\documentclass{article}

\usepackage[english]{babel}
\usepackage[utf8]{inputenc}
\usepackage{enumitem}
\usepackage{microtype}
\usepackage{graphicx}
\usepackage{booktabs} 
\usepackage{amsmath,amsfonts,amssymb}
\usepackage{mathtools}
\usepackage[round]{natbib}
\usepackage{url}            
\usepackage{nicefrac}       
\usepackage[dvipsnames]{xcolor}
\usepackage{multirow}
\usepackage{wrapfig}
\usepackage{setspace}
\usepackage{pifont}

\usepackage{algorithm}
\usepackage{algpseudocode}
\usepackage{titletoc}
\usepackage{capt-of}

\usepackage[backref=page,colorlinks=true,linkcolor=RedOrange, citecolor=NavyBlue]{hyperref}

\usepackage{xurl}
\hypersetup{pdftitle={From Task Mixtures to Specialized Experts},
               pdfauthor={Hojat Allah Salehi, Mehrdad Mahdavi, Andrew Arash Mahyari, M. Hadi Amini}}

\newtheorem{assumption}{Assumption}
\newtheorem{theorem}{Theorem}
\newtheorem{corollary}{Corollary}
\newtheorem{proposition}{Proposition}
\newtheorem{lemma}{Lemma}
\newtheorem{remark}{Remark}
\newtheorem{definition}{Definition}

\newcommand{\eqnref}[1]{Eq.~(\ref{#1})}
\newcommand{\Eqnref}[1]{Equation~(\ref{#1})}

\newcommand{\qed}{\hfill$\blacksquare$}

\makeatletter
\long\def\@makecaption#1#2{
  \vskip 0.8ex
  \setbox\@tempboxa\hbox{\small {\bf #1:} #2}
  \parindent 1.5em  
  \dimen0=\hsize
  \advance\dimen0 by -3em
  \ifdim \wd\@tempboxa >\dimen0
  \hbox to \hsize{
    \parindent 0em
    \hfil 
    \parbox{\dimen0}{\def\baselinestretch{0.96}\small
      {\bf #1.} #2
    } 
    \hfil}
  \else \hbox to \hsize{\hfil \box\@tempboxa \hfil}
  \fi
}
\makeatother

\title{\bf{\LARGE{From Task Mixtures to Specialized Experts}}}

\author{Hojat Allah Salehi$^{1,2}$ \quad Mehrdad Mahdavi$^{3}$ \quad Andrew Arash Mahyari$^{4}$ \quad M.~Hadi Amini$^{1,2}$ \vspace*{.4em} \\
$^{1}$Florida International University, Miami, FL, USA \\
$^{2}$Security, Optimization, and Learning for InterDependent \\networks laboratory (solid lab), Miami, FL, USA \\
$^{3}$The Pennsylvania State University, University Park, PA, USA \\
$^{4}$Florida Institute for Human and Machine Cognition (IHMC), Ocala, FL, USA \vspace*{.4em} \\
\texttt{hsalehi@fiu.edu, mzm616@psu.edu, amahyari@ihmc.us, amini@cs.fiu.edu}
}

\date{}
\begin{document}

\maketitle

\begin{abstract}
In collaborative foundation model fine-tuning, client data is rarely homogeneous. Instead, clients typically possess unknown mixtures of distinct data distributions, or \emph{tasks}. Conventional federated learning primarily addresses heterogeneity across clients without explicitly resolving latent task mixtures within each client. We study this setting as \emph{compound heterogeneity}, where data is heterogeneous both across and within clients. We study adaptation over a common frozen representation and show that, when tasks share the same feature geometry, the optimal model for a client's task mixture under squared loss is a convex combination of the optimal models for its underlying tasks. Thus, a single locally trained model represents the client's overall task mixture, while individual inputs may be drawn from different underlying task distributions. This motivates routing inputs to specialized experts, and we show that, when the task optima form a simplex, task-aligned routing achieves lower risk than any single adapted model for genuinely mixed clients. With access to a small set of task-labeled public samples, we derive a convex program to recover task experts and match them to their corresponding tasks. Our routing analysis shows that effective specialization requires input-dependent expert selection aligned with each client's task mixture.
Motivated by this analysis, we propose FedSEE. Across our experiments, FedSEE avoids the negative transfer observed in the evaluated baselines and improves performance by 2.9 points overall and 3.7 points for the worst-served quartile.
\end{abstract}

\section{Introduction}
\label{sec:intro}

Modern learning systems often adapt models using data distributed across users, institutions, or devices. A common distributed-learning abstraction assigns each participant a single local distribution $\mathcal{P}_i$, so non-IID structure arises across participants ($\mathcal{P}_i \neq \mathcal{P}_j$) \citep{McMahan2016Communication, fedprox, scaffold}. Yet a participant may itself come from multiple underlying data distributions, such as conversation, code, and prose for a personal assistant or records from different hospital departments. In this regime, heterogeneity exists both \emph{across} and \emph{within} participants, and each local dataset is a mixture whose composition varies across the population.

A latent-mixture view of client distributions has been studied in federated multi-task learning \citep{fedem}. We term the resulting combined within- and across-client variation \emph{compound heterogeneity}, modeling each client $i$'s data distribution $\mathcal{D}_i$ as an unknown mixture $\mathcal{D}_i=\sum_{m=1}^{M}\alpha_{im}\mathcal{P}_m$ of $M$ shared task distributions, where the latent mixture proportions $\alpha_i\in\Delta^{M-1}$ vary across clients. Specifically, we focus on the regime in which clients are themselves mixed and their mixture proportions differ across the population.

For such mixed clients, under a shared feature representation and squared loss, we show that the client optimum is a weighted combination of the task-specific optima, with weights given by the client's mixture proportions, $\theta_i^*=\sum_{m=1}^{M}\alpha_{im}\theta_m^*$ (Theorem~\ref{thm:mixture}). The resulting client optimum therefore captures the client's overall task mixture rather than recovering the individual task optima. In contrast, if the underlying task were known for each input, an ideal router could select the corresponding task optimum. When the task optima form a non-degenerate simplex, this idealized routing achieves strictly lower risk for every genuinely mixed client, while the two coincide for a pure client (Remark~\ref{rem:ideal_spec}). This shifts the learning objective from estimating one optimum per client to recovering the underlying task experts and learning to route among them. This naturally leads to a mixture-of-experts (MoE) formulation ~\citep{jacobs1991adaptive,shazeer2017outrageously} in which the task-specific optima define the experts and input-dependent routing determines how they are combined.

\begin{wrapfigure}{r}{0.45\textwidth}
\vspace{-15pt}
\centering
\includegraphics[width=\linewidth]{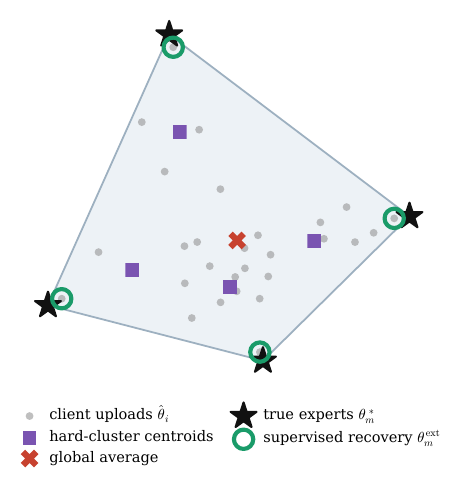}
\vspace{-20pt}
\caption{\textbf{Simplex geometry of compound heterogeneity.}
Task experts are vertices and client optima are barycentric points. Averaging stays inside the simplex, hard clustering collapses mixed clients to single clusters, and supervised extraction targets the vertices.}
\label{fig:geometry}
\vspace{-10pt}
\end{wrapfigure}

Recovering these task experts from mixed-client models is not straightforward. When the task optima form a simplex, the task experts correspond to its vertices and mixed-client optima lie within their convex hull. Averaging mixed-client optima therefore remains inside the simplex rather than recovering its vertices, while hard client clustering assigns each mixed client to a single cluster even when its optimum lies between task experts (Figure~\ref{fig:geometry}). The geometric structure that limits averaging and hard clustering also provides a direct geometric characterization of the task experts.

To recover the task experts, the server optimizes task-specific losses over convex combinations of the pooled client uploads. A small task-labeled public set provides these losses, allowing each extraction problem to target a specific task and yielding a convex recovery program (Theorem~\ref{thm:recovery}). The supervision also resolves the identity of the recovered experts. Without task-indexed information, the expert set may still be recovered, but one cannot determine which expert corresponds to which task (Theorem~\ref{thm:label}). As a result, experts learned on different clients cannot be reliably matched by task before aggregation.

Recovering the task experts is not sufficient for specialization because each input must also be routed to the expert corresponding to its underlying task. If inputs are routed to their corresponding task experts, the client's average expert usage naturally reflects its task mixture $\alpha_i$. Our analysis quantifies how both the client's average expert usage and the input-dependent routing decisions affect the resulting risk (Theorem~\ref{thm:routing_gap}). In particular, enforcing uniform average expert usage is incompatible with exact specialization when a client's task mixture is non-uniform (Proposition~\ref{prop:balanced_specialization}).

Motivated by this analysis, we propose \textbf{FedSEE} (Federated Supervised Expert Extraction), which combines task-specific LoRA experts~\citep{hu2022lora} with supervised extraction and mixture-aware routing. FedSEE shares the LoRA down-projection across experts, isolating task-specific adaptation in the remaining trainable parameters and enabling supervised expert extraction from mixed client updates. The resulting extraction weights indicate how strongly each client contributes to each recovered task expert, providing a proxy for the client's latent task mixture that guides expert selection and routing calibration. FedSEE also supports client-dependent training capacity by allowing each client to update only a subset of its available experts. We summarize our main contributions as follows. 
\begin{itemize}
    \item \textbf{Geometry under task mixtures.} We characterize how within-client task mixtures are represented after local optimization and show that, under the stated squared-loss conditions, each client optimum is a mixture-weighted combination of the underlying task optima, inducing a simplex geometry.

    \item \textbf{Expert recovery and identifiability.} We derive a task-supervised convex program for recovering task experts from mixed client uploads, together with recovery guarantees. We further characterize when task identities and the expert set itself cannot be determined from the available information.

    \item \textbf{Mixture-aware routing.} We characterize two requirements for specialization. Average expert usage should reflect the client's task mixture, while routing must remain input dependent to specialize individual inputs. 

    \item \textbf{FedSEE.} We instantiate these principles in a federated LoRA framework that extracts task-specific experts from mixed client updates and performs mixture-aware routing with client-specific calibration. Across our experiments, FedSEE improves over the strongest baseline by $2.9$ points overall and $3.7$ points on the worst-served quartile, and is the only evaluated method to avoid negative transfer across the five-point heterogeneity sweep.
\end{itemize}

\paragraph{Closest work.}
Related approaches span mixture modeling, federated expert construction, and centralized model composition. Mixture-based methods such as FedEM \citep{fedem} and FedSoft \citep{fedsoft} allow a client's data to combine multiple latent distributions, but jointly learn the corresponding components during federated optimization rather than recovering task-indexed experts from already mixed client adaptations. Hard clustering methods such as IFCA \citep{ifca} instead assign each client to a single component and therefore cannot directly represent clients spanning multiple tasks. Among federated LoRA and mixture-of-experts methods, FedLEASE \citep{fedlease} is closest to our setting, but constructs experts through client clustering and within-cluster aggregation rather than task-indexed recovery from mixed uploads. Centralized model and adapter composition methods \citep{ilharco2023editing,ties,zhao2025lego,huang2024lorahub} assume that task-specific models or adapters are already available and learn how to combine them. FedSEE addresses the inverse problem in which the observed client adaptations are mixtures of latent task experts that must first be recovered and identified. Appendix~\ref{app:related} provides a detailed comparison.

\section{Problem Formulation}
\label{sec:problem}
Consider $N$ clients and $M$ shared task distributions $\{\mathcal{P}_m\}_{m=1}^M$ on the sample space $\mathcal{X}\times\mathcal{Y}$. Under compound heterogeneity, the population distribution of client $i$ is
\begin{equation}
    \mathcal{D}_i \;=\; \sum\nolimits_{m=1}^{M} \alpha_{im}\mathcal{P}_m, \qquad \alpha_i \in \Delta^{M-1}, \label{eq:mixture}
\end{equation}
where $\alpha_i=(\alpha_{i1},\ldots,\alpha_{iM})$ is the client's latent task-mixture vector, unobserved by both the client and the server, and $\Delta^{M-1}$ is the probability simplex in $\mathbb{R}^M$. Throughout, $[K]:=\{1,\ldots,K\}$; $m,k\in[M]$ index tasks and $i,j\in[N]$ index clients.  We study personalization through a parameter block $\theta\in\mathbb{R}^{l\times r}$ acting linearly on a shared frozen representation. Let $f_0:\mathcal{X}\to\mathbb{R}^{l}$ denote the shared backbone and $\phi:\mathcal{X}\to\mathbb{R}^{r}$ the shared feature map. The personalized model is
\begin{equation}
    f_\theta(x) \;=\; f_0(x) \;+\; \theta\,\phi(x). \label{eq:model_class}
\end{equation}
For any non-negative measurable loss $\ell$, define the task and client population risks as
\begin{equation}
\begin{aligned}
    \mathcal{L}_m(\theta) &:= \mathbb{E}_{(x,y)\sim\mathcal{P}_m} [\ell(f_\theta(x),y)],\\
    \mathcal{L}_i(\theta) &:= \mathbb{E}_{(x,y)\sim\mathcal{D}_i} [\ell(f_\theta(x),y)] =\sum\nolimits_{m=1}^{M}\alpha_{im}\mathcal{L}_m(\theta).
\end{aligned}
\label{eq:risk_decomp}
\end{equation}
The final equality follows from linearity under~\eqnref{eq:mixture}, since the same loss and predictor are evaluated under each component distribution. For the exact geometry developed in Section~\ref{sec:theory}, we specialize to the squared loss $\ell(\hat y,y)=\|y-\hat y\|_2^2$. Appendix~\ref{app:general_loss} shows that for smooth general losses, including softmax cross-entropy, the optimum remains an approximate barycenter with deviation controlled by task-curvature dispersion. In this squared-loss setting, define the task and client population optima as
\begin{equation}
    \theta_m^* \in \arg\min_{\theta\in\mathbb{R}^{l\times r}} \mathcal{L}_m(\theta), \qquad \theta_i^* \in \arg\min_{\theta\in\mathbb{R}^{l\times r}} \mathcal{L}_i(\theta). \label{eq:optima}
\end{equation}
Under the conditions introduced in Section~\ref{sec:theory}, these minimizers exist and are unique.

In the finite-sample recovery abstraction, client $i$ holds $n_i$ i.i.d.\ samples from $\mathcal{D}_i$ and contributes a locally trained estimate $\hat{\theta}_i$ of $\theta_i^*$. The server observes the resulting pool $\{\hat{\theta}_i\}_{i=1}^{N}$ of parameter uploads together with the public reference sets. The learning objective is to recover the task experts $\{\theta_m^*\}_{m=1}^{M}$ and to route each input among the learned experts. Client $i$ uses a router $p_i:\mathcal{X}\to\Delta^{M-1}$, where $p_{im}(x)$ is the weight assigned to expert $m$, yielding,
\begin{equation}
    f_i(x) = f_0(x) + \sum\nolimits_{m=1}^{M} p_{im}(x)\,\theta_m\,\phi(x), \label{eq:objective}
\end{equation}
where $\{\theta_m\}_{m=1}^{M}$ denote the learned expert parameters, with exact recovery corresponding to $\theta_m=\theta_m^*$. Equivalently, $f_i(x)=f_{\theta_i^{\mathrm{rt}}(x)}(x)$ with $\theta_i^{\mathrm{rt}}(x):=\sum_m p_{im}(x)\theta_m \in\operatorname{conv}\{\theta_m\}_{m=1}^{M}$, extending the fixed parameter in~\eqnref{eq:model_class} to an input-dependent convex combination. We denote the resulting population risk by $\mathcal{R}_i := \mathbb{E}_{(x,y)\sim\mathcal{D}_i}[\ell(f_i(x),y)]$. Thus the learning problem has two coupled components: \emph{expert recovery} and \emph{input-dependent routing}. To make expert recovery task-indexed, we assume a small labeled reference set for each task.

\begin{assumption}[Task-Labeled Public Data]
\label{ass:public}
For each task $m\in[M]$, the server has access to a labeled reference set $\mathcal{D}_m^s$ containing $n_m$ i.i.d.\ samples from $\mathcal{P}_m$.
\end{assumption}

In standard benchmark settings, small labeled reference splits are commonly available. We define the empirical task-$m$ public risk as $\widehat{\mathcal{L}}_m^s(\theta) :=(1/n_m)\sum_{(x,y)\in\mathcal{D}_m^s} \ell(f_\theta(x),y)$. These task-specific losses provide the supervision used to associate recovered experts with their underlying tasks and to guide expert extraction.

\section{A General Framework for Federated Mixture Personalization}
\label{sec:theory}

We first characterize the geometry induced by the mixture model in~\eqnref{eq:mixture} and the model class in~\eqnref{eq:model_class}. We then use this structure to study supervised expert recovery and its identifiability limits, and finally derive the routing requirements for task-aligned specialization. Complete proofs and supporting theoretical results appear in Appendix~\ref{app:theory}.

\subsection{Barycentric Geometry of Client Optima}
\label{sec:thm1}

We analyze the squared-loss model under the following conditions on the shared representation and task distributions.
\begin{assumption}[Shared-Feature Linear Regression]
\label{ass:linear}
\emph{(i) Task realizability.} For each task $m$, data drawn from $\mathcal{P}_m$ satisfy $y = f_0(x) + \theta_m^*\phi(x) + \zeta_m$, with $\mathbb{E}[\zeta_m\mid x]=0$ and $\mathbb{E}[\zeta_m\zeta_m^\top\mid x]=\sigma_m^2 I_l$. \emph{(ii) Shared feature second moment.} $\Phi_m := \mathbb{E}_{\mathcal{P}_m}[\phi(x)\phi(x)^\top] = \Phi$ for all $m$. \emph{(iii) Feature non-degeneracy.} $\Phi \succ 0$.
\end{assumption}
Part (i) makes each task regression well specified relative to the shared representation. Part (ii) gives the exact common quadratic geometry. When the $\Phi_m$ differ, exact barycentricity is replaced by a controlled perturbation governed by the relative covariance dispersion $\rho_\Phi$ (Remark~\ref{rem:heterogeneous} in Appendix~\ref{app:geometry}). We write $\|C\|_\Phi^2 := \operatorname{tr}(C\Phi C^\top)$. Part (iii) ensures unique population minimizers.

\begin{theorem}[Barycentric Characterization of Client Optima]
\label{thm:mixture}
Under Assumption~\ref{ass:linear}, client $i$'s squared-loss population risk $\mathcal{L}_i$ has a unique minimizer given by
\begin{equation}
    \theta_i^* \;=\; \sum\nolimits_{m=1}^{M}\alpha_{im}\theta_m^*. \label{eq:thm1}
\end{equation}
\end{theorem}

Theorem~\ref{thm:mixture} shows that local optimization preserves the latent mixture weights in parameter space, as each client optimum is a barycentric combination of the task experts and therefore lies in their convex hull. To characterize this geometry, define
\begin{equation}
    \mathrm{sep} \;:=\; \min_{m \neq m'} \|\theta_m^* - \theta_{m'}^*\|_\mathrm{F}, \qquad h \;:=\; \min_{m \in [M]} \mathrm{dist}_\mathrm{F}\!\bigl(\theta_m^*,\, \mathrm{aff}\{\theta_k^* : k \neq m\}\bigr). \label{eq:sep_h}
\end{equation}
Here $\mathrm{sep}$ measures pairwise expert separation, while $h$ measures the minimum affine height. When $h>0$, the experts are affinely independent, their convex hull is a non-degenerate simplex, and $\alpha_i$ is the unique barycentric coordinate vector of $\theta_i^*$. Appendix~\ref{app:geometry} details the resulting face and mixing-gap geometry.

Under $h>0$, a mixed client's optimum is a non-vertex barycentric point. A task-oracle router that selects $\theta_m^*$ for each task-$m$ input incurs only the irreducible noise, with risk $\Lambda_i := \sum_m \alpha_{im}\sigma_m^2 l$. In contrast, the best fixed adapter $\theta_i^*$ incurs the mixing gap $G_i := \mathcal{L}_i(\theta_i^*)-\Lambda_i = \sum_m \alpha_{im}\|\theta_i^*-\theta_m^*\|_\Phi^2,$ which is strictly positive for every mixed client and vanishes exactly for a pure client (Proposition~\ref{prop:simplex_geometry}, Appendix~\ref{app:geometry}). This oracle gap motivates \emph{task-aligned specialization}. Approaching it with an input-dependent router additionally requires routing inputs toward their corresponding experts, as characterized in Section~\ref{sec:routing}.

\subsection{Supervised Expert Recovery and Identifiability}
\label{sec:recovery}

Theorem~\ref{thm:mixture} places client optima in the convex hull of the task experts. The task-indexed public losses allow the server to extract each expert by minimizing its task risk over convex combinations of the uploaded models. Using the population loss for task $m$, define

\begin{equation}
    \omega_m^* \in \underset{\omega_m\in\Delta^{N-1}}{\arg\min}\; \mathcal{L}_m\!\left( \sum\nolimits_{i=1}^{N}\omega_{m,i}\hat{\theta}_i \right), \qquad \theta_m^{\mathrm{ext}} = \sum\nolimits_{i=1}^{N}\omega_{m,i}^*\hat{\theta}_i . \label{eq:gen_extract}
\end{equation}

Under the squared-loss model, the task population risk satisfies $\mathcal{L}_m(\theta)-\mathcal{L}_m(\theta_m^*) =\|\theta-\theta_m^*\|_\Phi^2$, so~\eqnref{eq:gen_extract} computes the unique $\Phi$-projection of $\theta_m^*$ onto the convex hull of the uploads. The coefficients $\omega_m^*$ may be nonunique even though the recovered expert is unique. We next introduce a structural condition that enables a recovery guarantee.

\begin{assumption}[Near-Pure Client Anchors]
\label{ass:anchor}
For each task $m\in[M]$, there exists a client $i_m\in[N]$ satisfying $\alpha_{i_m,m}\geq 1-\varepsilon$ for some $\varepsilon\in[0,1/2)$.
\end{assumption}

This is an approximate separability condition analogous to anchor assumptions in topic modeling and separable non-negative matrix factorization~\citep{nmf}. It requires only one near-pure client per task. By Theorem~\ref{thm:mixture}, each such client optimum lies near the corresponding task expert and therefore provides a nearby feasible point for~\eqnref{eq:gen_extract}.

\begin{theorem}[Population Expert Recovery]
\label{thm:recovery}
Under Assumptions~\ref{ass:linear} and~\ref{ass:anchor}, with exact client uploads $\{\theta_i^*\}$ and population task losses, the extractor in~\eqnref{eq:gen_extract} satisfies $\|\theta_m^{\mathrm{ext}}-\theta_m^*\|_\mathrm{F} \leq \kappa(\Phi)^{1/2}\varepsilon\,\mathrm{diam}_\mathrm{F}(\Theta)$ for every $m\in[M]$, where $\kappa(\Phi)=\sigma_{\max}(\Phi)/\sigma_{\min}(\Phi)$ and $\mathrm{diam}_\mathrm{F}(\Theta)=\max_{m,k}\|\theta_m^*-\theta_k^*\|_\mathrm{F}$ for $\Theta=\{\theta_m^*\}_{m=1}^M$.
\end{theorem}

The recovery error decreases linearly with anchor impurity and vanishes for pure anchors. With estimated client uploads and finite public data, additional client-estimation and public-sampling errors arise and vanish as the corresponding sample sizes grow (Proposition~\ref{prop:finitesample}, Appendix~\ref{app:recovery}).

\paragraph{Two identifiability limits.}
The supervised extractor is task-indexed because its $m$th output is defined using the corresponding task loss $\mathcal{L}_m$. To isolate the role of this supervision, call an estimator \emph{label-blind} if it receives no task-indexed side information, and \emph{pool-only} if it uses only client uploads. We first show that label-blind observations cannot identify which expert corresponds to which task, even when the expert set itself is recoverable.

\begin{theorem}[Task-Identity Indeterminacy]
\label{thm:label}
Let $M\geq2$ and let $\mathcal O(\mathcal I):=\{\mathcal I^\pi:\pi\in S_M\}$ denote the permutation orbit of an instance $\mathcal I$. Every $\mathcal I^\pi\in\mathcal O(\mathcal I)$ induces identical label-free observables. Consequently, for any label-blind estimator $\mathcal A$,
\begin{equation}
\max_{\mathcal J\in\mathcal O(\mathcal I)} \mathbb E_{\mathcal A}\!\left[ \max_{m\in[M]} \|\hat\theta_m-\theta_m^*(\mathcal J)\|_{\mathrm F} \right] \geq \frac{1}{2}\mathrm{diam}_{\mathrm F}(\Theta) \geq \frac{\mathrm{sep}}{2}. \label{eq:label_lower_bound}
\end{equation}
This holds for every $N$, every anchor quality $\varepsilon$, including $\varepsilon=0$, and every amount of local data and computation.
\end{theorem}

Thus even perfect recovery of the vertex set cannot determine its task correspondence without task-indexed supervision. We next ask whether the unordered expert set itself can be recovered from client uploads.

\begin{theorem}[Vertex-Set Non-Identifiability from Interior Mixtures]
\label{thm:vertex}
Suppose $h>0$ and every client is $\gamma$-interior for some $\gamma\in(0,1]$, so that $\alpha_{im}\geq\gamma/M$ for all $i\in[N]$ and $m\in[M]$. Let $t:=\min\{\gamma,\mathrm{sep}/(2\,\mathrm{diam}_\mathrm{F}(\Theta))\}$. Then there exists a distinct valid instance $\tilde{\mathcal I}$ with its own expert set and mixture weights that induces the same upload pool as $\mathcal I$. For any pool-only estimator $\mathcal A$, even under the most favorable relabeling,
\begin{equation}
\max_{\mathcal J\in\{\mathcal I,\tilde{\mathcal I}\}} \mathbb E_{\mathcal A}\!\left[ \min_{\pi\in S_M}\max_{m\in[M]} \|\hat\theta_{\pi(m)}-\theta_m^*(\mathcal J)\|_\mathrm{F} \right] \geq \frac{M-1}{2M}\,t\,h .
\end{equation}
\end{theorem}

The two results isolate distinct requirements. Theorem~\ref{thm:label} shows that task-indexed side information is needed to identify the expert-to-task correspondence. Theorem~\ref{thm:vertex} shows that interior upload pools can be consistent with different expert sets, so upload-only vertex recovery requires sufficient boundary information. Near-pure client anchors provide a sufficient form of such information.

\subsection{Routing Requirements for Specialization}
\label{sec:routing}

After recovering the task experts, specialization also requires routing each input toward the appropriate expert. A task-aligned router should therefore satisfy two complementary properties. Its mean expert usage should reflect the client's task mixture $\alpha_i$, while its routing decisions should remain input dependent so that different tasks can activate different experts. We next characterize how these two properties affect personalized risk.

\begin{theorem}[Mean Calibration and Input-Dependent Routing]
\label{thm:routing_gap}
Suppose Assumption~\ref{ass:linear} holds, $h>0$, and $\|\phi(x)\|_2\leq Q$ almost surely. Let $D_\Theta:=\mathrm{diam}_{\mathrm F}(\Theta)$ and let $p_i:\mathcal X\to\Delta^{M-1}$ be any input-dependent routing policy. Define its mean usage as $\bar p_i:=\mathbb E_{x\sim\mathcal D_i}[p_i(x)]$ and its routing variation as $V_i(p_i):=\mathbb E_{x\sim\mathcal D_i} \|p_i(x)-\bar p_i\|_1$. Then the population risk $\mathcal R_i$ of routing over the true experts satisfies
\begin{align}
\sigma_{\min}(\Phi)h^2 \|\bar p_i-\alpha_i\|_\infty^2 -Q^2D_\Theta^2V_i(p_i) &\leq \mathcal R_i-\mathcal L_i(\theta_i^*) \nonumber\\
&\leq \frac{\sigma_{\max}(\Phi)D_\Theta^2}{4} \|\bar p_i-\alpha_i\|_1^2 +Q^2D_\Theta^2V_i(p_i). \label{eq:routing_gap}
\end{align}
\end{theorem}

Theorem~\ref{thm:routing_gap} separates the effects of mean calibration and input-dependent routing. If $p_i(x)\equiv\alpha_i$, the router recovers the optimal static adapter $\theta_i^*$ but cannot improve on it. Closing the mixing gap therefore requires input-dependent routing that is aligned with the underlying tasks. Corollary~\ref{cor:oracle_risk} shows that such task-aligned routing approaches the oracle risk as the routing and expert-recovery errors vanish.

\begin{proposition}[Uniform Mean Usage Limits Specialization]
\label{prop:balanced_specialization}
Suppose the $M$ experts admit a task matching $\pi$ such that, for every task $m$ present at client $i$, the router assigns its matched expert average probability at least $1-\beta$ on inputs from $\mathcal{P}_m$. Then $\|\bar p_i-\Pi_\pi\alpha_i\|_1\leq 2\beta$, where $\Pi_\pi$ is the permutation matrix for this matching. If additionally $\|\bar p_i-u\|_1\leq\delta_{\rm bal}$ for $u=\frac{1}{M}\mathbf 1$, then $\beta\geq\frac{1}{2}(\|\alpha_i-u\|_1-\delta_{\rm bal})_+$. Hence exact uniform mean usage is incompatible with exact specialization for any client with $\alpha_i\neq u$.
\end{proposition}

Proposition~\ref{prop:balanced_specialization} therefore identifies the client mixture $\alpha_i$, rather than uniform usage, as the appropriate mean-routing target for specialization. Since $\alpha_i$ is latent, FedSEE uses a contribution-based proxy $\tilde c_i$ for mixture calibration. Let $\hat{\alpha}_i$ denote the barycentric estimate of Proposition~\ref{prop:mixture_est}, satisfying $\|\hat{\alpha}_i-\alpha_i\|_1\leq\eta_C$, and define $\delta_i^{\mathrm{proxy}} :=\|\tilde c_i-\hat{\alpha}_i\|_1$. Then $\|\tilde c_i-\alpha_i\|_1 \leq\eta_C+\delta_i^{\mathrm{proxy}}$. Consequently, Pinsker's inequality gives $\|\bar p_i-\alpha_i\|_1 \leq \sqrt{2\,\mathrm{KL}(\tilde c_i\|\bar p_i)} +\eta_C+\delta_i^{\mathrm{proxy}}$ (Corollary~\ref{cor:kl_justify}). Thus anchoring mean routing toward $\tilde c_i$ controls mixture calibration up to the certified mixture-estimation error and the additional proxy error. Input-level specialization remains the responsibility of the task loss, as characterized by Corollary~\ref{cor:oracle_risk}. Section~\ref{sec:FedSEE} instantiates these two complementary objectives.

\section{FedSEE: The LoRA Instantiation}
\label{sec:FedSEE}
\begin{figure*}[t]
    \centering
    \includegraphics[width=\textwidth]{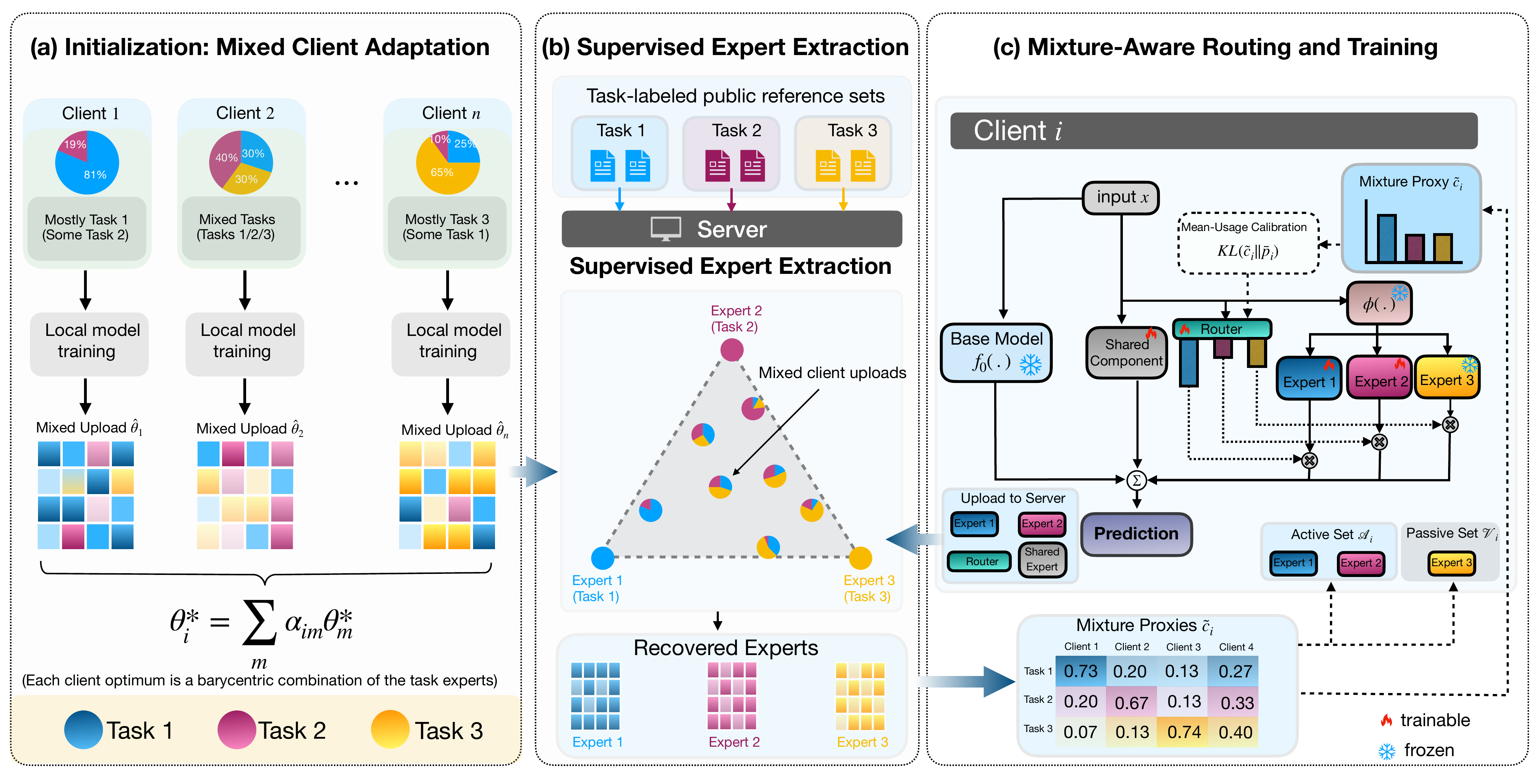}
    \vspace{-10pt}
    \caption{\textbf{FedSEE overview.} Supervised extraction recovers task experts from mixed client updates, followed by mixture-aware routing with client-specific calibration.}
    \label{fig:fedsee_overview}
    \vspace{-10pt}
\end{figure*}

\textbf{FedSEE} (Federated Supervised Expert Extraction) instantiates the framework with LoRA, as summarized in Figure~\ref{fig:fedsee_overview}. At a single adapted projection with $A^{\mathrm{global}}$ fixed, the routed LoRA component matches \eqnref{eq:model_class} with $\phi(x)=sA^{\mathrm{global}}x$ and $\theta=B$. Under the conditions of Theorem~\ref{thm:mixture}, this gives $B_i^*=\sum_m\alpha_{im}B_m^*$ exactly at the projection level. For the composed transformer, Appendix~\ref{app:fedsee_reduction} bounds the departure from this geometry under a local first-order approximation.

FedSEE also maintains an always-on shared LoRA expert $(A^{\mathrm{s}},B^{\mathrm{s}})$ to capture task-shared adaptation outside the routed expert pool. Its deployed forward pass is
\begin{equation}
    h(x) = \underbrace{W_0x+sB^{\mathrm{s}}A^{\mathrm{s}}x}_{\tilde f_0(x)} +\, s\sum\nolimits_{m\in\mathcal U_i} p_{im}(x)B_mA^{\mathrm{global}}x , \label{eq:forward_shared}
\end{equation}
where $s=\alpha_{\mathrm{LoRA}}/r$ and $\mathcal U_i$ is client $i$'s inference set. The shared expert is always active and is not routed. For a single projection, \eqnref{eq:forward_shared} therefore matches the routed model in~\eqnref{eq:objective} with the shared term absorbed into $\tilde f_0$.

\paragraph{Supervised expert extraction.}
FedSEE begins with one initialization round in which each client trains a common LoRA pair $(A^{(0)},B^{(0)})$ and returns $(A_i,B_i)$. The server sets $A^{\mathrm{global}}=N^{-1}\sum_i A_i$ and keeps it fixed thereafter. Let $\mathcal I^{(t)}:=\{(i,m') : B_{m'}^i \text{ is uploaded at round }t\}$ denote the current upload pool. Using the task-labeled public loss, expert $m$ is extracted as
\begin{equation}
\omega_m^* \!\in \!\underset{\omega_m\in\Delta^{|\mathcal I^{(t)}|-1}}{\arg\min}\! \widehat{\mathcal L}_m^s\!\left( \sum\nolimits_{(i,m')\in\mathcal I^{(t)}} \!\!\omega_{m,(i,m')}B_{m'}^i \right), \qquad B_m^{\mathrm{expert}} = \!\!\sum\nolimits_{(i,m')\in\mathcal I^{(t)}} \omega_{m,(i,m')}^*B_{m'}^i . \label{eq:omega_opt}
\end{equation}

FedSEE optimizes the $M$ taskwise programs jointly while constraining the raw client contributions $c_{mi}:=\sum_{m':(i,m')\in\mathcal I^{(t)}}\omega_{m,(i,m')}$ to satisfy $\sum_i c_{mi}=1$ and $\sum_m c_{mi}=M/N$. Balanced transport prevents repeated concentration on the same clients but can exclude independently optimal extraction weights. Theorem~\ref{thm:recovery} applies to independent extraction, while Proposition~\ref{prop:balanced_coupling} quantifies the coupling excess and gives exact recovery for transport-compatible pure anchors. Solver details appear in Appendix~\ref{app:fedsee_sinkhorn}. The same contribution matrix $C=[c_{mi}]$ also provides a client-specific relevance signal for capacity allocation and router calibration.

\paragraph{Mixture-aware capacity and routing.}
For client $i$, the contribution column $c_{\cdot i}$ ranks the recovered experts by their relevance to that client. FedSEE assigns the highest-ranked experts to an inference set $\mathcal U_i$ and selects an active subset $\mathcal A_i\subseteq\mathcal U_i$ for local updates. Experts in $\mathcal U_i\setminus\mathcal A_i$ remain available to the router but are detached during backpropagation, allowing inference and training capacity to differ. Normalizing the retained contributions over $\mathcal U_i$ gives the mixture proxy $\tilde c_i$ used to calibrate the router. Further details of the capacity assignment appear in Appendix~\ref{app:fedsee_capacity}, while Appendix~\ref{app:capacity_analysis} examines its empirical training-capacity tradeoff. Let $\psi_i=\{W_r^\ell,b_r^\ell\}_{\ell\in\mathcal P}$ denote the client $i$'s local copy of the per-layer router parameters. Local training uses
\begin{equation}
\min_{\{B_m\}_{m\in\mathcal A_i},\,(A^{\mathrm{s}}_i,B^{\mathrm{s}}_i),\,\psi_i}
\!\!\!\mathcal L_{\mathrm{task}}\!\left( \{B_m\}_{m\in\mathcal U_i},\psi_i;\mathcal D_i \right) + \frac{\beta_{\mathrm{KL}}}{|\mathcal P|} \sum\nolimits_{\ell\in\mathcal P} \mathrm{KL}\!\left(\tilde c_i\|\bar p_i^\ell\right), \quad \bar p_i^\ell:=\mathbb E_{x\sim\mathcal D_i}[p_i^\ell(x)] . 
\label{eq:client_obj}
\end{equation}

The shared pair $(A^{\mathrm{s}}_i,B^{\mathrm{s}}_i)$ is updated on the same objective and aggregated by FedAvg. The task loss learns input-dependent expert assignments, while the KL term calibrates mean usage toward the client-specific proxy $\tilde c_i$, as motivated by Section~\ref{sec:routing}. No uniform mean-usage penalty is used. After each round the routers are aggregated across clients, with the corresponding justification deferred to Appendix~\ref{app:fedsee_router}. For evaluation, client $i$ uses its personalized model \eqnref{eq:forward_shared}, comprising the locally updated active experts, shared expert, and router together with the recovered passive experts in $\mathcal U_i$. The complete procedure appears in Algorithm~\ref{alg:FedSEE} in Appendix~\ref{app:fedsee_algorithm}.

\section{Experiments}
\label{sec:experiments}
\paragraph{Synthetic validation.}
Under the assumptions of Section~\ref{sec:theory}, controlled experiments instantiate the simplex geometry illustrated in Figure~\ref{fig:geometry}. They confirm the main theoretical predictions. Expert-recovery error scales linearly with anchor impurity, task supervision resolves the identity ambiguity of label-free recovery, sufficiently interior upload pools exhibit the non-identifiability of Theorem~\ref{thm:vertex}, and mixture-aware routing avoids the cost of uniform calibration. Appendix~\ref{app:synthetic} provides the full construction and additional results.

\subsection{Natural Language Understanding Benchmark}
\label{sec:setup}

We evaluate FedSEE on six GLUE/SuperGLUE tasks~\citep{wang-etal-2018-glue,wang2019superglue}: SST-2, QNLI, CoLA, QQP, WiC, and BoolQ, forming $M=6$ tasks. Client task proportions follow a symmetric Dirichlet distribution with concentration $\alpha_{\mathrm{Dir}}$, where smaller values produce more concentrated mixtures and larger values approach uniform composition. Each client has disjoint 200-example validation and test sets matched to its realized task mixture. The validation set is used only for model selection, while the test set is held out for final personalized evaluation.

FedSEE assumes access to a small task-labeled public reference set, used only for server-side expert extraction in~\eqnref{eq:omega_opt}. We use $1{,}000$ public examples per task by default, all disjoint from client training, validation, and test data. Appendix~\ref{app:ablations} studies sensitivity to the amount of public supervision and shows that the advantage is retained with only 100 labeled examples per task. By default, all six recovered experts are available at inference while each client updates one active expert, denoted \textbf{FedSEE (L1)}. We compare against averaging methods (\textbf{FedIT}~\citep{zhang2024fedit}, \textbf{FFA-LoRA}~\citep{sun2024ffa}), personalized LoRA methods (\textbf{FedSA-LoRA}~\citep{guo2025fedsalora}, \textbf{FedDPA-T}~\citep{feddpa}), and clustering methods (\textbf{IFCA+LoRA}~\citep{ifca}, \textbf{FedLEASE}~\citep{fedlease}). Implementation details appear in Appendix~\ref{app:fedsee_hparams}, with conceptual comparisons in Appendix~\ref{app:related}.

\subsection{Compound Heterogeneity at
\texorpdfstring{$\alpha_{\mathrm{Dir}} = 0.1$}{alpha=0.1}}
\label{sec:main_results}

\paragraph{Protocol.}
We use RoBERTa-large~\citep{liu2019roberta} with $N=10$ fully participating clients, $T=10$ communication rounds, and $E=5$ local epochs. Each method selects its learning rate from $\{5\times10^{-4},10^{-3},5\times10^{-3}\}$ and is evaluated over three random seeds. Complete optimization and model-selection details appear in Appendix~\ref{app:fedsee_hparams}. We report overall personalized accuracy together with \emph{Worst-50\%} and \emph{Worst-25\%}, averaging the lowest $\lceil N/2\rceil$ and $\lceil N/4\rceil$ client accuracies, respectively. Per-task results appear in Appendix~\ref{app:main_results}.

\paragraph{Results.}
FedSEE achieves the highest overall personalized accuracy (Table~\ref{tab:main_results_combined}), improving over the strongest baseline from $79.75\%$ to $82.65\%$, a gain of $2.90$ points. FedSEE improves Worst-50\% accuracy from $72.46\%$ to $76.54\%$ and Worst-25\% accuracy from $69.10\%$ to $72.75\%$, corresponding to gains of $4.08$ and $3.65$ points, respectively. The strongest baseline varies by metric: FedSA-LoRA overall and FedLEASE on both client-level tail metrics. Component ablations appear in Appendix~\ref{app:ablations}.

\begin{table}[t]
\centering
\caption{Personalized accuracy (\%) under compound heterogeneity ($\alpha_{\mathrm{Dir}}=0.1$) with RoBERTa-large. Results are reported as mean $\pm$ standard deviation over three seeds.}
\label{tab:main_results_combined}
\renewcommand{\arraystretch}{0.9}
\setlength{\tabcolsep}{12pt}
\begin{tabular}{lccc}
\toprule
\textbf{Method} & \textbf{Overall} & \textbf{Worst-50\%} & \textbf{Worst-25\%} \\
\midrule
FedIT & $78.06 \pm 2.04$ & $69.35 \pm 5.17$ & $64.44 \pm 6.09$ \\
FFA-LoRA & $73.43 \pm 1.09$ & $64.64 \pm 2.27$ & $62.21 \pm 2.49$ \\
IFCA+LoRA & $79.42 \pm 0.99$ & $70.95 \pm 2.42$ & $67.37 \pm 1.49$ \\
FedSA-LoRA & $\underline{79.75 \pm 1.37}$ & $71.89 \pm 2.60$ & $66.78 \pm 2.83$ \\
FedDPA-T & $79.20 \pm 1.00$ & $71.40 \pm 3.74$ & $67.34 \pm 4.20$ \\
FedLEASE & $79.52 \pm 1.80$ & $\underline{72.46 \pm 3.99}$ & $\underline{69.10 \pm 3.77}$ \\
\midrule
\textbf{FedSEE (L1, Ours)} & $\mathbf{82.65 \pm 0.35}$ & $\mathbf{76.54 \pm 1.65}$ & $\mathbf{72.75 \pm 2.21}$ \\
\midrule
\textbf{Gain over best baseline} & $\mathbf{+2.90}$ & $\mathbf{+4.08}$ & $\mathbf{+3.65}$ \\
\bottomrule
\end{tabular}
\end{table}

\subsection{Where the Advantage Holds: A Heterogeneity Sweep}
\label{sec:alpha_sweep}
We next evaluate robustness from pure single-task clients to near-uniform mixtures. This sweep tests whether FedSEE's task-expert representation remains effective across regimes that increasingly favor client partitioning at one extreme and global aggregation at the other. We use \texttt{distilroberta-base} with $N=12$ clients and evaluate $\alpha_{\mathrm{Dir}}\in\{0,0.1,0.5,1.0,50\}$, denoting $\alpha_{\mathrm{Dir}}=50$ as \emph{uniform}. All methods use matched client partitions and random seeds. We also report \emph{local-only} training and a centralized task-specific reference trained on pooled task data. Table~\ref{tab:combined_alpha_sweep} reports FedSEE's absolute accuracy and paired differences from competing methods. Full construction appears in Appendix~\ref{app:fedsee_hparams}.

 \begin{wrapfigure}{r}{0.55\textwidth}
\centering
\vspace{-30pt}
\includegraphics[width=\linewidth]{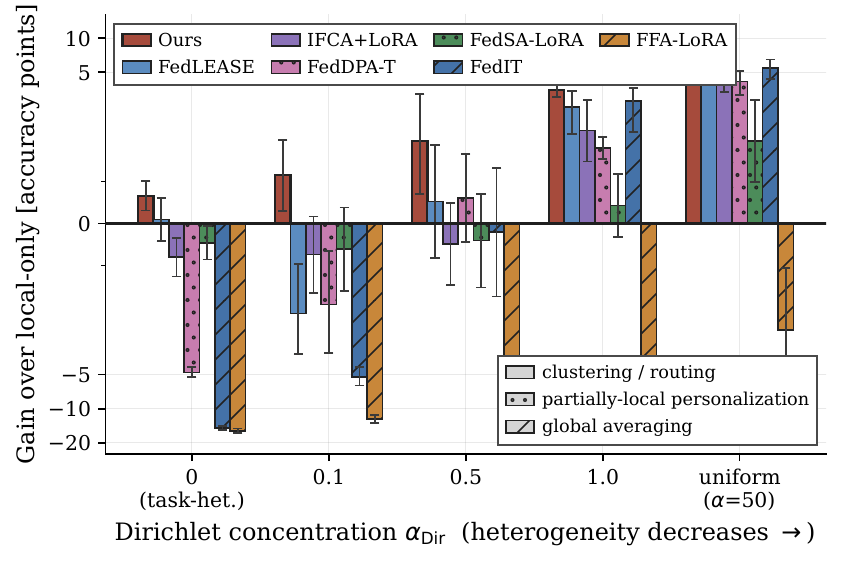}
\vspace{-20pt}
\caption{Gain over local-only training across heterogeneity ($y$-axis uses symlog scaling).}
    \label{fig:gain_over_floor}
\vspace{-15pt}
\end{wrapfigure}

\paragraph{Robustness across mixture regimes.}
FedSEE achieves the highest federated accuracy in four of five regimes and is within $0.38$ points of the strongest baseline at the near-uniform endpoint. The closest baseline changes across the sweep, while FedSEE uses the same task-expert representation and client-dependent routing throughout.

\paragraph{Positive transfer across heterogeneity.}
Figure~\ref{fig:gain_over_floor} shows that FedSEE improves over local-only training in every regime, with gains increasing from $0.65$ points for pure clients to $5.33$ points near uniform. In contrast, every competing federated method falls below local-only training in at least one regime. The training-capacity tradeoff across the heterogeneity sweep is analyzed in Appendix~\ref{app:capacity_analysis}.

\begin{table*}[t]
\centering
\caption{Personalized accuracy (\%) across the heterogeneity sweep.
Differences are method minus FedSEE, reported as mean $\pm$ standard deviation over three matched seeds.}
\vspace{5pt}
\label{tab:combined_alpha_sweep}
\renewcommand{\arraystretch}{1.05}
\setlength{\tabcolsep}{4pt}
\resizebox{\linewidth}{!}{%
\begin{tabular}{l|ccccc}
\toprule
\textbf{Method} & $\alpha_{\mathrm{Dir}}\!=\!0$ & $0.1$ & $0.5$ & $1.0$ & \textbf{uniform} \\
\midrule
\textbf{FedSEE (L1, Ours)} & $\mathbf{76.46}$ & $\mathbf{74.95}$ & $\mathbf{71.28}$ & $\mathbf{72.10}$ & $69.97$ \\
\midrule
\multicolumn{6}{l}{\emph{paired difference from FedSEE (L1)}} \\
\midrule
FedIT & $-15.47 \pm 0.46$ & $-6.40 \pm 1.07$ & $-2.18 \pm 0.42$ & $\underline{-0.60 \pm 1.02}$ & $+0.10 \pm 0.22$ \\

FFA-LoRA & $-16.30 \pm 0.62$ & $-13.43 \pm 1.09$ & $-9.16 \pm 2.15$ & $-10.14 \pm 2.20$ & $-7.86 \pm 1.33$ \\

FedSA-LoRA & $-0.83 \pm 0.42$ & $\underline{-1.59 \pm 0.65}$ & $-2.37 \pm 0.46$ & $-3.07 \pm 0.95$ & $-3.26 \pm 0.92$ \\

FedDPA-T & $-5.44 \pm 0.51$ & $-3.08 \pm 1.57$ & $\underline{-1.36 \pm 0.62}$ & $-1.72 \pm 0.83$ & $-1.18 \pm 0.31$ \\

IFCA+LoRA & $-1.46 \pm 0.39$ & $-1.89 \pm 0.87$ & $-2.46 \pm 0.44$ & $-1.31 \pm 0.43$ & $-0.19 \pm 2.09$ \\

FedLEASE & $\underline{-0.57 \pm 0.56}$ & $-3.29 \pm 1.41$ & $-1.44 \pm 0.33$ & $-0.74 \pm 0.47$ & $\underline{+0.38 \pm 1.07}$ \\
\midrule

\textit{local-only} & \textit{$-0.65 \pm 0.20$} & \textit{$-1.15 \pm 1.00$} & \textit{$-1.96 \pm 0.68$} & \textit{$-3.52 \pm 0.73$} & \textit{$-5.33 \pm 0.90$} \\

\textit{centralized} & \textit{$-0.67 \pm 0.27$} & \textit{$+0.21 \pm 0.29$} & \textit{$+3.76 \pm 1.18$} & \textit{$+3.52 \pm 0.50$} & \textit{$+6.15 \pm 0.74$} \\
\bottomrule
\end{tabular}%
}
\end{table*}

\section{Conclusion and Future Work}
\label{sec:conclusion}
We studied personalization under compound heterogeneity, where clients may contain mixtures of latent tasks. Under our shared-feature quadratic model, client optima are mixture-weighted barycenters of task optima, with affine independence inducing a simplex geometry. This exposes limitations of single-adapter aggregation and label-free expert construction, motivating task-supervised expert recovery and mixture-aware, input-dependent routing. FedSEE instantiates these principles with LoRA by extracting task-aligned experts from mixed client updates and routing them using client-specific mixture calibration. On natural language understanding tasks, FedSEE improves average and tail personalization and remains competitive from pure to near-uniform client populations. These results support explicit modeling of within-client mixture structure across heterogeneity regimes.

The present analysis suggests two natural directions for extending the framework. First, the exact barycentric characterization holds under the squared-loss model, while the general-loss analysis of Appendix~\ref{app:general_loss} introduces a perturbation controlled by the variation in task curvature. A sharper characterization of this perturbation, particularly for deep models trained with cross-entropy, would further connect the theoretical framework to practical fine-tuning settings. Second, the framework assumes that the number of underlying tasks $M$ is known. In practical federated settings, this quantity may itself need to be inferred from the client population. Extending supervised expert extraction to jointly estimate the number of latent tasks and their associated experts is an important direction for future work.

\section*{Acknowledgments}
This work is based upon the work partly supported by the National Center for Transportation Cybersecurity and Resiliency (TraCR) (a U.S. Department of Transportation National University Transportation Center) headquartered at Clemson University, Clemson, South Carolina, USA. Any opinions, findings, conclusions, and recommendations expressed in this material are those of the author(s) and do not necessarily reflect the views of TraCR, and the U.S. Government assumes no liability for the contents or use thereof. 

The work of Mehrdad Mahdavi was partially supported by National Science Foundation CAREER Award \#2239374. Any opinions, findings, and conclusions or recommendations expressed in this material are those of the author(s) and do not necessarily reflect the views of the National Science Foundation.

\bibliography{references}
\bibliographystyle{plainnat}

\newpage
\onecolumn
\appendix
\startcontents[appendix]

\section*{Appendix Contents}

\printcontents[appendix]{}{1}[2]{}
\clearpage

\section{Related Work}
\label{app:related}

\paragraph{Statistical Heterogeneity and Mixture Models.}The canonical challenge in federated learning is the divergence of client data distributions. Traditional methods mitigate this via proximal regularization (FedProx \citep{fedprox}), control variates (SCAFFOLD \citep{scaffold}),  meta-learning-based personalization (Per-FedAvg \citep{fallah2020perfedavg}; pFedMe \citep{pfedme}), model interpolation \citep{deng2020adaptive,mansour2020three}, personalized empirical risk minimization (PERM~\citep{deng2023distributed}). A related line of work explicitly models each client's data as a mixture of latent source distributions. FedEM \citep{fedem} jointly learns shared latent components and client-specific mixture weights, while FedSoft \citep{fedsoft} learns cluster and personalized models under soft client membership. These methods estimate the latent components as part of federated optimization. FedSEE studies a different recovery problem in which local optimization has already produced mixed client adaptations and the server must recover task-indexed experts from the resulting upload pool. Hard clustering methods such as IFCA \citep{ifca} instead assign each client to a single component. We evaluate this paradigm as IFCA+LoRA by pairing each cluster with its own dedicated adapter. Without task-indexed side information, latent components remain semantically unlabeled, and without sufficient boundary information even the unordered expert set may be non-identifiable (Theorems~\ref{thm:label} and~\ref{thm:vertex}). By leveraging task-labeled public data, FedSEE fixes the task correspondence and, under the anchor condition, recovers task-indexed experts with the guarantee of Theorem~\ref{thm:recovery}.

\paragraph{Federated LoRA Fine-Tuning.}Early federated PEFT frameworks such as FedIT \citep{zhang2024fedit} apply standard federated aggregation to LoRA parameters. Because separately averaging the two LoRA factors does not generally reproduce the average of their products, $\bar{B}\bar{A}\neq\overline{BA}$, subsequent methods focus on reducing this aggregation mismatch. FedEx-LoRA \citep{singhal2024fedexlora} corrects the residual aggregation error, FLoRA \citep{flora} stacks client updates along the rank dimension, FLoRA-NA \citep{nguyenflora-na} constructs surrogate aggregated factors, and FedRot-LoRA \citep{zhang2026fedrot} aligns equivalent low-rank factorizations before aggregation. These methods target an accurate global adapted model rather than the recovery of task-indexed experts. Other approaches exploit the asymmetric roles of the LoRA factors. FedSA-LoRA \citep{guo2025fedsalora} shares only the $A$ factor, while FFA-LoRA \citep{sun2024ffa} fixes one randomly initialized factor and optimizes only the other. FedSEE instead learns a shared projection $A^{\mathrm{global}}$ during initialization and then freezes it. Beyond reducing aggregation ambiguity, this common projection gives the theoretical model a shared feature space in which client optima in $B$-space become mixture-weighted combinations of task optima (Theorem~\ref{thm:mixture}).

\paragraph{Federated Mixture-of-Experts.}A complementary line of work preserves heterogeneity through multiple adapters or experts. MoLE \citep{wu2024mole} learns gating functions over a collection of independently trained LoRA experts. In federated settings, FedLEASE \citep{fedlease} clusters client representations to construct domain-specific experts and subsequently uses an adaptive mixture-of-experts mechanism to select among them. Related personalized approaches such as FedDPA \citep{feddpa} maintain global and local adapters and combine them through instance-dependent weighting. FedSEE differs from clustering-based expert construction in two respects. First, each mixed client upload may contribute fractionally to multiple recovered task experts through the extraction weights and the resulting contribution matrix $C$. Second, the recovered experts are optimized against task-labeled public risks, which assigns them explicit task identities rather than identities induced by client clusters. Resource-heterogeneous FedLoRA methods such as FlexLoRA \citep{bai2024flexlora} instead adapt LoRA rank to client capabilities and use SVD-based redistribution of the aggregated update. FedSEE addresses a different form of system heterogeneity by separating the experts available for inference from those a client can actively update through its capacity-aware active/passive partition.

\paragraph{Centralized Model and Adapter Composition.}A parallel non-federated literature studies combining independently trained task-specific models or adapters through weight arithmetic (Task Arithmetic \citep{ilharco2023editing}), conflict resolution (TIES-Merging \citep{ties}), orthogonalization (ZipLoRA \citep{ziplora}; DO-Merging \citep{zheng2025domerge}), or modular decomposition (LoRA-LEGO \citep{zhao2025lego}). LoRAHub \citep{huang2024lorahub} uses a small number of examples from an unseen task to determine composition weights over a collection of already trained LoRA modules. These methods solve a forward composition problem in which task-specific models or adapters are already available and the goal is to combine them. FedSEE instead addresses an inverse extraction problem. In the federated mixture setting, task-specific experts are not directly observed and the server receives adaptations that already mix information from multiple tasks. FedSEE uses task-labeled public data to recover and identify the latent task experts from this upload pool before learning how to route among them.

\section{Experimental Details and Additional Results}
\label{app:experiments}
This appendix provides the experimental details and additional results supporting Section~\ref{sec:experiments}. We first present controlled synthetic experiments that isolate the predictions of the theoretical analysis. We then give the complete specification of the natural-language experiments, followed by detailed benchmark results, the training-capacity analysis, component ablations, and client-mixture estimation results.

\subsection{Synthetic Validation of the Theory}
\label{app:synthetic}

This subsection provides controlled validation of the theoretical results in Section~\ref{sec:theory}. We construct a synthetic setting in which the assumptions can be enforced directly and examine the barycentric characterization, expert-recovery guarantees, identifiability limits, and routing predictions. These experiments isolate the behavior predicted by the theory before turning to the transformer experiments of Section~\ref{sec:experiments}.
 
\subsubsection{Synthetic Setup and Barycentric Validation}
\label{app:synthetic_setup}
We instantiate the LoRA model of Section~\ref{sec:FedSEE} directly: $\phi(x) = Ax$ with $A \in \mathbb{R}^{r \times d}$ frozen, $f_0(x) = W_0x$, and $\theta = B$. Data for task $m$ is generated as $y = W_0x + \theta_m^*\phi(x) + \zeta$ with $\zeta \sim \mathcal{N}(0, \sigma^2 I)$, so Assumption~\ref{ass:linear} holds by construction. Every task shares the input second moment $\Sigma = I$, hence $\Phi_m = AA^\top = \Phi$ for all $m$ and Assumption~\ref{ass:linear}(ii) holds exactly; $A$ is drawn with a prescribed singular spectrum so that $\kappa(\Phi)$ is set directly, and the experts $\{\theta_m^*\}$ are centered and rescaled to a prescribed $\mathrm{diam}_\mathrm{F}(\Theta)$. Unless noted we use $d = 60$, $r = 16$, $l = 12$, $M = 6$, $\kappa(\Phi) = 4$, $\mathrm{diam}_\mathrm{F}(\Theta) = 6$, $\sigma = 0.5$, $N = 12$ clients, and non-anchor mixtures drawn from $\mathrm{Dir}(0.5)$. For each task one anchor client is planted with $\alpha_{i_m,m} = 1-\varepsilon$ and $\alpha_{i_m,m'} = \varepsilon/(M-1)$, so that $\varepsilon$ is directly controlled; the failure of Assumption~\ref{ass:anchor} is imposed by drawing every client $\gamma$-interior instead. Clients solve their least-squares problem in closed form, and the server solves~\eqnref{eq:gen_extract} and~\eqnref{eq:gen_bary} as quadratic programs on the simplex by FISTA with exact simplex projection. On an adversarial instance admitting a non-vertex optimum the solver agrees with exact SLSQP to four decimals ($w = (0.085, 0.744, 0.172)$, objective $0.4404$ against $4.0400$ at the anchor), confirming that the vertex solutions reported below are not artifacts of the optimizer.
 
\paragraph{Barycentric convergence (Theorem~\ref{thm:mixture}).} Theorem~\ref{thm:mixture} predicts that the population client optimum equals the mixture-weighted barycenter of the task experts. Mean error $\|\hat\theta_i - \sum_m \alpha_{im}\theta_m^*\|_\mathrm{F}$ over
$12$ clients, against local sample size: $0.801$, $0.366$, $0.166$, $0.083$, $0.036$ at $n_i = 2\times10^2, 10^3, 5\times10^3, 2\times10^4, 10^5$. Successive ratios are 2.19, 2.21, 1.99, and 2.29, closely matching the $n^{-1/2}$ predictions 2.24, 2.24, 2.00, and 2.24.

\begin{table}[t]
\centering
\caption{Expert recovery error against anchor impurity.}
\label{tab:app_thm2}
\begin{tabular}{lcccccc}
\toprule
$\varepsilon$ & $0$ & $0.05$ & $0.10$ & $0.20$ & $0.30$ & $0.40$ \\
\midrule
exact uploads                 & $0.000$ & $0.228$ & $0.455$ & $0.911$ & $1.366$ & $1.821$ \\
finite, $n_i = 10^3$          & $0.167$ & $0.380$ & $0.596$ & $1.041$ & $1.574$ & $1.980$ \\
finite, $n_i = 1.6\times10^4$ & $0.044$ & $0.242$ & $0.483$ & $0.930$ & $1.387$ & $1.836$ \\
bound $\kappa(\Phi)^{1/2}\varepsilon\, \mathrm{diam}_\mathrm{F}(\Theta)$ & $0.000$ & $0.600$ & $1.200$ & $2.400$ & $3.600$ & $4.800$ \\
\bottomrule
\end{tabular}
\end{table}
\subsubsection{Expert Recovery and Finite-Sample Behavior}
\label{app:synthetic_recovery}

\begin{figure}[t]
\centering
\includegraphics[width=\linewidth]{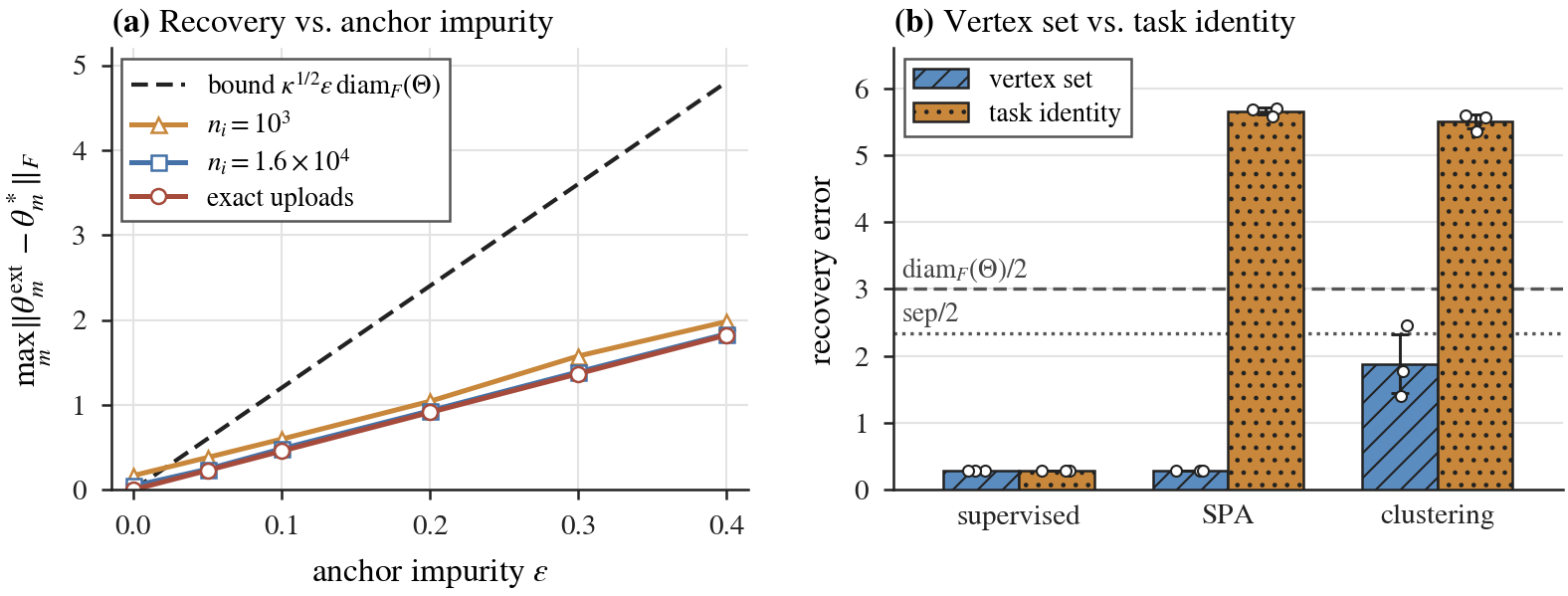}
\vspace{-15pt}
\caption{\textbf{Synthetic validation of expert recovery.}
(a) Expert-recovery error against anchor impurity, with the bound of Theorem~\ref{thm:recovery}.
(b) Vertex-set versus task-identity recovery; dashed and dotted lines mark
$\mathrm{diam}_F(\Theta)/2$ and $\mathrm{sep}/2$ (Theorem~\ref{thm:label}).
Bars show mean $\pm$ standard deviation over three seeds; circles are individual seeds.}
\label{fig:synthetic}
\end{figure}

\paragraph{Anchor impurity and finite-sample recovery (Theorem~\ref{thm:recovery} and Proposition~\ref{prop:finitesample}).} Theorem~\ref{thm:recovery} and
Proposition~\ref{prop:finitesample} predict complementary effects of anchor impurity and finite sampling. Figure~\ref{fig:synthetic}(a) and Table~\ref{tab:app_thm2} show that expert-recovery error is exactly linear in anchor impurity under exact uploads and vanishes at $\varepsilon=0$. The finite-sample floor decomposes as Proposition~\ref{prop:finitesample} predicts: holding $n_i = 2\times10^4$ and varying the public set, the error falls $0.087 \to 0.039$ as $n_m$ goes $10^2 \to 1.6\times10^4$; holding $n_m = 1.6\times10^4$ and varying local data, it falls $0.352 \to 0.043$ as $n_i$ goes $250 \to 1.6\times10^4$, halving for each fourfold increase.

\paragraph{Effect of the conditioning factor.}
The bound of Theorem~\ref{thm:recovery} is never violated, and its $\varepsilon$-dependence is tight, but the conditioning factor behaves in two distinct regimes. When each anchor's residual mass is spread evenly over the remaining tasks, program~\eqnref{eq:gen_extract} places weight $1.00$ on the anchor: the recovered expert \emph{is} the anchor upload, the error equals $\varepsilon\,\|\theta_m^* - \tfrac{1}{M-1}\sum_{m'\neq m}\theta_{m'}^*\|_\mathrm{F}$, and is therefore independent of $\Phi$ in this regime. Measured error is then invariant to $\kappa(\Phi)$ across $1$ to $1024$, and remains so when the experts are placed on eigendirections of $\Phi$ so that the Gram matrix induced in mixture coordinates has condition number equal to $\kappa(\Phi)$ exactly. The factor becomes active precisely when the pool admits a convex combination strictly better than any anchor, so that the optimum lies on a face rather than a vertex: with anchor impurity concentrated on a single high-curvature coordinate and complementary clients present, the minimizer has $\max_j \omega_{m,j} \in [0.39, 0.75]$ and the error rises monotonically with conditioning, $1.200 \to 1.894 \to 3.679 \to 4.194$ at $\kappa(\Phi) = 1, 16, 256, 4096$. At $\kappa(\Phi)=1$ the measured error $1.2000$ coincides with the bound $1.200$, so the constant is attained. Thus the $\varepsilon$-dependence is tight in this regime, while for larger $\kappa(\Phi)$ the bound remains conservative.

\subsubsection{Identifiability and Scaling Limits}
\label{app:synthetic_identifiability}
\paragraph{Task identity and vertex-set recovery (Theorems~\ref{thm:label} and~\ref{thm:vertex}).}
Figure~\ref{fig:synthetic}(b) and Table~\ref{tab:app_thm34} separate vertex-set recovery from task-identity recovery. Here $\mathrm{sep}/2 = 2.333$ and $\mathrm{diam}_\mathrm{F}(\Theta)/2 = 3.000$.

 \begin{figure}[t]
\centering
\includegraphics[width=\linewidth]{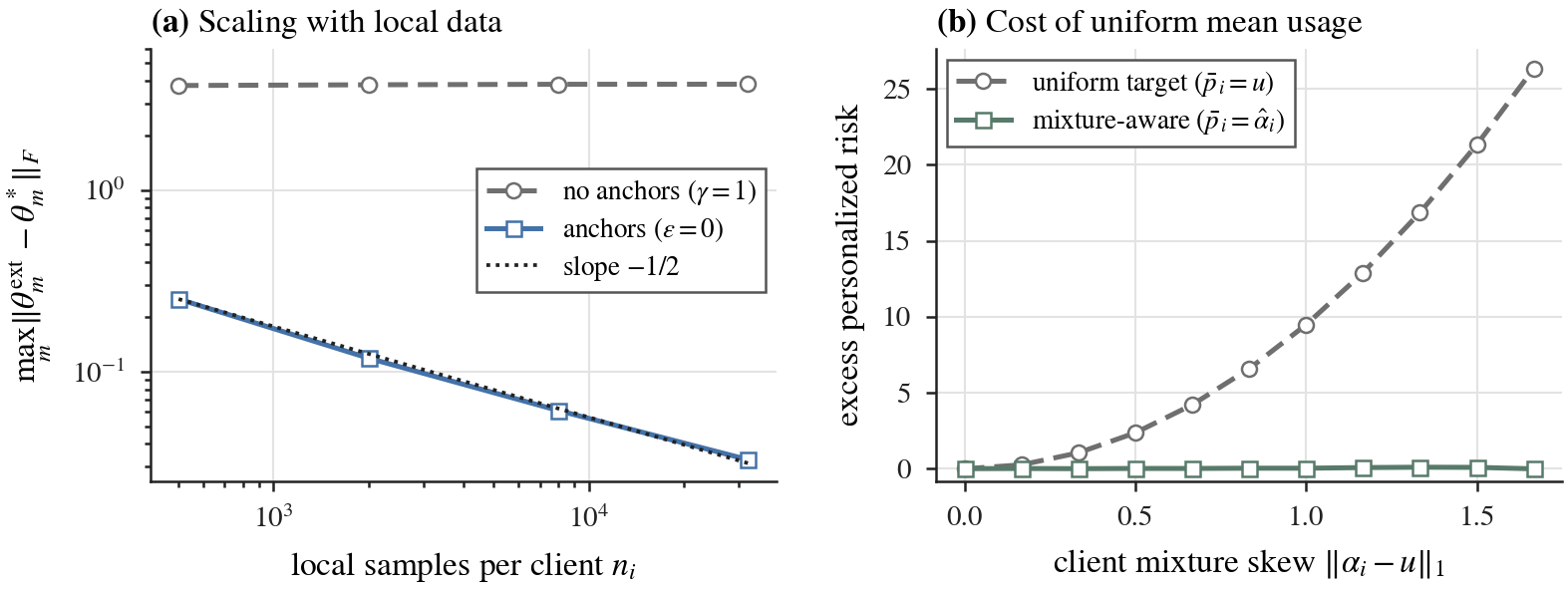}
\vspace{-15pt}
\caption{\textbf{Synthetic validation of identifiability limits and routing.}
(a) Recovery error against local data size with and without anchors (Theorem~\ref{thm:vertex}).
(b) Excess personalized risk of constant routers under uniform (load-balancing) versus
mixture-aware mean usage (Theorem~\ref{thm:routing_gap}).}
\label{fig:synthetic_appendix}
\end{figure}

\begin{table}[!htbp]
\centering
\caption{Recovery error with and without anchors, three seeds,
$N = 12$, $n_i = 4000$, $n_m = 1000$.}
\label{tab:app_thm34}
\begin{tabular}{lcccc}
\toprule
& \multicolumn{2}{c}{anchors, $\varepsilon = 0.05$} & \multicolumn{2}{c}{no anchors, $\gamma = 1$} \\
\cmidrule(lr){2-3}\cmidrule(lr){4-5}
Method & vertex set & task identity & vertex set & task identity \\
\midrule
supervised & $\mathbf{0.287}$ & $\mathbf{0.287}$ & $3.760$ & $3.760$ \\
SPA        & $\mathbf{0.287}$ & $5.658$ & $3.770$ & $3.818$ \\
clustering & $1.877$ & $5.507$ & $3.780$ & $3.815$ \\
\bottomrule
\end{tabular}
\end{table}

Increasing the federation size does not reduce the unanchored error floor. At $N = 20$ the no-anchor error is $3.754, 3.789, 3.813, 3.816$ as $n_i$ goes $500 \to 3.2\times10^4$, and remains nearly unchanged as $N$ grows from $12$ to $60$; the exact-anchor error decreases from $0.251$ to $0.033$ over the same range at the $n_i^{-1/2}$ rate (Figure~\ref{fig:synthetic_appendix}(a)).

\subsubsection{Mixture-Aware Routing}
\label{app:synthetic_routing}

\paragraph{Mean calibration and balanced routing (Theorem~\ref{thm:routing_gap} and
Corollary~\ref{cor:balanced_cost}).} To isolate the mean-calibration component of Theorem~\ref{thm:routing_gap}, we evaluate constant routing policies, for which $V_i(p_i)=0$. The resulting excess risk is available in closed form as $\sum_m \alpha_{im}\|\sum_k p_k\theta_k^* - \theta_m^*\|_\Phi^2$ minus its value at $p = \alpha_i$, so the comparison does not depend on an optimization procedure. Numerical minimization confirms the minimizer is $\alpha_i$ to three decimals. Sweeping a client from uniform toward a vertex, the excess risk of a balanced router is $0, 1.053, 4.212, 9.477, 16.847, 26.324$ --- exactly $26.32\,\tau^2$ in the interpolation parameter, and thus quadratic in $\|\alpha_i - u\|_1$ (Figure~\ref{fig:synthetic_appendix}(b)). Using the full pipeline instead, with experts from~\eqnref{eq:gen_extract} and mixtures from~\eqnref{eq:gen_bary}, over the $24$ non-anchor clients of four seeds: $\|\hat\alpha_i - \alpha_i\|_1 = 0.056$, excess risk $0.031$ when anchored to $\tilde c_i$ and $7.526$ under load balancing.
\subsection{Geometry and Routing Beyond the Exact Model}
\label{app:controlled_geometry}

The exact barycentric characterization in Section~\ref{sec:theory} is exact under the shared-feature squared-loss model. We next examine whether the same structure remains visible in a substantially different regime with a deep transformer, cross-entropy loss, and LoRA adaptation. We focus on two questions motivated directly by the theory. First, whether mixed-client updates retain a task-simplex organization when the exact barycentric relation no longer holds. Second, whether the client-level task mixture is sufficient for specialization, or whether mixed clients still require input-dependent routing.

\paragraph{Controlled setup.}
We use QNLI, SST-2, and WiC and construct nine clients at prescribed locations of the resulting three-task simplex, with three pure clients, three edge clients, and three interior clients. We use \texttt{distilroberta-base} with rank-$4$ LoRA adapters applied to the query and value projections. In an initialization round, each client trains both LoRA factors and the server averages the resulting $A$ matrices to obtain a shared $A^{\mathrm{global}}$. The averaged initialization update in $B$ is absorbed into the frozen base model, after which $A^{\mathrm{global}}$ remains fixed and all residual $B$ matrices are reinitialized to zero. We then train one pooled expert for each task and one local adapter for each client for exactly $240$ optimizer steps under the same learning-rate schedule. This common optimization clock makes their learned updates directly comparable without differences caused only by training duration. We repeat the experiment over three random seeds. For the geometry analysis, we compare the six value-projection $B$ blocks layer by layer and normalize distances within each layer by the maximum pairwise distance between the three task experts. The routing diagnostic subsequently freezes these task experts and trains only the input-dependent router.

\begin{figure}[t]
    \centering
    \includegraphics[width=0.8\linewidth]{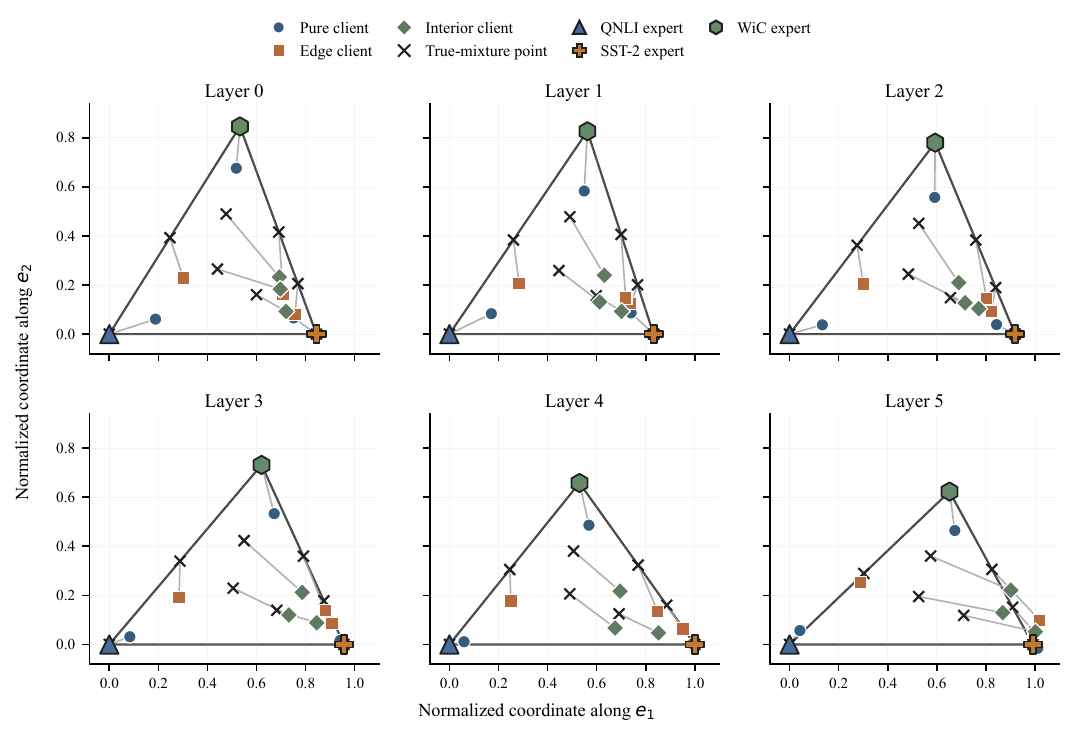}
    \caption{\textbf{Expert-plane geometry across layers.}
    Colored markers show projected client updates and crosses show the locations induced by their realized task mixtures, averaged over three seeds.}
    \label{fig:deep_expert_plane}
\end{figure}

\paragraph{Expert-plane geometry.}
We first examine how the learned client updates are organized relative to the task experts. For each layer and seed, the three task-expert updates define a two-dimensional affine plane in the LoRA parameter space. We construct an orthonormal basis of this plane by taking the QNLI-to-SST-2 direction as the first axis and the component toward WiC orthogonal to it as the second, and then project each client update onto this plane. Because the basis is orthonormal, the resulting expert triangle preserves the pairwise distances between the three task experts for that layer. To compare the geometry across seeds and layers, coordinates are normalized by the maximum pairwise expert distance within each seed and layer before averaging across seeds. The crosses in Figure~\ref{fig:deep_expert_plane} show the locations induced by the realized client mixtures, while the colored markers show the corresponding learned client projections.

Figure~\ref{fig:deep_expert_plane} reveals a consistent task-simplex organization throughout the adapted layers. Pure clients remain associated with their corresponding task vertices, edge clients follow the faces spanned by their active tasks, and interior clients occupy the interior of the expert triangle. Moreover, the relative organization of these client types remains visible despite changes in the shape of the expert triangle across layers. This persistence suggests that the task-mixture structure predicted by the theoretical analysis continues to organize the learned updates well beyond the exact shared-feature squared-loss setting.

\paragraph{Quantifying the simplex approximation.}
We next quantify how well the expert simplex explains the full client updates, including components that are not visible in the two-dimensional projections. For client $i$ and layer $\ell$, we define the normalized reconstruction residual as
\[
R_i^\ell(w) = \frac{ \left\| B_i^\ell - \sum_{m=1}^{3} w_m B_m^\ell \right\|_\mathrm{F} }{ D^\ell }, \qquad D^\ell = \max_{m \neq m'} \left\| B_m^\ell - B_{m'}^\ell \right\|_\mathrm{F} .
\]
Each row of Table~\ref{tab:deep_geometry} corresponds to a different choice of the reconstruction weights $w$. The true-mixture residual uses $w=\alpha_i$ and measures the discrepancy from the barycentric location specified by the realized client mixture. The best convex-hull fit minimizes the residual over $w\in\Delta^{2}$ and therefore gives the closest point in the expert simplex. The best affine fit minimizes the same residual subject only to $\sum_m w_m=1$, allowing the reconstruction to move anywhere in the affine plane spanned by the experts. As reference controls, the uniform mixture uses $w=(1/3,1/3,1/3)$, while the shuffled mixture permutes the entries of $\alpha_i$ across the three experts, preserving the mixture proportions while breaking their correspondence with task identity.

\begin{table}[t]
\centering
\caption{Normalized reconstruction residuals of learned client updates. Results are mean $\pm$ standard deviation over three seeds of the per-seed median across clients and layers.}
\label{tab:deep_geometry}
\setlength{\tabcolsep}{6pt}
\begin{tabular}{lc}
\toprule
\textbf{Reference} & \textbf{Normalized residual} \\
\midrule
Best affine fit & $0.444 \pm 0.029$ \\
Best convex-hull fit & $0.444 \pm 0.029$ \\
True mixture $\alpha_i$ & $0.493 \pm 0.023$ \\
Uniform mixture & $0.579 \pm 0.015$ \\
Shuffled mixture & $0.650 \pm 0.012$ \\
\bottomrule
\end{tabular}
\end{table}

The expert simplex provides a substantially better reconstruction than the uniform and shuffled controls. In particular, the shuffled-mixture residual shows that preserving the mixture proportions is not sufficient when those proportions are assigned to the wrong task experts. The nearly identical affine and convex-hull residuals provide an additional geometric indication that allowing the reconstruction to leave the convex region spanned by the experts offers essentially no improvement. Together with Figure~\ref{fig:deep_expert_plane}, these results show that the task simplex remains a meaningful organizing structure for the learned updates in the deep-model setting.

\begin{figure}[t]
    \centering
    \includegraphics[width=0.5\linewidth]
    {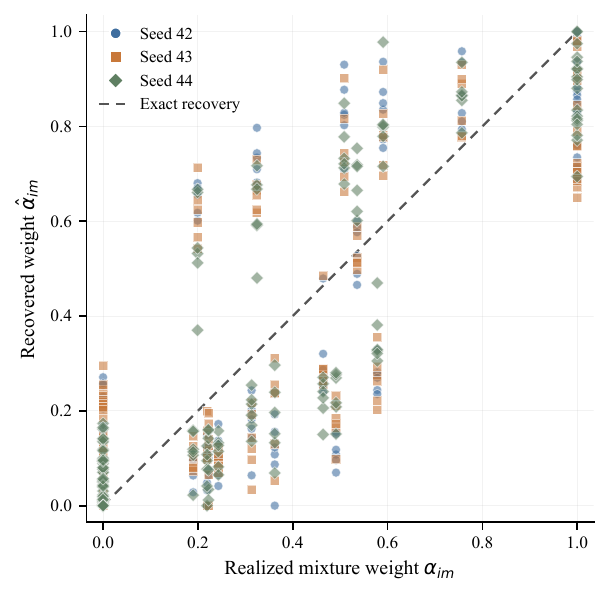}
    \caption{\textbf{Recovery of client mixture coordinates.}
    Recovered simplex coefficients versus realized task proportions across
    clients, tasks, layers, and seeds. The dashed line denotes exact recovery.}
    \label{fig:alpha_recovery}
\end{figure}

\paragraph{Mixture-coordinate recovery.}
The optimal weights of the convex-hull reconstruction also provide an estimate of the client's task composition. Specifically, for each client $i$ and layer
$\ell$, let
\[
\hat{\alpha}_i^\ell \in \arg\min_{w\in\Delta^2} \left\| B_i^\ell-\sum_{m=1}^{3} w_m B_m^\ell \right\|_\mathrm{F}.
\]
Figure~\ref{fig:alpha_recovery} compares each recovered coordinate
$\hat{\alpha}_{im}^\ell$ with the corresponding realized mixture weight
$\alpha_{im}$ across clients, tasks, layers, and seeds. The recovered
coordinates closely track the realized proportions, with a correlation of
$0.814 \pm 0.040$ and a mean absolute error of $0.157 \pm 0.017$ across
seeds. The support structure is particularly well preserved. For edge clients,
whose mixtures contain exactly two tasks, the recovered mass assigned to the
absent task is only $0.080 \pm 0.021$. Thus, the convex representation captures
not only the overall simplex organization observed above, but also meaningful
information about the client's underlying task composition.

\paragraph{Input-dependent routing for mixed clients.}
We next ask whether knowing the client-level task mixture is sufficient for specialization once the task experts are fixed. We compare three expert combinations. The uniform baseline uses the same weights $p_i(x)=u$ for every input, while the true-mixture baseline fixes $p_i(x)=\alpha_i$. Both are static combinations and therefore contain no input-level routing information. The input-dependent router instead produces $p_i(x)$ from the current example.

\begin{table}[t]
\centering
\caption{Validation task cross-entropy under static expert mixtures and input-dependent routing. Lower is better.}
\label{tab:deep_routing}
\setlength{\tabcolsep}{5pt}
\begin{tabular}{lccc}
\toprule
\textbf{Client type}
& \textbf{Uniform mixture}
& \textbf{True mixture}
& \textbf{Input-dependent} \\
\midrule
Pure     & $0.672$ & $0.487$ & $\mathbf{0.481}$ \\
Edge     & $0.684$ & $0.625$ & $\mathbf{0.497}$ \\
Interior & $0.691$ & $0.679$ & $\mathbf{0.554}$ \\
\bottomrule
\end{tabular}
\end{table}

The three settings reveal two complementary effects. First, the uniform combination performs worst across all client types, showing that treating the experts equally ignores useful information about the client's task composition. 
\begin{wrapfigure}{r}{0.55\textwidth}
\vspace{-5pt}
    \centering
    \includegraphics[width=1\linewidth]
    {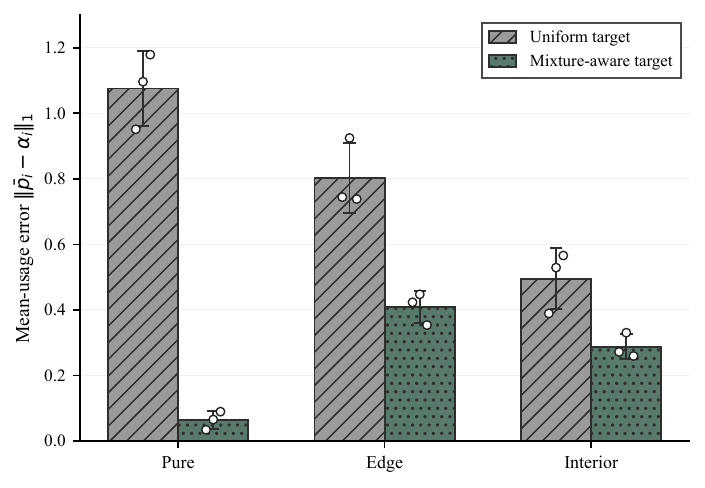}
    \caption{\textbf{Mixture-aware calibration of average expert usage.}
    Mean usage mismatch under uniform and mixture-aware KL targets. Bars show mean $\pm$ standard deviation over three seeds.}
    \label{fig:router_calibration}
\vspace{-30pt}
\end{wrapfigure}
Replacing the uniform weights with the realized mixture $\alpha_i$ substantially improves the pure and edge cases and also improves the interior clients. Second, for mixed clients, the true mixture remains a static client-level summary and therefore cannot determine which expert should process an individual example. Input-dependent routing further reduces the cross-entropy from $0.625$ to $0.497$ for edge clients and from $0.679$ to $0.554$ for interior clients. In contrast, for pure clients $\alpha_i$ is one-hot and already identifies the relevant expert, making the true-mixture combination and the input-dependent router nearly identical. These results separate the roles of client composition and input-level specialization: nonuniform client mixtures motivate mixture-aware expert usage, while mixed clients additionally require routing conditioned on the current input.

\paragraph{Mixture-aware router calibration.}
We finally isolate the role of the calibration target in the routing objective. We train two otherwise identical input-dependent routers and change only the target of the KL regularizer on the mean expert usage, using either the uniform distribution or the realized client mixture $\alpha_i$. This comparison therefore separates average-usage calibration from the input-level discrimination learned through the task loss.

Replacing the uniform target with $\alpha_i$ reduces the overall mean-usage mismatch
$\|\bar p_i-\alpha_i\|_1$ from $0.791$ to $0.253$. Figure~\ref{fig:router_calibration} shows that this improvement is consistent across pure, edge, and interior clients and is particularly pronounced for structured nonuniform mixtures. For pure clients, the mean probability assigned to the correct expert increases from $0.462$ to $0.968$. For edge clients, the mean probability assigned to the absent expert decreases from $0.289$ to $0.024$. In contrast, the validation task cross-entropy remains nearly unchanged at $0.509$ and $0.511$ for uniform and mixture-aware calibration, respectively. Unlike the full-training ablation in Appendix~B.6, this diagnostic freezes the task experts and optimizes only the router, isolating mean-usage calibration from its effect on expert training. These results indicate that the task loss supplies the input-dependent discrimination required for specialization, while the mixture-aware KL term aligns the router's average expert usage with the client composition.

Together, these experiments show that the task-simplex structure remains visible in the deep-model setting and that the client mixture continues to provide meaningful information about expert composition. At the same time, specialization of mixed clients requires input-dependent routing, while mixture-aware calibration controls the router's average expert usage.

\subsection{Experimental Setup}
\label{app:fedsee_hparams}
This subsection specifies the data construction, evaluation protocol, baseline implementations, model configurations, and optimization settings used in Section~\ref{sec:experiments}. Unless stated otherwise, these settings are held fixed within the corresponding experimental protocol.

\subsubsection{Tasks, Public Data, and Evaluation}

\paragraph{Task pool.}
We use six classification tasks from GLUE and SuperGLUE~\citep{wang-etal-2018-glue,wang2019superglue}: SST-2, QNLI, CoLA, QQP, WiC, and BoolQ. These tasks form a common pool of $M=6$ task distributions. The same task pool is used in the headline comparison, heterogeneity sweep, and ablations.

\paragraph{Public reference data.} For each task $m$, we reserve $1{,}000$ labeled examples as the public reference set $\mathcal D_m^s$. These examples are disjoint from all client training, validation, and test data and are never used for local model training or evaluation. FedSEE uses them only at the server to evaluate the task-indexed public losses required by the supervised extraction program~\eqnref{eq:omega_opt}. The evaluated baselines do not use these public examples. The public-set-size ablation varies this resource explicitly to quantify the sensitivity of FedSEE to the amount of server-side supervision.

\paragraph{Validation and personalized evaluation.}
Each client is assigned disjoint validation and test sets of $200$ examples each, with task compositions matched to its realized training mixture. The validation set is used only for hyperparameter and model selection. The test set is not used for local training, server-side extraction, router calibration, or model selection and is evaluated only after the experimental configuration has been fixed. For each random seed, all compared methods use the same client training, validation, and test partitions. Overall accuracy is the mean of the per-client test accuracies. We additionally report the mean test accuracy of the worst-served $50\%$ and $25\%$ of clients to measure personalization quality away from the population average.

\subsubsection{Compound-Client Construction}
\label{app:partition}

The client partitions are constructed to control both within-client and across-client heterogeneity while keeping the underlying set of tasks fixed. Each task contributes at most $5{,}000$ training examples to the allocation pool, or all available examples when its source training set is smaller. In particular, WiC contributes $4{,}428$ examples.

\paragraph{Pure-client endpoint.}
The condition denoted $\alpha_{\mathrm{Dir}}=0$ is not sampled from a Dirichlet distribution. It is a deterministic pure-client reference in which every client receives examples from exactly one task. In the $N=12$ heterogeneity sweep, two clients are assigned to each of the six tasks. Each task's available pool is divided between its assigned clients, with any integer-division remainder assigned to the final client. This realizes the vertex regime of the simplex geometry in Section~\ref{sec:theory}.

\paragraph{Mixed-client regimes.}
For $\alpha_{\mathrm{Dir}}\in\{0.1,0.5,1.0,50\}$, one task-proportion vector is drawn for each client from the symmetric distribution $\operatorname{Dirichlet}(\alpha_{\mathrm{Dir}}\mathbf 1)$. The $\alpha_{\mathrm{Dir}}=50$ condition is denoted \emph{uniform} in the figures and tables because its draws are strongly concentrated around equal task proportions. Smaller concentration parameters produce more peaked and more diverse client mixtures, whereas larger values move clients toward similar, near-uniform compositions.

Before converting a sampled mixture into integer data allocations, task weights below $0.05$ are set to zero and the remaining entries are renormalized. This preprocessing rule is applied identically to the client partitions used by all compared methods.

\paragraph{Supply-constrained allocation.}
Each client's mixture vector is scaled by a common nominal sample budget obtained from the total available task supply and the number of clients. This produces a requested number of examples from each task. For a task whose aggregate requests exceed its available supply, all client requests for that task are reduced proportionally until the total matches the available pool. When aggregate demand falls below supply, the unused remainder is left unallocated rather than redistributed. Consequently, the realized amount of training data varies with the sampled client mixtures. Table~\ref{tab:app_realized_volume} reports the realized pooled client training--validation allocation at each heterogeneity level.

\begin{table}[!htbp]
\centering
\caption{Realized pooled client training--validation allocation per task across the heterogeneity sweep. ``Uniform'' denotes $\alpha_{\mathrm{Dir}}=50$.}
\label{tab:app_realized_volume}
\renewcommand{\arraystretch}{1.2}
\begin{tabular}{l|ccccc}
\toprule
\textbf{Task} & $\alpha_{\mathrm{Dir}}\!=\!0$ & $0.1$ & $0.5$ & $1.0$ & \textbf{uniform} \\
\midrule
SST-2 & 5000 & 4997 & 4995 & 4993 & 4716 \\
QNLI  & 5000 & 4169 & 4996 & 4686 & 4728 \\
QQP   & 5000 & 2901 & 4439 & 3045 & 4987 \\
CoLA  & 5000 & 4998 & 2794 & 4665 & 4876 \\
WiC   & 4428 & 4427 & 4425 & 3618 & 4422 \\
BoolQ & 5000 & 4047 & 3639 & 4994 & 4994 \\
\bottomrule
\end{tabular}
\end{table}

\subsubsection{Baseline Implementations}

We compare against six federated LoRA methods spanning global aggregation, personalization, and clustering. For a given experiment, all methods use the same client partitions, pretrained backbone, communication-round budget, local-epoch budget, batch size, maximum sequence length, LoRA rank, and adapted transformer modules unless the method itself requires a different treatment of the LoRA factors. Method-specific aggregation and personalization follow the respective published procedures.

\paragraph{FedIT.}
FedIT~\citep{zhang2024fedit} applies standard federated averaging to both LoRA factors $A$ and $B$, producing a single global adapter shared by all clients.

\paragraph{FFA-LoRA.}
FFA-LoRA~\citep{sun2024ffa} keeps the LoRA projection $A$ fixed and aggregates the trainable $B$ updates, producing a single shared adapter without averaging two independently changing LoRA factors.

\paragraph{FedSA-LoRA.}
FedSA-LoRA~\citep{guo2025fedsalora} exploits the asymmetric roles of the two LoRA factors by aggregating $A$ while retaining a personalized client-side $B$ component.

\paragraph{FedDPA-T.}
FedDPA-T~\citep{feddpa} maintains global and personalized adapter components and combines them at inference through its test-time personalization procedure. We use the test-time variant throughout the reported comparisons.

\paragraph{IFCA+LoRA.}
We instantiate IFCA~\citep{ifca} with LoRA by maintaining a separate LoRA adapter for each cluster and iteratively assigning each client to a single cluster using the IFCA hard-assignment procedure. A client updates only the adapter associated with its assigned cluster.

\paragraph{FedLEASE.}
FedLEASE~\citep{fedlease} clusters client $B$ updates using cosine similarity, constructs one expert per cluster, and applies its published routing mechanism over the resulting expert pool.

The conceptual relationships among these methods and FedSEE are discussed separately in Appendix~\ref{app:related}. The description here concerns their experimental instantiation.

\subsubsection{FedSEE Configuration}

\paragraph{Initialization and shared projection.}
FedSEE begins with one initialization round. The server broadcasts a common randomly initialized LoRA pair $(A^{(0)},B^{(0)})$, and each client trains both factors for $E_{\mathrm{init}}$ local epochs. The server then forms $A^{\mathrm{global}}=N^{-1}\sum_i A_i$ and keeps this projection fixed for all subsequent task-specific experts. The complete initialization and communication procedure is given in Algorithm~\ref{alg:FedSEE}.

\paragraph{Expert extraction.}
The server solves~\eqnref{eq:omega_opt} once per communication round using $50$ coefficient-update steps at learning rate $0.1$. The coefficient matrix is projected onto the balanced-transport constraints using $20$ Sinkhorn iterations every $5$ updates, and the resulting contribution matrix is smoothed across rounds with EMA rate $\lambda=0.9$. We use an optimization regularization coefficient of $0.01$ during the coefficient updates. The extraction procedure itself is described in Appendix~\ref{app:fedsee_sinkhorn}.

\paragraph{Inference and training capacity.}
FedSEE uses $M=6$ recovered experts and makes all six available at inference, so $U_i=6$. The default configuration updates at most one expert per client, $L_i=1$, which we denote \textbf{FedSEE (L1)}. The full-capacity variant \textbf{FedSEE (L6)} sets $L_i=6$ while retaining the same inference pool. Experts with normalized client-specific contribution below $\tau=0.05$ are excluded from the active set, so $L_i$ remains an upper bound rather than a strict quota. The capacity construction is defined in Appendix~\ref{app:fedsee_capacity}, and its empirical effect is examined in Appendix~\ref{app:capacity_analysis}.

\paragraph{Router and shared expert.}
We set the router-calibration weight to $\beta_{\mathrm{KL}}=10^{-3}$ and use router learning rate $\eta_r=10\,\eta_B$. The shared LoRA expert is trained at learning rate $\eta_B$ and aggregated across clients using uniform FedAvg. Router parameterization and aggregation are described in Appendix~\ref{app:fedsee_router}.

Table~\ref{tab:hparams} summarizes the settings shared across methods and the additional FedSEE-specific settings. Following the NLU protocol of FedLEASE~\citep{fedlease}, all reported transformer runs use AdamW with a frozen pretrained backbone and a classification head that is frozen after initialization.

\begin{table}[!htbp]
\centering
\caption{Complete shared and FedSEE-specific hyperparameter
configuration.}
\label{tab:hparams}
\renewcommand{\arraystretch}{1.15}
\begin{tabular}{ll}
\toprule
\multicolumn{2}{l}{\emph{Shared across compared methods}} \\
\midrule
LoRA rank $r$ & $4$ \\
LoRA scaling $\alpha_{\mathrm{LoRA}}$ & $8$ \\
LoRA dropout & $0.1$ \\
Target modules & query and value projections in all self-attention layers \\
Backbone parameters & frozen; LoRA-MoE query/value projections use $W_0$ without bias \\
Optimizer & AdamW \\
Weight decay & $0.01$ \\
Gradient-norm clipping & $1.0$ \\
Batch size & $64$ \\
Maximum sequence length & $128$ tokens \\
Local epochs per round $E$ & $5$ \\
Communication rounds $T$ & $10$ \\
Personal validation-set size & $200$ examples per client \\
Personal test-set size & $200$ examples per client \\
\midrule
\multicolumn{2}{l}{\emph{FedSEE-specific}} \\
\midrule
Task experts $M$ & $6$ \\
Initialization epochs $E_{\mathrm{init}}$ & 5 \\
Inference capacity $U_i$ & $6$ \\
Training capacity $L_i$ & $1$ by default; $6$ in the L6 ablation \\
Extraction steps & $50$ per communication round \\
Extraction learning rate & $0.1$ \\
Extraction regularization coefficient & $0.01$ \\
Sinkhorn projection & $20$ iterations every $5$ extraction steps \\
Contribution EMA rate $\lambda$ & $0.9$ \\
Router calibration weight $\beta_{\mathrm{KL}}$ & $10^{-3}$ \\
Router learning rate & $\eta_r=10\,\eta_B$ \\
Active-set contribution floor $\tau$ & $0.05$ \\
Shared-expert learning rate & $\eta_B$ \\
Shared-expert aggregation & uniform FedAvg \\
Public reference set & $1{,}000$ examples per task for FedSEE \\
\bottomrule
\end{tabular}
\end{table}

\subsubsection{Headline Compound-Heterogeneity Protocol}

The headline comparison in Section~\ref{sec:main_results} uses $\alpha_{\mathrm{Dir}}=0.1$ and RoBERTa-large~\citep{liu2019roberta}, with $N=10$ fully participating clients. All methods train for $T=10$ communication rounds with $E=5$ local epochs per round. The supply-constrained allocation yields approximately $2{,}000$--$2{,}500$ training examples per client. For each method, we select the learning rate from $\{5\times10^{-4},10^{-3},5\times10^{-3}\}$ using the mean client validation accuracy on one fixed tuning seed. The selected learning rate is then fixed and evaluated on the held-out test sets using the tuning seed and two additional seeds, giving three reported runs per method. We report the mean and standard deviation across these runs.

\subsubsection{Heterogeneity-Sweep Protocol}

The heterogeneity sweep in Section~\ref{sec:alpha_sweep} uses \texttt{distilroberta-base}, $N=12$ clients, $T=10$ communication rounds, and $E=5$ local epochs. We evaluate five population regimes: the deterministic pure-client endpoint denoted $\alpha_{\mathrm{Dir}}=0$, Dirichlet concentrations $\alpha_{\mathrm{Dir}}\in\{0.1,0.5,1.0\}$, and the near-uniform $\alpha_{\mathrm{Dir}}=50$ endpoint.

All methods use the fixed expert learning rate $\eta_B=5\times10^{-4}$ and the three shared random seeds $\{42,43,44\}$. For a given seed, every method uses the same realized client partition. Between-method differences are therefore computed seed by seed. For each method, we report the mean paired difference from FedSEE and its sample standard deviation across the three matched seeds.

\paragraph{Local-only and centralized references.}
The sweep additionally includes two non-federated reference conditions. \textit{Local-only} trains each client independently on its own data and therefore measures performance without information sharing across clients. The \textit{centralized} reference pools the available training examples by task and trains one task-specific model on each pooled task dataset, which is then evaluated on examples from the corresponding task. These conditions are included as empirical references rather than as theoretical lower or upper bounds.

\subsubsection{Ablation Protocol}

Unless an ablation directly changes the corresponding component, the component ablations follow the headline compound-heterogeneity setting of Section~\ref{sec:main_results}: RoBERTa-large, $\alpha_{\mathrm{Dir}}=0.1$, $N=10$ clients, $T=10$ communication rounds, and $E=5$ local epochs. This keeps the remaining experimental conditions fixed while isolating each design choice.

The public-supervision ablation varies $|\mathcal D_m^s|\in\{100,250,500,1000\}$ examples per task, with $1{,}000$ as the default. The router-calibration ablation compares the default client-specific calibration with uniform mean-usage calibration and with removal of the first-moment calibration term. The expert-count ablation varies the task-labeled public reference sets: $M=3$ retains SST-2, QNLI, and CoLA, $M=6$ covers all client tasks, and $M=9$ additionally includes RTE, MRPC, and WNLI, which are absent from the clients; for WNLI, all 635 available training examples are used. The shared-expert ablation removes the always-active pair $(A^{\mathrm{s}},B^{\mathrm{s}})$ while leaving the routed expert mechanism unchanged.

Training capacity is evaluated separately with \texttt{distilroberta-base} across the heterogeneity sweep. We compare $L_i=1$ and $L_i=6$ while holding $U_i=6$ fixed. The corresponding results are reported in Appendix~\ref{app:capacity_analysis}.

\subsubsection{Computational Environment}

All reported transformer experiments were executed on a GPU cluster with NVIDIA H100 and A100 accelerators. Each experimental run used a single-process federated simulation loop.

\subsection{Detailed Results under Compound Heterogeneity}
\label{app:main_results}
Table~\ref{tab:main_results_per_task} gives the per-task breakdown for the RoBERTa-large experiment in Section~\ref{sec:main_results}. The main text focuses on client-level overall and tail metrics because these directly evaluate personalized performance under compound heterogeneity. The task-level results provide a complementary view of where the aggregate gains arise.  FedSEE obtains the highest accuracy on QNLI, WiC, and BoolQ and remains competitive on SST-2, CoLA, and QQP, while producing the strongest overall and client-tail performance.

\begin{table*}[!htbp]
\centering
\caption{Per-task and personalized accuracy (\%) under compound
heterogeneity ($\alpha_{\mathrm{Dir}}=0.1$) with RoBERTa-large. Results are reported as mean $\pm$ standard deviation over three seeds.}
\label{tab:main_results_per_task}
\renewcommand{\arraystretch}{1.2}
\setlength{\tabcolsep}{4pt}
\resizebox{\linewidth}{!}{%
\begin{tabular}{l|cccccc|ccc}
\toprule
\textbf{Method} & \textbf{SST-2} & \textbf{QNLI} & \textbf{CoLA} & \textbf{QQP} & \textbf{WiC} & \textbf{BoolQ} & \textbf{Overall} & \textbf{Worst-50\%} & \textbf{Worst-25\%} \\
\midrule

FedIT & $91.76 \pm 2.35$ & $86.16 \pm 1.77$ & $78.28 \pm 3.25$ & $70.81 \pm 25.73$ & $61.62 \pm 1.88$ & $72.41 \pm 5.30$ & $78.06 \pm 2.04$ & $69.35 \pm 5.17$ & $64.44 \pm 6.09$ \\

FFA-LoRA & $89.86 \pm 2.87$ & $77.67 \pm 7.35$ & $75.32 \pm 3.03$ & $71.80 \pm 15.43$ & $61.25 \pm 0.70$ & $62.90 \pm 2.80$ & $73.43 \pm 1.09$ & $64.64 \pm 2.27$ & $62.21 \pm 2.49$ \\

IFCA+LoRA & $\underline{93.04 \pm 1.13}$ & $83.69 \pm 6.18$ & $78.90 \pm 1.78$ & $\mathbf{86.02 \pm 1.95}$ & $61.85 \pm 3.45$ & $67.82 \pm 6.72$ & $79.42 \pm 0.99$ & $70.95 \pm 2.42$ & $67.37 \pm 1.49$ \\

FedSA-LoRA & $\mathbf{93.86 \pm 1.66}$ & $86.66 \pm 4.40$ & $\mathbf{81.00 \pm 1.36}$ & $79.98 \pm 4.39$ & $63.95 \pm 1.61$ & $71.97 \pm 9.16$ & $\underline{79.75 \pm 1.37}$ & $71.89 \pm 2.60$ & $66.78 \pm 2.83$ \\

FedDPA-T & $92.08 \pm 1.07$ & $\underline{87.64 \pm 3.53}$ & $77.82 \pm 2.08$ & $73.91 \pm 18.85$ & $64.32 \pm 2.89$ & $\underline{75.47 \pm 3.82}$ & $79.20 \pm 1.00$ & $71.40 \pm 3.74$ & $67.34 \pm 4.20$ \\

FedLEASE & $91.40 \pm 2.57$ & $83.21 \pm 1.66$ & $79.63 \pm 5.07$ & $71.66 \pm 14.99$ & $\underline{65.67 \pm 2.75}$ & $75.28 \pm 10.52$ & $79.52 \pm 1.80$ & $\underline{72.46 \pm 3.99}$ & $\underline{69.10 \pm 3.77}$ \\

\midrule

\textbf{FedSEE (L1, Ours)} & $92.54 \pm 1.39$ & $\mathbf{92.29 \pm 1.95}$ & $\underline{80.47 \pm 0.90}$ & $\underline{84.67 \pm 1.74}$ & $\mathbf{66.24 \pm 4.96}$ & $\mathbf{77.73 \pm 3.86}$ & $\mathbf{82.65 \pm 0.35}$ & $\mathbf{76.54 \pm 1.65}$ & $\mathbf{72.75 \pm 2.21}$ \\

\bottomrule
\end{tabular}%
}
\end{table*}

\subsection{Training-Capacity Analysis}
\label{app:capacity_analysis}

Appendix~\ref{app:fedsee_capacity} defines the active and passive expert sets controlled by the training capacity $L_i$. Here we examine how this choice affects empirical performance across different heterogeneity regimes.

\paragraph{Training-capacity tradeoff.}
We evaluate the effect of training capacity by fixing the expert pool at $M=6$ and the inference capacity at $U_i=6$, while varying only the maximum number of task-specific experts updated by each client. Specifically, we compare the default $L_i=1$ configuration with full-capacity training at $L_i=6$. Both variants retain all six experts during inference and therefore use the same inference capacity and routing architecture. Table~\ref{tab:capacity} reports the comparison across the full heterogeneity sweep.

FedSEE (L1) is strongest at the pure-client endpoint, improving over full-capacity training by $0.89$ points at $\alpha_{\mathrm{Dir}}=0$. As client mixtures become broader, L6 obtains small gains of $0.25$, $0.92$, $0.05$, and $0.31$ points at $\alpha_{\mathrm{Dir}}=0.1$, $0.5$, $1.0$, and the near-uniform regime, respectively. This behavior is consistent with the role of the client-specific contribution profile. For concentrated mixtures, local updates can be focused on the most relevant expert, whereas broader mixtures provide useful local signal for a larger portion of the expert pool. Importantly, the two configurations remain within one percentage point across every regime despite L1 updating at most one task-specific expert per client instead of six. We therefore use $L_i=1$ as the default configuration, which retains the full six-expert inference pool while substantially reducing task-specific training and communication cost.

\begin{table}[!htbp]
\centering
\caption{Training-capacity ablation: overall accuracy (\%) across the heterogeneity sweep.}
\label{tab:capacity}
\renewcommand{\arraystretch}{1.15}
\setlength{\tabcolsep}{4pt}
\begin{tabular}{l|ccccc}
\toprule
\textbf{Variant} & $\alpha_{\mathrm{Dir}}\!=\!0$ & $0.1$ & $0.5$ & $1.0$ & \textbf{uniform} \\
\midrule
FedSEE (L1) & $\mathbf{76.46 \pm 0.30}$ & $74.95 \pm 0.63$ & $71.28 \pm 1.76$ & $72.10 \pm 0.80$ & $69.97 \pm 1.37$ \\

FedSEE (L6) & $75.57 \pm 0.23$ & $\mathbf{75.20 \pm 0.40}$ & $\mathbf{72.20 \pm 1.12}$ & $\mathbf{72.15 \pm 0.93}$ & $\mathbf{70.28 \pm 0.41}$ \\
\bottomrule
\end{tabular}
\end{table}

\subsection{Component Ablations}
\label{app:ablations}

We isolate the main design choices of FedSEE under the headline compound-heterogeneity setting of Section~\ref{sec:main_results}. Unless an ablation directly modifies the corresponding component, we use RoBERTa-large, $\alpha_{\mathrm{Dir}}=0.1$, $N=10$ clients, $T=10$ communication rounds, and $E=5$ local epochs, following the protocol in Appendix~\ref{app:fedsee_hparams}. Each ablation changes one component of the default FedSEE configuration while keeping the remaining settings fixed. Table~\ref{tab:ablation} reports overall personalized accuracy and the change relative to the default configuration. The default row corresponds to the configuration used in Section~\ref{sec:main_results}, with $|\mathcal D_m^s|=1{,}000$, mixture-aware router calibration, six recovered experts, and the shared expert.

\begin{table}[!htbp]
\centering
\caption{Component ablations under the headline compound-heterogeneity setting. Results are reported as mean $\pm$ standard deviation over three seeds. $\Delta$ denotes the change in mean accuracy relative to the default configuration.}
\label{tab:ablation}
\renewcommand{\arraystretch}{1.15}
\setlength{\tabcolsep}{5pt}
\begin{tabular}{llcc}
\toprule
\textbf{Component} & \textbf{Setting} & \textbf{Overall (\%)} & $\Delta$ \\
\midrule

Default & FedSEE (L1) & $\mathbf{82.65 \pm 0.35}$ & -- \\

\midrule

\multirow{3}{*}{Public reference set}
& $|\mathcal D_m^s|=100$ & $81.27 \pm 0.74$ & $-1.38$ \\
& $|\mathcal D_m^s|=250$ & $81.92 \pm 0.90$ & $-0.73$ \\
& $|\mathcal D_m^s|=500$ & $82.49 \pm 0.39$ & $-0.16$ \\

\midrule

\multirow{2}{*}{Router calibration}
& uniform target & $80.85 \pm 0.42$ & $-1.80$ \\
& none & $81.35 \pm 0.75$ & $-1.30$ \\

\midrule

\multirow{2}{*}{Expert count}
& 3 experts & $82.01 \pm 0.59$ & $-0.64$ \\
& 9 experts & $82.31 \pm 0.54$ & $-0.34$ \\

\midrule

Shared expert & removed & $82.03 \pm 0.19$ & $-0.62$ \\

\bottomrule
\end{tabular}
\end{table}

\paragraph{Public supervision.}
We vary the task-labeled public reference set used for server-side extraction in~\eqnref{eq:omega_opt} from $100$ to the default $1{,}000$ examples per task. Reducing the public reference set produces a gradual decrease in accuracy, from $82.65\%$ with $1{,}000$ examples per task to $81.27\%$ with $100$. Importantly, the $100$-example setting uses only one tenth of the default public supervision, yet still achieves $81.27\%$ overall accuracy. This remains $1.52$ points above the strongest evaluated baseline, FedSA-LoRA at $79.75\%$, under the same headline setting reported in Table~\ref{tab:main_results_combined}. Increasing the reference set to $500$ examples per task recovers most of the default performance, reaching $82.49\%$, only $0.16$ points below the $1{,}000$-example setting. These results indicate that additional public supervision is beneficial, while the performance advantage over the evaluated baselines is retained even when the public reference set is reduced substantially.

\paragraph{Mixture-aware router calibration.}
We compare the default client-specific calibration target $\tilde c_i$ in~\eqnref{eq:client_obj} with a uniform mean-usage target and with removing the mean-usage calibration term entirely. Replacing the client-specific target with a uniform target reduces accuracy from $82.65\%$ to $80.85\%$, while removing the calibration term gives $81.35\%$. Mixture-aware calibration therefore improves over no calibration by $1.30$ points in this setting, whereas uniform calibration is $0.50$ points below using no calibration term. This ordering is consistent with the routing analysis in Section~\ref{sec:routing} and Proposition~\ref{prop:balanced_specialization}, where uniform mean expert usage can be mismatched to clients with non-uniform task mixtures. Empirically, the results suggest that the choice of calibration target matters rather than mean-usage calibration being beneficial by itself.

\paragraph{Expert count.}
The default configuration uses six public task anchors matching the six client tasks. With $M=3$, public supervision is retained only for SST-2, QNLI, and CoLA, while $M=9$ additionally introduces RTE, MRPC, and WNLI, which are absent from the clients. The six-expert default achieves $82.65\%$, compared with $82.01\%$ for $M=3$ and $82.31\%$ for $M=9$. The modest reductions indicate robustness to both incomplete and extraneous task coverage in the public reference set.

\paragraph{Shared expert.}
FedSEE includes the always-active shared LoRA pair $(A^{\mathrm{s}},B^{\mathrm{s}})$ in~\eqnref{eq:forward_shared} to capture adaptation that need not be assigned to a task-specific expert. Removing this component while leaving the routed expert mechanism unchanged reduces overall accuracy from $82.65\%$ to $82.03\%$, a decrease of $0.62$ percentage points. This modest reduction indicates that the shared expert provides a useful complementary adaptation pathway, allowing structure common across tasks to be represented without being absorbed into the routed expert pool. At the same time, performance remains close to the full model after its removal, suggesting that the task-specific routed experts account for the majority of the adaptation benefit rather than relying on the shared component.

\subsection{Client-Mixture Estimation}
\label{app:mixture_recovery_results}

Proposition~\ref{prop:mixture_est} characterizes client-mixture recovery through barycentric coordinates of the recovered task experts. Here we instead examine whether FedSEE's contribution-based mixture proxy tracks the known proportions used to construct the federated partitions.

\begin{figure}[t]
    \centering
    \includegraphics[width=\linewidth]{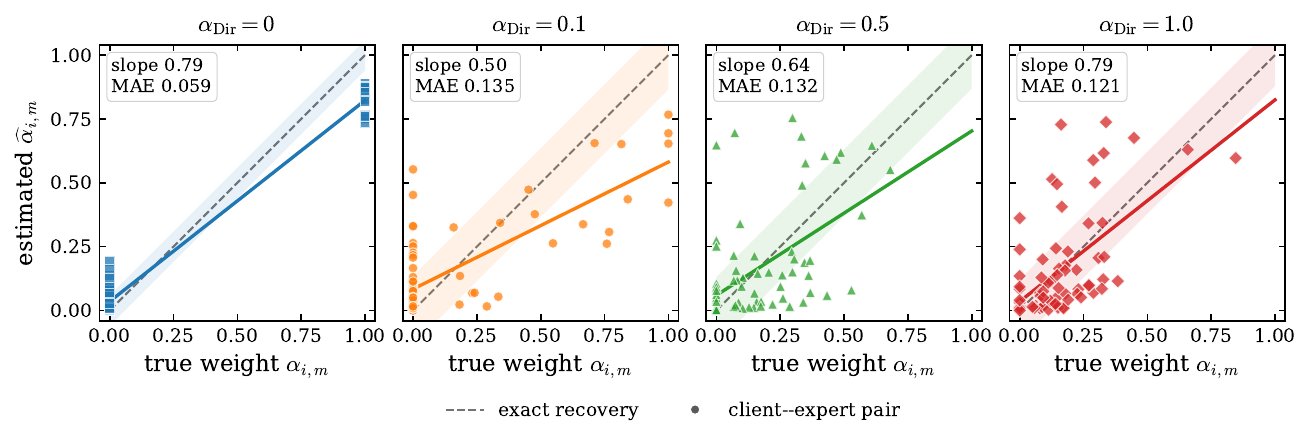}
    \vspace{-20pt}
    \caption{\textbf{Client-mixture estimation across heterogeneity regimes.}
    Contribution-based estimate $\hat c_{i,m}$ versus the ground-truth task proportion $\alpha_{i,m}$ for each client--task pair. The dashed diagonal indicates exact recovery, the solid line shows the least-squares fit, and each panel reports the fitted slope and mean absolute error (MAE).}
    \label{fig:mixture_recovery}
\end{figure}

For each client, we normalize its column of the smoothed contribution matrix $C$ to unit sum, yielding a contribution-based estimate $\hat c_i$, and average this estimate across the ten communication rounds. We compare these estimates with the task proportions used by the Dirichlet partitioner, which are not provided to the server during training. Figure~\ref{fig:mixture_recovery} reports all client--task pairs. At the pure-client endpoint $\alpha_{\mathrm{Dir}}=0$, the dominant task is correctly identified for every client, with a mean absolute error of $0.059$. In the mixed regimes, the corresponding MAEs are $0.135$, $0.132$, and $0.121$ for $\alpha_{\mathrm{Dir}}\in\{0.1,0.5,1.0\}$, respectively. Across all regimes, the contribution-based estimates remain positively associated with the underlying task proportions, showing that FedSEE's extraction contributions retain meaningful information about each client's mixture composition.

\section{FedSEE Algorithm and Implementation Details}
\label{app:fedsee}
\label{app:fedsee_algorithm}

Section~\ref{sec:FedSEE} introduces the main components of FedSEE. This appendix provides the corresponding algorithmic and implementation details, including the complete training procedure, the projection-level reduction underlying the LoRA instantiation, the joint balanced-extraction solver, the contribution-based capacity assignment, and the router parameterization and aggregation. Theoretical guarantees for these components are developed in Appendix~\ref{app:theory}, while the experimental configurations used in our evaluation are reported in Appendix~\ref{app:experiments}.

Algorithm~\ref{alg:FedSEE} states one full run of FedSEE. \textsc{Extract} solves the supervised extraction program jointly over all $M$ experts and returns the recovered experts together with the raw client-contribution matrix $C$, whose rows sum to one and whose client columns sum to $M/N$ under the balanced-transport constraint. \textsc{Assign} selects client $i$'s inference set from $c_{\cdot i}$, normalizes the retained contributions to obtain the unit-sum target $\tilde c_i$, and partitions the inference set into active and passive experts. \textsc{ClientUpdate} then runs $E$ local epochs on~\eqnref{eq:client_obj}, taking full gradients on the active blocks and stop-gradient on the passive blocks. Only active blocks are uploaded.

\begin{algorithm}[!htbp]
\caption{FedSEE: Federated Supervised Expert Extraction}
\label{alg:FedSEE}
\begin{algorithmic}[1]
\Require Public data $\{\mathcal{D}^s_m\}_{m=1}^M$, capacities $\{L_i,U_i\}_{i=1}^N$, communication rounds $T$
\Require Epochs $E_{\mathrm{init}}, E$; learning rates $\eta_B, \eta_r$; EMA rate $\lambda$;
         calibration weight $\beta_{\mathrm{KL}}$; contribution floor $\tau$
\Ensure Experts $\{B^{\mathrm{expert}}_m\}_{m=1}^M$, global projection $A^{\mathrm{global}}$,
        shared pair $(A^{\mathrm{s}},B^{\mathrm{s}})$, router $\psi$
\Statex
\State \textbf{Initialize:} broadcast LoRA pair $(A^{(0)},B^{(0)})$
\State Clients train for $E_{\mathrm{init}}$ epochs and return $\{(A_i, B_i)\}_{i=1}^N$
\State $A^{\mathrm{global}} \gets \frac{1}{N}\sum_{i=1}^N A_i$ \Comment{fixed thereafter}
\State Each client initializes $\psi_i^{(0)}$ and $(A_i^{\mathrm{s}},B_i^{\mathrm{s}})$ locally, with $A_i^{\mathrm{s}}\gets\text{Kaiming}()$ and $B_i^{\mathrm{s}}\gets\mathbf{0}$
\State $\bigl(\{B^{\mathrm{expert}}_m\}, C^{(1)}\bigr) \gets \textsc{Extract}\bigl(\{B_i\}_{i=1}^N, W_0\bigr)$ \Comment{initial experts}
\Statex
\For{$t = 1,\dots,T$}
    \State Server broadcasts $\{B^{\mathrm{expert}}_m\}_{m=1}^M$ and, for $t>1$, $(A^{\mathrm{s}},B^{\mathrm{s}})$ and $\psi^{(t-1)}$
    \For{client $i \in \{1,\dots,N\}$ \textbf{in parallel}}
        \State $(\mathcal{U}_i,\mathcal{A}_i,\mathcal{V}_i,\tilde{c}_i) \gets \textsc{Assign}\bigl(C^{(t)}_{\cdot i}, U_i, L_i, \tau\bigr)$
        \State $\bigl(\{B^i_m\}_{m\in\mathcal{A}_i}, (A^{\mathrm{s}}_i,B^{\mathrm{s}}_i), \psi_i\bigr)
               \gets \textsc{ClientUpdate}\bigl(i, \mathcal{U}_i, \mathcal{A}_i, \tilde{c}_i, t\bigr)$
    \EndFor
    \State $(A^{\mathrm{s}},B^{\mathrm{s}}) \gets \frac{1}{N}\sum_{i=1}^N (A^{\mathrm{s}}_i, B^{\mathrm{s}}_i)$,
           \quad $\psi^{(t)} \gets \frac{1}{N}\sum_{i=1}^N \psi_i$
    \State $\bigl(\{B^{\mathrm{expert}}_m\}, C^{\ast}\bigr) \gets
           \textsc{Extract}\bigl(\{B^i_m\}, W_0 + s B^{\mathrm{s}} A^{\mathrm{s}}\bigr)$ \Comment{update experts}
    \State $C^{(t+1)} \gets (1-\lambda)C^{(t)} + \lambda C^{\ast}$ \Comment{EMA of contributions}
\EndFor
\Statex
\Procedure{ClientUpdate}{$i, \mathcal{U}_i, \mathcal{A}_i, \tilde{c}_i, t$}
    \State $B_m \gets B^{\mathrm{expert}}_m$ for $m\in\mathcal{U}_i$; for $t>1$, $(A^{\mathrm{s}}_i,B^{\mathrm{s}}_i)\gets(A^{\mathrm{s}},B^{\mathrm{s}})$ and $\psi_i\gets\psi^{(t-1)}$
    \State Let $\Pi_m[\cdot]$ be the identity for $m\in\mathcal{A}_i$ and stop-gradient for $m\in\mathcal{V}_i$
    \For{$e = 1,\dots,E$}
        \State $h(x) \gets \Bigl(W_0 + s B^{\mathrm{s}}_i A^{\mathrm{s}}_i
               + s\sum_{m\in\mathcal{U}_i} p_{im}(x)\,\Pi_m[B_m]\, A^{\mathrm{global}}\Bigr)x$
        \State Compute $\mathcal{L}$ by \eqnref{eq:client_obj} 
        \State Update $\{B_m\}_{m\in\mathcal{A}_i}$, $(A^{\mathrm{s}}_i,B^{\mathrm{s}}_i)$ with $\eta_B$ and $\psi_i$ with $\eta_r$
    \EndFor
    \State \Return $\{B_m\}_{m\in\mathcal{A}_i}$, $(A^{\mathrm{s}}_i,B^{\mathrm{s}}_i)$, $\psi_i$
\EndProcedure
\Statex
\Procedure{Assign}{$c_{\cdot i}, U_i, L_i, \tau$}
    \State $\mathcal{U}_i \gets \operatorname{Top-}U_i\{c_{mi}\}_{m=1}^M$
    \State $\tilde c_{mi} \gets c_{mi}\big/\sum_{k\in\mathcal{U}_i}c_{ki}$ for $m\in\mathcal{U}_i$
    \State $\mathcal{A}_i \gets \operatorname{Top-}L_i \{m\in\mathcal{U}_i:\tilde c_{mi}\ge\tau\}$
    \State $\mathcal{V}_i \gets \mathcal{U}_i\setminus\mathcal{A}_i$
    \State \Return $(\mathcal{U}_i,\mathcal{A}_i,\mathcal{V}_i,\tilde c_i)$
\EndProcedure
\Statex
\Procedure{Extract}{$\mathcal{P}$, $\tilde{W}$} \Comment{$\mathcal{P}$: upload pool}
    \State Solve by projected gradient descent with Sinkhorn projection (Appendix~\ref{app:fedsee_sinkhorn}):
    \Statex \hspace{\algorithmicindent}
        $\boldsymbol{\Omega}^* \in \arg\min_{\boldsymbol{\Omega}} \sum_{m=1}^M
        \widehat{\mathcal{L}}_m^s\!\bigl(\textstyle\sum_{p\in\mathcal P}\omega_{mp}B_p;\, A^{\mathrm{global}},\tilde W\bigr)$
    \Statex \hspace{\algorithmicindent}
        \textbf{s.t.}\; $\omega_{mp}\ge0$,\; $\sum_{p\in\mathcal P}\omega_{mp}=1\;\forall m$,\;
        $\sum_{m=1}^M c_{mi}=M/N\;\forall i$, \; where $c_{mi}:= \sum_{p:\,\mathrm{client}(p)=i}\omega_{mp}$
    \State \Return $\bigl\{B^{\mathrm{expert}}_m = \sum_p\omega^*_{mp}B_p\bigr\}_{m=1}^M$ and $C^\ast=(c^\ast_{mi})$
\EndProcedure
\end{algorithmic}
\end{algorithm}
\subsection{Projection-Level Reduction}
\label{app:fedsee_projection}

\Eqnref{eq:forward_shared} describes one adapted linear projection, with the layer and module indices suppressed. Conditional on its incoming representation $x$, the routed LoRA contribution is linear in each $B_m$ because $A^{\mathrm{global}}$ is fixed. The always-on shared term can therefore be absorbed exactly into $\tilde f_0(x)=W_0x+sB^{\mathrm{s}}A^{\mathrm{s}}x$ at that projection. Hence the reduction to~\eqnref{eq:objective} is exact at the frozen-feature projection level.

The corresponding extension from this projection-level reduction to the full transformer is analyzed in Appendix~\ref{app:fedsee_reduction}.

\subsection{Joint Extraction with Balanced Transport}
\label{app:fedsee_sinkhorn}

\paragraph{Solver.}
The optimization problem in~\eqnref{eq:omega_opt} is solved by warm-started projected gradient descent. We differentiate the public loss through the reconstructed expert $B_m$ only and reach $\boldsymbol{\Omega}$ by the chain rule
\[
\frac{\partial\widehat{\mathcal{L}}_m^s} {\partial\omega_{m,(i,m')}} = \left\langle B^i_{m'}, \frac{\partial\widehat{\mathcal{L}}_m^s}{\partial B_m} \right\rangle .
\]
Gradients with respect to each reconstructed $B_m$ and each row of $\boldsymbol{\Omega}$ are clipped to norm $0.1$, and the coefficients are clipped to be nonnegative between Sinkhorn projections. Thus, each optimization step requires one forward and backward pass per task, independent of the size of the upload pool.

\paragraph{Balanced coupling.}
FedSEE couples the $M$ taskwise extraction problems through client-level contribution constraints to prevent different experts from repeatedly concentrating on the same clients. For task $m$ and client $i$, define
\[
c_{mi}:= \sum_{p:\mathrm{client}(p)=i}\omega_{mp}.
\]
Since the extraction weights for each task sum to one, $\sum_i c_{mi}=1$. The total contribution mass across the $M$ experts is therefore $M$. FedSEE distributes this mass uniformly across the $N$ clients by imposing
\[
\sum_m c_{mi}=\frac{M}{N},
\]
and enforces the resulting row and column marginals using Sinkhorn projection.

\paragraph{Client normalization.}
The balanced-transport constraint is imposed on the raw contribution matrix $C$ during expert extraction. After extraction, client $i$ normalizes the contributions of the experts retained in its inference set:
\[
\tilde c_{mi} = \frac{c_{mi}} {\sum_{k\in\mathcal U_i}c_{ki}}, \qquad m\in\mathcal U_i,
\]
so that $\sum_{m\in\mathcal U_i}\tilde c_{mi}=1$. When $\mathcal U_i=[M]$, this reduces to $\tilde c_{mi}=(N/M)c_{mi}$. This client-side normalization does not alter the balanced-transport constraints or the recovered experts. The effect of balanced coupling on expert recovery is analyzed in Proposition~\ref{prop:balanced_coupling} of Appendix~\ref{app:balanced_recovery}.


\subsection{The Mixture Proxy and Capacity Assignment}
\label{app:fedsee_capacity}

The barycentric analysis of Section~\ref{sec:theory} motivates a client-specific mixture profile over the recovered experts. FedSEE obtains this profile directly from the client contributions produced by the joint extraction procedure. These contributions provide the client-specific relevance signal used for both capacity assignment and router calibration.

\paragraph{Temporal smoothing.}
FedSEE does not use the raw contribution matrix from each extraction round directly. Instead, it maintains a temporally smoothed contribution matrix for capacity assignment and router calibration. The smoothing is applied only to the contribution matrix, not to the extraction weights $\boldsymbol{\Omega}$ or the recovered expert parameters. Let $C^{*,(t)}\in\mathbb{R}^{M\times N}$ denote the raw balanced contribution matrix returned by the extraction program at round $t$. The deployed update is
\begin{equation}
    C^{(t+1)} = (1-\lambda)C^{(t)} + \lambda C^{*,(t)}, \qquad \lambda\in(0,1). \label{eq:gen_smooth}
\end{equation}
The feasibility and tracking properties of this update are established in Proposition~\ref{prop:c_ema} of Appendix~\ref{app:c_ema_stability}.

For client $i$, the column $c_{\cdot i}^{(t)}=(c_{1i}^{(t)},\ldots,c_{Mi}^{(t)})$ summarizes its contribution across the recovered experts. After restricting this profile to the client's inference set, the retained entries are normalized to obtain $\tilde c_i$, which is used to rank experts for capacity assignment and to calibrate the router.

\paragraph{Active, passive, and inference sets.}
FedSEE separates training and inference capacity through two per-client budgets. The training capacity $L_i\in\{1,\ldots,M\}$ limits the number of experts updated by backpropagation, while the inference capacity $U_i\in\{L_i,\ldots,M\}$ limits the number of experts participating in the forward pass. Ranking the recovered experts by their client-specific contributions gives
\begin{equation}
    \mathcal{U}_i = \operatorname{Top-}U_i\{c_{mi}\}_{m}, \quad \tilde{c}_{mi} = \tfrac{c_{mi}}{\sum_{m'\in\mathcal{U}_i} c_{m'i}}, \quad \mathcal{A}_i = \operatorname{Top-}L_i \{\tilde{c}_{mi} : \tilde{c}_{mi}\ge\tau\}, \quad \mathcal{V}_i = \mathcal{U}_i\setminus\mathcal{A}_i , \label{eq:capacity}
\end{equation}
where $\mathcal U_i$, $\mathcal A_i$, and $\mathcal V_i$ denote the inference, active, and passive expert sets, respectively. The normalization is performed after applying the inference cap, so $\tilde c_i$ remains a unit-sum distribution over $\mathcal U_i$ when $U_i<M$. The threshold $\tau$ is then applied together with $L_i$ to determine the active set. Consequently, $L_i$ is an upper bound rather than a strict quota, and clients with concentrated contribution profiles may update fewer than $L_i$ experts.

Active experts receive gradients for both the expert parameters and the router. Passive experts remain in the forward pass, but their $B_m$ blocks are detached during backpropagation. Their routing probabilities therefore continue to receive task-loss gradients while their expert parameters remain unchanged on client $i$. Only active expert blocks are uploaded, so $\mathcal A_i$ also determines the expert-specific communication cost of the round. The effect of training capacity on empirical performance is evaluated in Appendix~\ref{app:capacity_analysis}.

\subsection{Router Parameterization and Aggregation}
\label{app:fedsee_router}

FedSEE uses a lightweight linear router at each routed layer. For an input token representation $x$, the router produces expert logits through $W_r^\ell x+b_r^\ell$. Experts outside the client-specific inference set $\mathcal U_i$ are masked before applying the softmax, yielding routing probabilities over the experts available to client $i$. The KL term in~\eqnref{eq:client_obj} is applied at each routed layer and averaged over $\mathcal P$. Its mean-routing statistic $\bar p_i^\ell$ is estimated on each minibatch by averaging the routing probabilities over its examples and token positions. The router is optimized with a separate learning rate $\eta_r$.

After each communication round, the server averages the router parameters across participating clients and broadcasts the resulting shared router for the next round. Each client retains its own inference set $\mathcal U_i$ and mixture-calibration target $\tilde c_i$, so the shared router is trained under different client-specific masks and calibration targets across the federation. The theoretical justification for using a shared routing function is given in Appendix~\ref{app:routing}.


\section{Theoretical Results and Proofs}
\label{app:theory}
This appendix provides the complete proofs and supporting theoretical results for the framework and its FedSEE instantiation. We first develop the exact squared-loss analysis, including the barycentric geometry, expert recovery, identifiability limits, mixture estimation, and routing guarantees. We then extend the framework beyond the exact quadratic setting and analyze the additional approximations introduced by the FedSEE instantiation.

Appendix~\ref{app:geometry} develops the barycentric geometry and the mixing gap. Appendix~\ref{app:recovery} establishes the population and finite-sample recovery guarantees, Appendix~\ref{app:identifiability} gives the identifiability limits, Appendix~\ref{app:mixture_estimation} studies mixture estimation and static reconstruction, and Appendix~\ref{app:routing} analyzes task-aligned routing. Appendix~\ref{app:general_loss} extends these results beyond the squared-loss setting, while Appendix~\ref{app:instantiations} discusses additional model-class instantiations. Finally, Appendices~\ref{app:fedsee_reduction}, \ref{app:balanced_recovery}, and~\ref{app:c_ema_stability} analyze the deep-model approximation, balanced expert extraction, and temporal smoothing used by FedSEE.

\paragraph{Notation.}
Recall that $\|C\|_\Phi^2:=\operatorname{tr}(C\Phi C^\top)$. Since $\Phi\succ0$,
\[
\sigma_{\min}(\Phi)\|C\|_{\mathrm F}^2 \leq \|C\|_\Phi^2 \leq \sigma_{\max}(\Phi)\|C\|_{\mathrm F}^2.
\]
Hence conversion between the $\Phi$-weighted and Frobenius norms introduces at most the factor $\kappa(\Phi)^{1/2}$, where $\kappa(\Phi)=\sigma_{\max}(\Phi)/\sigma_{\min}(\Phi)$. We use $[K]:=\{1,\ldots,K\}$ for positive integers $K$.

\subsection{Barycentric Geometry and the Mixing Gap}
\label{app:geometry}

\paragraph{Proof of Theorem~\ref{thm:mixture} (Mixture recovery in \texorpdfstring{$\theta$}{theta}-space).}

\textbf{Step 1. Specialize the mixture-risk decomposition.} By~\eqnref{eq:risk_decomp}, the client population risk decomposes as
\[
\mathcal{L}_i(\theta) = \sum_{m=1}^{M}\alpha_{im}\mathcal{L}_m(\theta).
\]
Under the squared loss used in Section~\ref{sec:theory}, this becomes
\begin{equation}
    \mathcal{L}_i(\theta) = \sum_{m=1}^{M} \alpha_{im} \mathbb{E}_{(x,y)\sim\mathcal{P}_m} \left[ \|y-f_0(x)-\theta\phi(x)\|_2^2 \right]. \label{eq:loss_decomp}
\end{equation}

\textbf{Step 2. Apply the task-wise realizability model.} By Assumption~\ref{ass:linear}(i), under $\mathcal{P}_m$,
\[
y-f_0(x) = \theta_m^*\phi(x)+\zeta_m.
\]
Substituting into~\eqnref{eq:loss_decomp} gives
\begin{equation}
    \mathcal{L}_i(\theta) = \sum_{m=1}^{M}\alpha_{im} \mathbb{E}_{\mathcal{P}_m} \left[ \|(\theta_m^*-\theta)\phi(x)+\zeta_m\|_2^2 \right]. \label{eq:loss_realizable}
\end{equation}

Since $\mathbb{E}[\zeta_m \mid x] = 0$, the cross term $2\mathbb{E}[\langle(\theta_m^*-\theta)\phi(x),\,\zeta_m\rangle]$ vanishes. Expanding the remaining square gives
\begin{equation}
    \mathcal{L}_i(\theta) = \sum_{m=1}^M \alpha_{im}\Bigl[ \mathbb{E}_{\mathcal{P}_m}\bigl[ \|(\theta_m^* - \theta)\phi(x)\|^2\bigr] + \sigma_m^2 l\Bigr]. \label{eq:expanded}
\end{equation}

\textbf{Step 3. Evaluate the quadratic term using the feature covariance.} Using $\|Cv\|^2 = \mathrm{tr}(Cvv^\top C^\top)$, linearity of trace and expectation, and Assumption~\ref{ass:linear}(ii) ($\Phi_m = \Phi$), we obtain
\begin{align}
    \mathbb{E}_{\mathcal{P}_m}\bigl[ \|(\theta - \theta_m^*)\phi(x)\|^2\bigr] &= \mathrm{tr}\bigl((\theta-\theta_m^*) \mathbb{E}_{\mathcal{P}_m}[\phi\phi^\top] (\theta-\theta_m^*)^\top\bigr) \notag\\
    &= \mathrm{tr}\bigl((\theta-\theta_m^*)\Phi(\theta-\theta_m^*)^\top\bigr) = \|\theta - \theta_m^*\|_\Phi^2,
\end{align}
where $\Phi = \mathbb{E}[\phi\phi^\top]$. Denoting $\Lambda_i = \sum_m \alpha_{im}\sigma_m^2 l$ as the constant term:
\begin{equation}
    \mathcal{L}_i(\theta) = \sum_{m=1}^M \alpha_{im}\|\theta - \theta_m^*\|_\Phi^2 + \Lambda_i. \label{eq:loss_quadratic}
\end{equation}

\textbf{Step 4. Solve the first-order optimality condition.} Using $\nabla_\theta\|\theta - C\|_\Phi^2 = 2(\theta-C)\Phi$, we obtain
\begin{equation}
    \nabla_\theta\mathcal{L}_i(\theta) = 2\sum_{m=1}^M\alpha_{im}(\theta - \theta_m^*)\Phi = 2\Bigl(\theta - \sum_{m=1}^M\alpha_{im}\theta_m^*\Bigr)\Phi.
\end{equation}
Setting to zero and right-multiplying by $\Phi^{-1}$, which exists by Assumption~\ref{ass:linear}(iii), gives
\begin{equation}
    \theta_i^* = \sum_{m=1}^M \alpha_{im} \theta_m^*.
\end{equation}

\textbf{Step 5. Uniqueness and global minimality.} The Hessian of $\mathcal{L}_i$ with respect to $\mathrm{vec}(\theta)$ is $2(\Phi \otimes I_l) \succ 0$, since $\Phi \succ 0$ by Assumption~\ref{ass:linear}(iii). Hence $\mathcal{L}_i$ is strictly convex in $\theta$, and the stationary point above is the unique global minimizer. \qed

\begin{remark}[Heterogeneous input covariance]
\label{rem:heterogeneous}
Without Assumption~\ref{ass:linear}(ii), the gradient in Step 4 yields a covariance-weighted average rather than a pure mixture average:
\begin{equation}
    \theta_i^* = \Bigl(\sum_m \alpha_{im}\theta_m^*\Phi_m\Bigr) \Bigl(\sum_m\alpha_{im}\Phi_m\Bigr)^{-1}, \quad \Phi_m := \mathbb{E}_{\mathcal{P}_m}[\phi\phi^\top].
\end{equation}
The bias relative to the true mixture $\bar{\theta}_i^{\mathrm{mix}} := \sum_m\alpha_{im}\theta_m^*$ admits a clean closed form. Writing $\bar{\Phi}_i := \sum_m \alpha_{im}\Phi_m$ for the mixture-averaged feature covariance and using $\sum_m \alpha_{im}(\Phi_m - \bar{\Phi}_i) = 0$, the optimality condition gives
\begin{equation}
    \theta_i^* - \bar{\theta}_i^{\mathrm{mix}} = \Bigl(\sum_m \alpha_{im}(\theta_m^* - \bar{\theta}_i^{\mathrm{mix}}) (\Phi_m - \bar{\Phi}_i)\Bigr)\,\bar{\Phi}_i^{-1},
\end{equation}
so that, bounding $\|\theta_m^* - \bar{\theta}_i^{\mathrm{mix}}\|_\mathrm{F} \leq \mathrm{diam}_\mathrm{F}(\Theta)$ and $\sigma_{\min}(\bar{\Phi}_i) \geq \min_m \sigma_{\min}(\Phi_m)$ (Weyl's inequality),
\begin{equation}
    \|\bar{\theta}_i^{\mathrm{mix}} - \theta_i^*\|_\mathrm{F} \;\leq\; \rho_\Phi \cdot \mathrm{diam}_\mathrm{F}(\Theta), \qquad \rho_\Phi \;:=\; \frac{\max_{m,m'}\|\Phi_m - \Phi_{m'}\|_2} {\min_m \sigma_{\min}(\Phi_m)}.
\end{equation}
The controlling quantity is the \emph{dispersion} of the per-task feature covariances, not the conditioning of any single one: $\rho_\Phi$ depends on how much the $\Phi_m$ differ from one another, and vanishes exactly when they coincide.  As the task feature covariances align, $\rho_\Phi\to0$ and Theorem~\ref{thm:mixture} is recovered in the limit. In particular, when all $\Phi_m$ are equal the bias is exactly zero even if the shared $\Phi$ is ill-conditioned --- a regime the worst-case condition number $\max_m\kappa(\Phi_m)$ would have spuriously penalized.
\end{remark}

\begin{proposition}[Simplex faces and mixing-gap geometry]
\label{prop:simplex_geometry}
Suppose the task experts are affinely independent, equivalently $h>0$, and let $S_i:=\{m:\alpha_{im}>0\}$. Then
\[
\theta_i^* \in \operatorname{relint} \operatorname{conv}\{\theta_m^*:m\in S_i\}.
\]
Hence $\theta_i^*$ lies in the relative interior of the full simplex iff $\alpha_{im}>0$ for every $m$, and is a vertex iff client $i$ is pure.

Moreover, the mixing gap satisfies
\begin{equation}
G_i = \frac12 \sum_{m,k} \alpha_{im}\alpha_{ik} \|\theta_m^*-\theta_k^*\|_\Phi^2, \label{eq:mixing_pairwise}
\end{equation}
and consequently
\begin{equation}
\frac{\sigma_{\min}(\Phi)\mathrm{sep}^2}{2} \left(1-\|\alpha_i\|_2^2\right) \le G_i \le \frac{\sigma_{\max}(\Phi) \mathrm{diam}_{\mathrm F}(\Theta)^2}{2} \left(1-\|\alpha_i\|_2^2\right). \label{eq:mixing_impurity_bound}
\end{equation}
Thus $G_i=0$ iff client $i$ is pure.
\end{proposition}

\paragraph{Proof of Proposition~\ref{prop:simplex_geometry}.}

Since $h>0$, the task experts are affinely independent and their convex hull is an $(M-1)$-simplex. By Theorem~\ref{thm:mixture},
\[
\theta_i^*=\sum_m\alpha_{im}\theta_m^*.
\]
Barycentric coordinates in a non-degenerate simplex are unique. A point lies in the relative interior of the face generated by a set of vertices exactly when its coefficients on those vertices are strictly positive and all remaining coefficients vanish. The face, full-simplex, and vertex claims therefore follow.

For the mixing gap, let $\bar\theta_i:=\sum_m\alpha_{im}\theta_m^*=\theta_i^*$. The weighted variance identity under the inner product induced by $\Phi$ yields
\begin{align*}
G_i &= \sum_m\alpha_{im} \|\theta_m^*-\bar\theta_i\|_\Phi^2\\
&= \frac12 \sum_{m,k} \alpha_{im}\alpha_{ik} \|\theta_m^*-\theta_k^*\|_\Phi^2.
\end{align*}
For $m\neq k$,
\[
\sigma_{\min}(\Phi)\mathrm{sep}^2 \le \|\theta_m^*-\theta_k^*\|_\Phi^2 \le \sigma_{\max}(\Phi) \mathrm{diam}_{\mathrm F}(\Theta)^2.
\]
Finally,
\[
\sum_{m\neq k}\alpha_{im}\alpha_{ik} = 1-\sum_m\alpha_{im}^2 = 1-\|\alpha_i\|_2^2,
\]
which gives~\eqnref{eq:mixing_impurity_bound}. Because $\mathrm{sep}>0$, equality $G_i=0$ holds exactly when $\|\alpha_i\|_2^2=1$, i.e., when $\alpha_i$ is a simplex vertex. \qed

Proposition~\ref{prop:simplex_geometry} shows that the gap between static adaptation and task-specific specialization is determined by the client's mixture structure and vanishes exactly for pure clients. We next study when the underlying task experts can be recovered from mixed client uploads.

\subsection{Supervised Expert Recovery and Finite-Sample Guarantees}
\label{app:recovery}

\paragraph{Proof of Theorem~\ref{thm:recovery} (Population expert recovery).}

\textbf{Step 1. Bound the anchor client's distance from the task expert.} By Theorem~\ref{thm:mixture}, anchor client $i_m$ satisfies:
\begin{equation}
    \theta_{i_m}^* - \theta_m^* = \sum_{m'\neq m}\alpha_{i_m,m'}(\theta_{m'}^* - \theta_m^*).
\end{equation}
Taking Frobenius norms and applying the triangle inequality:
\begin{equation}
    \|\theta_{i_m}^* - \theta_m^*\|_\mathrm{F} \leq \underbrace{\sum_{m'\neq m}\alpha_{i_m,m'}}_{ =\,1-\alpha_{i_m,m}\,\leq\,\varepsilon} \cdot\mathrm{diam}_\mathrm{F}(\Theta) \leq \varepsilon\cdot\mathrm{diam}_\mathrm{F}(\Theta).
\end{equation}

\textbf{Step 2. Translate to the $\Phi$-norm.} Using $\|C\|_\Phi \leq \sigma_{\max}(\Phi)^{1/2}\|C\|_\mathrm{F}$:
\begin{equation}
    \|\theta_{i_m}^* - \theta_m^*\|_\Phi \leq \sigma_{\max}(\Phi)^{1/2} \cdot\varepsilon\cdot\mathrm{diam}_\mathrm{F}(\Theta).
\end{equation}

\textbf{Step 3. Apply the projection bound.} The server minimizes $\|\theta - \theta_m^*\|_\Phi^2$ over $\mathrm{conv}(\{\theta_i^*\})$. Since $\theta_{i_m}^* \in \mathrm{conv}(\{\theta_i^*\})$ is feasible and $\theta_m^{\mathrm{ext}}$ is the minimizer:
\begin{equation}
    \|\theta_m^{\mathrm{ext}} - \theta_m^*\|_\Phi \leq \|\theta_{i_m}^* - \theta_m^*\|_\Phi \leq \sigma_{\max}(\Phi)^{1/2} \cdot\varepsilon\cdot\mathrm{diam}_\mathrm{F}(\Theta).
\end{equation}

\textbf{Step 4. Convert back to Frobenius norm.} Using $\|C\|_\mathrm{F} \leq \sigma_{\min}(\Phi)^{-1/2}\|C\|_\Phi$:
\begin{equation}
    \|\theta_m^{\mathrm{ext}} - \theta_m^*\|_\mathrm{F} \leq \frac{\sigma_{\max}(\Phi)^{1/2}}{\sigma_{\min}(\Phi)^{1/2}} \cdot\varepsilon\cdot\mathrm{diam}_\mathrm{F}(\Theta) = \kappa(\Phi)^{1/2}\cdot\varepsilon\cdot\mathrm{diam}_\mathrm{F}(\Theta).
\end{equation}
\qed

\begin{corollary}[Exact recovery with pure clients]
\label{cor:exact}
If $\varepsilon = 0$, then $\theta_m^{\mathrm{ext}} = \theta_m^*$ for all $m \in [M]$.
\end{corollary}
 
$\varepsilon = 0$ implies $\theta_{i_m}^* = \theta_m^*$ by Theorem~\ref{thm:mixture}. Since $\theta_m^*$ is feasible, the projection objective achieves value $\|\theta_m^* - \theta_m^*\|_\Phi^2 = 0$, uniquely attained at $\theta_m^{\mathrm{ext}} = \theta_m^*$.

In the finite-sample setting, clients upload estimated parameters $\hat{\theta}_i$ rather than population optima $\theta_i^*$, and the server evaluates empirical task losses on finite public reference sets. The following boundedness and tail conditions provide the concentration required for the finite-sample recovery analysis.
 
\begin{assumption}[Boundedness]
\label{ass:bounded}
The feature map is bounded, $\|\phi(x)\|_2 \leq Q$ almost surely, and the uploaded parameters satisfy $\|\hat{\theta}_i\|_\mathrm{F} \leq \theta_{\max}$ for all $i \in [N]$.
\end{assumption}

\begin{assumption}[Finite-Sample Public Data]
\label{ass:public_tail}
In addition to Assumption~\ref{ass:public}, each public set $\mathcal D_m^s$ is independent of the client-side data used to construct the uploaded parameters. Moreover, conditional on $x$, the task noise is $\sigma_m$-sub-Gaussian: for every $v\in\mathbb{R}^l$ and $t\in\mathbb{R}$,
\[
\mathbb{E}\!\left[ \exp\!\left(t\,v^\top\zeta_m\right)\mid x \right] \leq \exp\!\left( \frac{t^2\sigma_m^2\|v\|_2^2}{2} \right).
\]
\end{assumption}

\begin{proposition}[Finite-sample expert recovery]
\label{prop:finitesample}
Under Assumptions~\ref{ass:public}--\ref{ass:anchor}, \ref{ass:bounded}, and~\ref{ass:public_tail}, suppose the upload pool is fixed independently of the public sample used for extraction and 
\[
\|\hat{\theta}_i-\theta_i^*\|_{\mathrm F} \leq \delta_{\mathrm{loc}} \qquad\text{for all }i\in[N].
\]
Condition on the uploaded parameters $\{\hat{\theta}_i\}_{i=1}^N$ and define 
\[
\mathcal S := \operatorname{span}\{ \hat{\theta}_i-\hat{\theta}_1: i=2,\ldots,N \}, \qquad d_{\mathcal S}:=\dim(\mathcal S)\leq N-1,
\]
together with 
\[
\chi_\Phi := \frac{Q^2}{\sigma_{\min}(\Phi)}.
\]
There exist universal constants $C_0,C_1>0$ such that, whenever 
\[
n_m \geq C_1\chi_\Phi \left( d_{\mathcal S}+\log\frac{2}{\delta} \right),
\]
with probability at least $1-\delta$ over the public sample $\mathcal{D}_m^s$,
\begin{align}
\|\theta_m^{\mathrm{ext}}-\theta_m^*\|_{\mathrm F} \leq{}& \underbrace{ \kappa(\Phi)^{1/2}\varepsilon\, \mathrm{diam}_{\mathrm F}(\Theta) }_{\textup{(I) anchor error}} \nonumber\\
&+ \underbrace{ \kappa(\Phi)^{1/2} \sqrt{ \bigl( 2\varepsilon\,\mathrm{diam}_{\mathrm F}(\Theta) +\delta_{\mathrm{loc}} \bigr)\delta_{\mathrm{loc}} } }_{\textup{(II) client estimation}} \nonumber\\
&+ \underbrace{ \frac{ C_0 \sqrt{ \sigma_m^2+ Q^2 \bigl( \mathrm{diam}_{\mathrm F}(\Theta) +\delta_{\mathrm{loc}} \bigr)^2 } }{ \sigma_{\min}(\Phi)^{1/2} } \sqrt{ \frac{ d_{\mathcal S}+\log(2/\delta) }{ n_m } } }_{\textup{(III) public-data estimation}} . \label{eq:prop1}
\end{align}
In particular, when $\varepsilon=0$, Term~\textup{(II)} reduces to $\kappa(\Phi)^{1/2}\delta_{\mathrm{loc}}$. More generally, 
\[
\sqrt{ \bigl( 2\varepsilon\,\mathrm{diam}_{\mathrm F}(\Theta) +\delta_{\mathrm{loc}} \bigr)\delta_{\mathrm{loc}} } \leq \delta_{\mathrm{loc}} + \sqrt{ 2\varepsilon\,\mathrm{diam}_{\mathrm F}(\Theta) \,\delta_{\mathrm{loc}} },
\]
so Terms~\textup{(I)} and~\textup{(II)} both vanish as $\varepsilon,\delta_{\mathrm{loc}}\to0$, while Term~\textup{(III)} vanishes at the parametric $n_m^{-1/2}$ rate for fixed upload geometry.
\end{proposition}

 \paragraph{Proof of Proposition~\ref{prop:finitesample}.} We use the $\Phi$-projection characterization throughout. By~\eqnref{eq:loss_quadratic}, the population public loss for task $m$ obeys $\mathcal{L}_m(\theta) - \mathcal{L}_m(\theta_m^*) = \|\theta - \theta_m^*\|_\Phi^2$, so every population minimizer over a convex feasible set is the $\Phi$-projection of $\theta_m^*$ onto that set. Recall the norm equivalence $\sigma_{\min}(\Phi)^{1/2}\|\cdot\|_\mathrm{F} \leq \|\cdot\|_\Phi \leq \sigma_{\max}(\Phi)^{1/2}\|\cdot\|_\mathrm{F}$.

\textbf{A ladder of intermediate experts.} Let
\begin{align*}
\theta_m^{(0)} &:= \arg\min_{\theta \in \mathrm{conv}\{\theta_i^*\}} \mathcal{L}_m(\theta) && \text{(population loss, \emph{true} uploads),}\\
\theta_m^{(1)} &:= \arg\min_{\theta \in \mathrm{conv}\{\hat{\theta}_i\}} \mathcal{L}_m(\theta) && \text{(population loss, \emph{noisy} uploads),}\\
\theta_m^{(2)} &:= \arg\min_{\theta \in \mathrm{conv}\{\hat{\theta}_i\}} \widehat{\mathcal{L}}_m^s(\theta) = \theta_m^{\mathrm{ext}} && \text{(empirical loss, noisy uploads).}
\end{align*}
By the triangle inequality,
\begin{equation}
    \|\theta_m^{\mathrm{ext}} - \theta_m^*\|_\mathrm{F} \leq \underbrace{\|\theta_m^{(0)} - \theta_m^*\|_\mathrm{F}}_{(A)} + \underbrace{\|\theta_m^{(1)} - \theta_m^{(0)}\|_\mathrm{F}}_{(B1)} + \underbrace{\|\theta_m^{(2)} - \theta_m^{(1)}\|_\mathrm{F}}_{(B2)}.
\end{equation}
\textbf{Term (A) --- anchor error.} This is exactly Theorem~\ref{thm:recovery}:
\begin{equation}
    (A) \leq \kappa(\Phi)^{1/2}\cdot\varepsilon\cdot \mathrm{diam}_\mathrm{F}(\Theta),
\end{equation}
which is Term~\textup{(I)}. We also record the $\Phi$-norm form $\|\theta_m^{(0)} - \theta_m^*\|_\Phi \leq \sigma_{\max}(\Phi)^{1/2} \varepsilon\,\mathrm{diam}_\mathrm{F}(\Theta)$, used below.

\textbf{Term (B1) --- client noise via a variational projection estimate.} Both $\theta_m^{(0)}$ and $\theta_m^{(1)}$ are $\Phi$-projections of the \emph{same} point $z := \theta_m^*$ onto two convex sets, $K := \mathrm{conv}\{\theta_i^*\}$ and $\hat{K} := \mathrm{conv}\{\hat{\theta}_i\}$. Their Hausdorff distance satisfies $d_H(K,\hat{K}) \leq \max_i\|\hat{\theta}_i - \theta_i^*\|_\mathrm{F} \leq \delta_{\mathrm{loc}}$ in Frobenius norm, hence $\eta := d_H^\Phi(K,\hat{K}) \leq \sigma_{\max}(\Phi)^{1/2}\delta_{\mathrm{loc}}$ in the $\Phi$-metric. We \emph{cannot} invoke non-expansiveness here, because the projected point is fixed while the set moves --- the resulting dependence is H\"older-$\tfrac12$, not Lipschitz. Instead, write $p := \theta_m^{(0)} = \Pi_K^\Phi(z)$ and $p' := \theta_m^{(1)} = \Pi_{\hat K}^\Phi(z)$, and choose $q \in K$ with $\|q - p'\|_\Phi \leq \eta$ and $q' \in \hat{K}$ with $\|q' - p\|_\Phi \leq \eta$ (possible since $d_H^\Phi \leq \eta$). The variational characterizations of the two projections, $\langle z - p,\, q - p\rangle_\Phi \leq 0$ and $\langle z - p',\, q' - p'\rangle_\Phi \leq 0$, give
\begin{align}
    \langle z - p,\, p - p'\rangle_\Phi &= \langle z - p,\, p - q\rangle_\Phi + \langle z - p,\, q - p'\rangle_\Phi \;\geq\; -\,\|z - p\|_\Phi\,\eta, \\
    \langle z - p',\, p - p'\rangle_\Phi &= \langle z - p',\, p - q'\rangle_\Phi + \langle z - p',\, q' - p'\rangle_\Phi \;\leq\; \|z - p'\|_\Phi\,\eta.
\end{align}
Subtracting,
\begin{equation}
    \|p - p'\|_\Phi^2 = \langle z - p',\, p - p'\rangle_\Phi - \langle z - p,\, p - p'\rangle_\Phi \leq \bigl(\|z - p\|_\Phi + \|z - p'\|_\Phi\bigr)\,\eta. \label{eq:varproj}
\end{equation}
For $\|z - p'\|_\Phi$, the perturbed anchor $\hat{\theta}_{i_m} \in \hat{K}$ is feasible with $\|\theta_m^* - \hat{\theta}_{i_m}\|_\mathrm{F} \leq \varepsilon\,\mathrm{diam}_\mathrm{F}(\Theta) + \delta_{\mathrm{loc}}$, so $\|z - p'\|_\Phi \leq \sigma_{\max}(\Phi)^{1/2} (\varepsilon\,\mathrm{diam}_\mathrm{F}(\Theta) + \delta_{\mathrm{loc}})$. Substituting this and the Term~(A) bound on $\|z-p\|_\Phi$ together with $\eta$ into~\eqnref{eq:varproj},
\begin{equation}
    \|p - p'\|_\Phi^2 \leq \sigma_{\max}(\Phi)\, \bigl(2\varepsilon\,\mathrm{diam}_\mathrm{F}(\Theta) + \delta_{\mathrm{loc}}\bigr)\, \delta_{\mathrm{loc}}.
\end{equation}
Converting to Frobenius norm via $\|\cdot\|_\mathrm{F} \leq \sigma_{\min}(\Phi)^{-1/2}\|\cdot\|_\Phi$,
\begin{equation}
    (B1) = \|\theta_m^{(1)} - \theta_m^{(0)}\|_\mathrm{F} \leq \kappa(\Phi)^{1/2}\, \sqrt{\bigl(2\varepsilon\,\mathrm{diam}_\mathrm{F}(\Theta) + \delta_{\mathrm{loc}}\bigr)\,\delta_{\mathrm{loc}}}\,,
\end{equation}
which is Term~\textup{(II)}. At $\varepsilon = 0$ this is exactly $\kappa(\Phi)^{1/2}\delta_{\mathrm{loc}}$, recovering the Lipschitz rate.

\textbf{Term (B2) --- empirical versus population loss over a fixed feasible set.} Now $\theta_m^{(1)}$ and $\theta_m^{(2)}$ minimize, respectively, the population and empirical public losses over the same convex set
\[
\hat K := \operatorname{conv}\{ \hat{\theta}_1,\ldots,\hat{\theta}_N \}.
\]
Conditional on the uploaded parameters, $\hat K$ is fixed and the public sample remains i.i.d.

Define 
\[
\mathcal S := \operatorname{span}\{ \hat{\theta}_i-\hat{\theta}_1: i=2,\ldots,N \}, \qquad d_{\mathcal S}:=\dim(\mathcal S),
\]
and let 
\[
\Delta := \theta_m^{(2)}-\theta_m^{(1)}.
\]
Since both parameters lie in $\hat K$, 
\[
\Delta\in\mathcal S.
\]
For a public sample $\mathcal D_m^s=\{(x_j,y_j)\}_{j=1}^{n_m}$, write 
\[
\widehat{\mathbb E}_m[h] := \frac{1}{n_m}\sum_{j=1}^{n_m}h(x_j,y_j), \qquad \mathbb E_m[h] := \mathbb E_{(x,y)\sim\mathcal P_m}[h(x,y)],
\]
and define the residual at the population in-class optimum 
\[
r_1(x,y) := y-f_0(x)-\theta_m^{(1)}\phi(x).
\]
Because $\theta_m^{(2)}$ minimizes the empirical squared loss over $\hat K$,
\begin{equation}
\widehat{\mathbb E}_m \|\Delta\phi(x)\|_2^2 \leq 2\widehat{\mathbb E}_m \langle r_1(x,y),\Delta\phi(x)\rangle . \label{eq:b2_basic}
\end{equation}
On the other hand, first-order optimality of $\theta_m^{(1)}$ for the population loss over the convex set $\hat K$ gives 
\[
\left\langle \nabla\mathcal L_m(\theta_m^{(1)}), \Delta \right\rangle_{\mathrm F} \geq 0.
\]
Since 
\[
\nabla\mathcal L_m(\theta_m^{(1)}) = -2\mathbb E_m[ r_1(x,y)\phi(x)^\top],
\]
this is equivalent to 
\[
\mathbb E_m \langle r_1(x,y),\Delta\phi(x)\rangle \leq 0.
\]
Combining with~\eqnref{eq:b2_basic},
\begin{equation}
\widehat{\mathbb E}_m \|\Delta\phi(x)\|_2^2 \leq 2 \left| (\widehat{\mathbb E}_m-\mathbb E_m) \langle r_1(x,y),\Delta\phi(x)\rangle \right|. \label{eq:b2_centered}
\end{equation}

\textit{Step (i): restricted empirical quadratic form.} Choose a $\Phi$-orthonormal basis $U_1,\ldots,U_{d_{\mathcal S}}$ of $\mathcal S$, i.e.
\[
\langle U_j,U_k\rangle_\Phi := \operatorname{tr}(U_j\Phi U_k^\top) = \mathbf 1\{j=k\}.
\]
For $a\in\mathbb R^{d_{\mathcal S}}$, write
\[
\Delta(a):=\sum_{k=1}^{d_{\mathcal S}}a_kU_k, \qquad \|\Delta(a)\|_\Phi=\|a\|_2,
\]
and define
\[
V(x) := \bigl[ U_1\phi(x),\ldots, U_{d_{\mathcal S}}\phi(x) \bigr] \in\mathbb R^{l\times d_{\mathcal S}}.
\]
Then
\[
\mathbb E_m[V(x)^\top V(x)] = I_{d_{\mathcal S}}.
\]
Furthermore,
\[
\|V(x)^\top V(x)\|_{\mathrm{op}} = \sup_{\|a\|_2=1} \|\Delta(a)\phi(x)\|_2^2 \leq \frac{Q^2}{\sigma_{\min}(\Phi)} = \chi_\Phi.
\]
Therefore, a matrix Chernoff bound for sums of independent positive semidefinite matrices gives, with probability at least $1-\delta/2$,
\begin{equation}
\widehat{\mathbb E}_m \|\Gamma\phi(x)\|_2^2 \geq \frac12\|\Gamma\|_\Phi^2 \qquad \text{for every }\Gamma\in\mathcal S, \label{eq:b2_rsc}
\end{equation}
provided
\[
n_m \geq C\chi_\Phi \log\frac{2d_{\mathcal S}}{\delta}.
\]
The sample-size condition in the proposition is stronger than this requirement and therefore suffices.

\textit{Step (ii): the centered linear process.} Let
\[
R_m := \|\theta_m^{(1)}-\theta_m^*\|_{\mathrm F}.
\]
Because $\theta_m^{(1)}\in\hat K$, there is a convex combination of the true uploads within $\delta_{\mathrm{loc}}$ of $\theta_m^{(1)}$. Since every true client optimum lies in $\operatorname{conv}(\Theta)$,
\begin{equation}
R_m \leq \mathrm{diam}_{\mathrm F}(\Theta) +\delta_{\mathrm{loc}}. \label{eq:b2_R}
\end{equation}
Writing
\[
r_1(x,y) = (\theta_m^*-\theta_m^{(1)})\phi(x) +\zeta_m,
\]
consider any $\Gamma\in\mathcal S$ with $\|\Gamma\|_\Phi=1$. The centered random variable
\[
Z_\Gamma := \langle r_1(x,y),\Gamma\phi(x)\rangle - \mathbb E_m \langle r_1(x,y),\Gamma\phi(x)\rangle
\]
has variance bounded by
\begin{equation}
\operatorname{Var}(Z_\Gamma) \leq \sigma_m^2+Q^2R_m^2. \label{eq:b2_var}
\end{equation}
Indeed, the noise contribution has variance
\[
\mathbb E_m[ \sigma_m^2\|\Gamma\phi(x)\|_2^2] = \sigma_m^2,
\]
while the signal contribution satisfies
\[
\mathbb E_m \langle (\theta_m^*-\theta_m^{(1)})\phi(x), \Gamma\phi(x) \rangle^2 \leq Q^2R_m^2 \mathbb E_m\|\Gamma\phi(x)\|_2^2 = Q^2R_m^2.
\]
The cross term vanishes because $\mathbb E[\zeta_m\mid x]=0$.

Moreover,
\[
\|\Gamma\|_{\mathrm F} \leq \sigma_{\min}(\Phi)^{-1/2},
\]
so
\[
\|\Gamma\phi(x)\|_2 \leq \sqrt{\chi_\Phi}.
\]
Assumption~\ref{ass:public_tail} therefore makes the noise component sub-Gaussian at scale at most $\sigma_m\sqrt{\chi_\Phi}$, while the centered signal component is bounded at scale at most a universal constant times $Q R_m\sqrt{\chi_\Phi}$. Consequently $Z_\Gamma$ satisfies a Bernstein-type tail bound with variance proxy
\[
v_m^2 := \sigma_m^2+Q^2R_m^2
\]
and scale at most $C\sqrt{\chi_\Phi}\,v_m$.

Apply this scalar Bernstein bound on a $1/2$-net of the unit sphere in the $d_{\mathcal S}$-dimensional space $\mathcal S$. Such a net has cardinality at most $5^{d_{\mathcal S}}$. A union bound and the standard net extension therefore give, with probability at least $1-\delta/2$,
\begin{align}
&\sup_{\substack{\Gamma\in\mathcal S\\ \|\Gamma\|_\Phi=1}} \left| (\widehat{\mathbb E}_m-\mathbb E_m) \langle r_1(x,y),\Gamma\phi(x)\rangle \right| \nonumber\\
&\qquad\leq C v_m \left[ \sqrt{ \frac{ d_{\mathcal S}+\log(2/\delta) }{ n_m } } + \sqrt{\chi_\Phi}\, \frac{ d_{\mathcal S}+\log(2/\delta) }{ n_m } \right]. \label{eq:b2_linear_raw}
\end{align}
Under the proposition's sample-size condition
\[
n_m \geq C_1\chi_\Phi \left( d_{\mathcal S}+\log\frac{2}{\delta} \right),
\]
the second term in brackets is dominated by the first. Hence, after enlarging the universal constant,
\begin{equation}
\left| (\widehat{\mathbb E}_m-\mathbb E_m) \langle r_1(x,y),\Gamma\phi(x)\rangle \right| \leq C v_m \|\Gamma\|_\Phi \sqrt{ \frac{ d_{\mathcal S}+\log(2/\delta) }{ n_m } } \label{eq:b2_linear}
\end{equation}
simultaneously for every $\Gamma\in\mathcal S$.

\textit{Step (iii): parameter error.} Applying~\eqnref{eq:b2_rsc} and~\eqnref{eq:b2_linear} to $\Delta=\theta_m^{(2)}-\theta_m^{(1)}$ in \eqnref{eq:b2_centered} gives
\[
\frac12\|\Delta\|_\Phi^2 \leq 2C v_m \|\Delta\|_\Phi \sqrt{ \frac{ d_{\mathcal S}+\log(2/\delta) }{ n_m } }.
\]
If $\Delta=0$ there is nothing to prove. Otherwise, dividing by $\|\Delta\|_\Phi$ and absorbing numerical constants yields
\[
\|\Delta\|_\Phi \leq C_0 v_m \sqrt{ \frac{ d_{\mathcal S}+\log(2/\delta) }{ n_m } }.
\]
Using~\eqnref{eq:b2_R} and $\|\Delta\|_{\mathrm F} \leq \sigma_{\min}(\Phi)^{-1/2}\|\Delta\|_\Phi$, we obtain
\begin{equation}
(B2) \leq \frac{ C_0 \sqrt{ \sigma_m^2+ Q^2 \bigl( \mathrm{diam}_{\mathrm F}(\Theta) +\delta_{\mathrm{loc}} \bigr)^2 } }{ \sigma_{\min}(\Phi)^{1/2} } \sqrt{ \frac{ d_{\mathcal S}+\log(2/\delta) }{ n_m } },
\end{equation}
which is Term~\textup{(III)}.

\textbf{Collecting all terms.} Combining (A), (B1), and (B2),
\begin{align}
\|\theta_m^{\mathrm{ext}}-\theta_m^*\|_{\mathrm F} \leq{}& \underbrace{ \kappa(\Phi)^{1/2}\varepsilon\, \mathrm{diam}_{\mathrm F}(\Theta) }_{\textup{(I)}} \nonumber\\
&+ \underbrace{ \kappa(\Phi)^{1/2} \sqrt{ \bigl( 2\varepsilon\,\mathrm{diam}_{\mathrm F}(\Theta) +\delta_{\mathrm{loc}} \bigr)\delta_{\mathrm{loc}} } }_{\textup{(II)}} \nonumber\\
&+ \underbrace{ \frac{ C_0 \sqrt{ \sigma_m^2+ Q^2 \bigl( \mathrm{diam}_{\mathrm F}(\Theta) +\delta_{\mathrm{loc}} \bigr)^2 } }{ \sigma_{\min}(\Phi)^{1/2} } \sqrt{ \frac{ d_{\mathcal S}+\log(2/\delta) }{ n_m } } }_{\textup{(III)}} .
\end{align}
This is exactly the bound in Proposition~\ref{prop:finitesample}. \qed

The bound separates three sources of recovery error: anchor impurity, local estimation error, and finite public-sample error. These terms vanish respectively as $\varepsilon\to0$, $\delta_{\mathrm{loc}}\to0$, and $n_m\to\infty$.

\begin{remark}[On the coupling of the error terms]
\label{rem:cross_term}
Term~(II) is not fully decoupled from the anchor error: when $\varepsilon>0$ the perturbed feasible hull $\mathrm{conv}\{\hat{\theta}_i\}$ is displaced from a set whose $\Phi$-projection of $\theta_m^*$ is itself $O(\varepsilon)$ away, producing the cross-term $\sqrt{\varepsilon\,\delta_{\mathrm{loc}}}$. This interaction is intrinsic to projecting a fixed point onto a moving convex set --- the dependence on the set is H\"older-$\tfrac12$ in Hausdorff distance, not Lipschitz --- and it vanishes both as $\varepsilon\to0$ (pure anchors, Lipschitz recovery) and as $\delta_{\mathrm{loc}}\to0$ (exact uploads). The interaction therefore does not alter consistency of the finite-sample recovery bound: together with Term~\textup{(III)} of Proposition~\ref{prop:finitesample}, the full recovery error vanishes as $\varepsilon\to0$, $\delta_{\mathrm{loc}}\to0$, and $n_m\to\infty$ for fixed upload geometry.
\end{remark}

\subsection{Identifiability Limits}
\label{app:identifiability}

Throughout, an \emph{instance} $\mathcal{I} = (\{\theta_m^*\}_{m=1}^M, \{\alpha_i\}_{i=1}^N)$ satisfies Assumptions~\ref{ass:linear}, with task $m$ generated by $y = f_0(x) + \theta_m^*\phi(x) + \zeta_m$ over the common input marginal; by Theorem~\ref{thm:mixture} it induces the upload pool $\theta_i^* = \sum_m \alpha_{im} \theta_m^*$.

\paragraph{Proof of Theorem~\ref{thm:label} (Label indeterminacy).}

\textbf{Step 1. The permuted instance is observationally equivalent.} Write $\theta_m'(\mathcal{I}^\pi) = \theta_{\pi(m)}^*$, $\alpha'_{im} = \alpha_{i\pi(m)}$, and $\mathcal{P}'_m = \mathcal{P}_{\pi(m)}$. The three label-free observables coincide. \emph{(i) Client data.} Under $\mathcal{I}^\pi$, client $i$ draws from $\sum_m \alpha'_{im}\mathcal{P}'_m = \sum_m \alpha_{i\pi(m)}\mathcal{P}_{\pi(m)} = \sum_k \alpha_{ik}\mathcal{P}_k$ (reindex $k = \pi(m)$), exactly client $i$'s distribution under $\mathcal{I}$. \emph{(ii) Upload pool.} By Theorem~\ref{thm:mixture}, $\theta_i^*(\mathcal{I}^\pi) = \sum_m \alpha'_{im} \theta'_m = \sum_k \alpha_{ik} \theta_k^* = \theta_i^*(\mathcal{I})$. \emph{(iii) Unlabeled corpus.} $\frac{1}{M}\sum_m \mathcal{P}'_m = \frac{1}{M}\sum_k \mathcal{P}_k$ is permutation-invariant.

\textbf{Step 2. Two-point argument.} Choose a pair $(m,m')$ attaining
\[
\|\theta_m^*-\theta_{m'}^*\|_{\mathrm F} = \mathrm{diam}_{\mathrm F}(\Theta),
\]
and let $\pi$ transpose this pair. By Step~1, $\mathcal I$ and $\mathcal I^\pi$ induce identical observables, so $\mathcal A$ has the same output distribution on both instances. Couple the two runs to have the same realization $(\hat\theta_1,\ldots,\hat\theta_M)$. Then
\begin{align*}
& \max_k \|\hat\theta_k-\theta_k^*(\mathcal I)\|_{\mathrm F} + \max_k \|\hat\theta_k-\theta_k^*(\mathcal I^\pi)\|_{\mathrm F} \\
&\qquad\ge \|\hat\theta_m-\theta_m^*\|_{\mathrm F} + \|\hat\theta_m-\theta_{m'}^*\|_{\mathrm F} \\
&\qquad\ge \|\theta_m^*-\theta_{m'}^*\|_{\mathrm F} = \mathrm{diam}_{\mathrm F}(\Theta).
\end{align*}
Taking expectations shows that at least one of the two observationally equivalent instances has task-indexed error at least $\mathrm{diam}_{\mathrm F}(\Theta)/2$. Since both belong to $\mathcal O(\mathcal I)$,~\eqnref{eq:label_lower_bound} follows.

\begin{remark}[Scope]
The argument uses only that the abstract tasks are exchangeable within the linear model. If, in a given application, tasks carry a priori distinguishable structure known to the estimator (e.g., disjoint label vocabularies), the permutation symmetry is broken by that side information---which is itself a form of task supervision, consistent with our thesis.
\end{remark}

\paragraph{Proof of Theorem~\ref{thm:vertex} (Vertex-set non-identifiability).}

\textbf{Step 1. Shrunk instance.} Let $\bar{\theta} := \frac{1}{M}\sum_m \theta_m^*$ and, for $t \in (0,1)$, set
\begin{equation}
    \tilde{\theta}_m := (1 - t)\,\theta_m^* + t\,\bar{\theta}, \qquad \tilde{\alpha}_{im} := \frac{\alpha_{im} - t/M}{1 - t}.
\end{equation}
Since each client is $\gamma$-interior and $t \leq \gamma$, $\alpha_{im} \geq \gamma/M \geq t/M$, so $\tilde{\alpha}_{im} \geq 0$; and $\sum_m \tilde{\alpha}_{im} = (1 - M\cdot t/M)/(1-t) = 1$. Thus $\tilde{\mathcal{I}} := (\{\tilde{\theta}_m\}, \{\tilde{\alpha}_i\})$ is a valid instance.

\textbf{Step 2. Identical pools.} For every $i$,
\begin{align}
    \sum_m \tilde{\alpha}_{im} \tilde{\theta}_m &= \sum_m \frac{\alpha_{im} - t/M}{1-t} \bigl[(1-t)\theta_m^* + t\bar{\theta}\bigr] = \sum_m (\alpha_{im} - t/M)\theta_m^* + \frac{t}{1-t}\bar{\theta}\sum_m(\alpha_{im} - t/M) \notag\\
    &= \theta_i^* - t\bar{\theta} + \frac{t}{1-t}\bar{\theta}(1-t) = \theta_i^*,
\end{align}
so $\mathcal{P}(\tilde{\mathcal{I}}) = \mathcal{P}(\mathcal{I})$ and any pool-only estimator has the same output law on both.

\textbf{Step 3. Vertex-set geometry.} With $\bar{\theta}_{-m} := \frac{1}{M-1}\sum_{k \neq m} \theta_k^*$, $\|\tilde{\theta}_m - \theta_m^*\|_\mathrm{F} = t\|\theta_m^* - \bar{\theta}\|_\mathrm{F} = t\frac{M-1}{M}\|\theta_m^* - \bar{\theta}_{-m}\|_\mathrm{F} \geq t\frac{M-1}{M}h$, since $\bar{\theta}_{-m} \in \mathrm{aff}\{\theta_k^* : k \neq m\}$. Also $\|\theta_m^* - \bar{\theta}\|_\mathrm{F} \leq \mathrm{diam}_\mathrm{F}(\Theta)$, so the displacement is at most $t\,\mathrm{diam}_\mathrm{F}(\Theta) \leq \mathrm{sep}/2$ by the choice of $t$. For $k \neq m$, $\|\theta_m^* - \tilde{\theta}_k\|_\mathrm{F} \geq \mathrm{sep} - t\,\mathrm{diam}_\mathrm{F}(\Theta) \geq \mathrm{sep}/2$. Hence the minimal distance between the two vertex sets is
\begin{equation}
    d := \min_{m,k}\|\theta_m^* - \tilde{\theta}_k\|_\mathrm{F} \;\geq\; t\frac{M-1}{M}h \;>\; 0. \label{eq:d_bound}
\end{equation}

\textbf{Step 4. Matching.} For any common output $\hat{\mathcal{V}}$, let $e_1 := \min_\pi \max_m \|\hat{\theta}_{\pi(m)} - \theta_m^*\|_\mathrm{F}$ and $e_2 := \min_\pi \max_m \|\hat{\theta}_{\pi(m)} - \tilde{\theta}_m\|_\mathrm{F}$. Let $\pi_1$ and $\pi_2$ attain these minima. For any $m$, set $k := \pi_2^{-1}(\pi_1(m))$. Then
\[
\|\theta_m^*-\tilde{\theta}_k\|_\mathrm{F} \leq \|\theta_m^*-\hat{\theta}_{\pi_1(m)}\|_\mathrm{F} + \|\hat{\theta}_{\pi_2(k)}-\tilde{\theta}_k\|_\mathrm{F} \leq e_1+e_2.
\]
By~\eqnref{eq:d_bound}, $e_1+e_2\geq d$ for every realization. Therefore $\mathbb E_{\mathcal A}[e_1]+\mathbb E_{\mathcal A}[e_2]\geq d$, so
\[
\max\{\mathbb E_{\mathcal A}[e_1],\mathbb E_{\mathcal A}[e_2]\} \geq \frac{d}{2} \geq \frac{M-1}{2M}\,t\,h,
\]
which proves the claim. \qed

\begin{remark}[Relation between interiority and anchor quality]
\label{rem:optimality}
Under the $\gamma$-interior condition, every client's purity is at most $1-(M-1)\gamma/M$, so the best achievable anchor impurity satisfies $\varepsilon\geq(M-1)\gamma/M$. For $\gamma\leq\mathrm{sep}/(2\,\mathrm{diam}_{\mathrm F}(\Theta))$, Theorem~\ref{thm:vertex} gives a pool-only lower bound linear in $\gamma$. Hence, on families for which $\varepsilon=\Theta(\gamma)$, pool-only vertex recovery has a linear lower bound in anchor impurity, matching the linear dependence of Theorem~\ref{thm:recovery} up to conditioning and simplex geometry. This lower bound concerns upload-only recovery; supervised recovery is distinguished separately in Remark~\ref{rem:anchor_two_reasons}.
\end{remark}

\begin{remark}[Distinct roles of anchor information]
\label{rem:anchor_two_reasons}
For pool-only estimators, Theorem~\ref{thm:vertex} gives an information-theoretic obstruction when the population remains $\gamma$-interior: without anchors the upload pool alone is consistent with multiple vertex sets (Theorem~\ref{thm:vertex}). The supervised program is not subject to this particular obstruction --- the instances $\mathcal{I}$ and $\tilde{\mathcal{I}}$ of Theorem~\ref{thm:vertex} induce \emph{different} labeled public losses, since $\tilde{\theta}_m \neq \theta_m^*$, so the $\boldsymbol{\Omega}$ program can distinguish them. The reason the program nonetheless requires anchors is \emph{algorithmic rather than information-theoretic}: its expert is constrained to $\mathrm{conv}\{\text{uploads}\}$, and when no near-pure client exists for task $m$, the true vertex $\theta_m^*$ lies strictly \emph{outside} this hull, so no convex combination of uploads can reach it --- exactly the $O(\varepsilon)$ projection gap (Term~(I)) of Theorem~\ref{thm:recovery}. These mechanisms are distinct. Theorem~\ref{thm:vertex} establishes an information-theoretic limitation of pool-only recovery, whereas the anchor requirement of the supervised extraction program is a feasibility limitation induced by its convex-hull constraint.
\end{remark}

\begin{proposition}[Supervised identification margin]
\label{prop:margin}
Let $\mathcal{V} = \{V_1, \ldots, V_M\}$ be any candidate vertex set with $\max_m \min_k \|V_k - \theta_m^*\|_\mathrm{F} \leq \eta$ for
\begin{equation}
    \eta \;\leq\; \frac{\mathrm{sep}}{2\bigl(1 + \kappa(\Phi)^{1/2}\bigr)}, \label{eq:eta_cond}
\end{equation}
and define the supervised assignment $\hat{\sigma}(m) := \arg\min_{k \in [M]} \mathcal{L}_m(V_k)$. \textup{(a)} With the population public loss, $\hat{\sigma}$ is exactly correct, and the loss margin over every incorrect candidate is at least $\tfrac{1}{4}\,\sigma_{\min}(\Phi)\,\mathrm{sep}^2$. \textup{(b)} Under Assumptions~\ref{ass:linear} and~\ref{ass:bounded} with sub-Gaussian noise $\|\zeta_m\|_{\psi_2} \leq \sigma_m$, there are universal constants $c, C > 0$ such that if
\begin{equation}
    n_m \;\geq\; C\,(K/G)^2\,\log(M^2/\delta), \quad K := Q^2\mathrm{diam}_\mathrm{F}(\Theta)^2 + \sigma_m Q\,\mathrm{diam}_\mathrm{F}(\Theta), \quad G := \tfrac{1}{4}\sigma_{\min}(\Phi)\,\mathrm{sep}^2,
\end{equation}
then with probability at least $1 - \delta$ the empirical assignment $\arg\min_k \widehat{\mathcal{L}}_m^s(V_k)$ is correct simultaneously for all $m \in [M]$.
\end{proposition}

\paragraph{Proof of Proposition~\ref{prop:margin} (Supervised identification margin).}

\textbf{Part (a).} From~\eqnref{eq:loss_quadratic} specialized to the pure task-$m$ distribution, $\mathcal{L}_m(\theta) - \mathcal{L}_m(\theta_m^*) = \|\theta - \theta_m^*\|_\Phi^2$. Let $V_{k(m)}$ satisfy $\|V_{k(m)} - \theta_m^*\|_\mathrm{F} \leq \eta$, and let $V_j$ be within $\eta$ of some $\theta_{m'}^*$, $m' \neq m$ (once $\eta < \mathrm{sep}/2$, guaranteed by~\eqnref{eq:eta_cond}, the candidate-to-vertex assignment is injective). Then
\begin{align}
    \mathcal{L}_m(V_j) - \mathcal{L}_m(V_{k(m)}) &= \|V_j - \theta_m^*\|_\Phi^2 - \|V_{k(m)} - \theta_m^*\|_\Phi^2 \geq \sigma_{\min}(\Phi)(\mathrm{sep} - \eta)^2 - \sigma_{\max}(\Phi)\eta^2,
\end{align}
using $\|V_j - \theta_m^*\|_\mathrm{F} \geq \mathrm{sep} - \eta$ and norm equivalence. By~\eqnref{eq:eta_cond}, $\eta \leq \mathrm{sep}/(2(1+\kappa^{1/2})) \leq \mathrm{sep}/4$, so $\sigma_{\min}^{1/2}(\mathrm{sep} - \eta) \geq \tfrac34\sigma_{\min}^{1/2}\mathrm{sep}$ while $\sigma_{\max}^{1/2}\eta = \sigma_{\min}^{1/2}\kappa^{1/2}\eta \leq \tfrac12\sigma_{\min}^{1/2}\mathrm{sep}$. Therefore the gap is at least $(\tfrac{9}{16} - \tfrac14)\sigma_{\min}(\Phi)\mathrm{sep}^2 \geq \tfrac14\sigma_{\min}(\Phi)\mathrm{sep}^2 = G$, so the population argmin selects $V_{k(m)}$.

\textbf{Part (b).} Fix $m$ and a pair $(V_{k(m)}, V_j)$. For $(x,y) \sim \mathcal{P}_m$ let $g(x,y) := \ell(V_j,x,y) - \ell(V_{k(m)},x,y)$. With $\Delta_j := V_j - \theta_m^*$, $\Delta_k := V_{k(m)} - \theta_m^*$ and $y - f_0(x) = \theta_m^*\phi(x) + \zeta_m$, $\mathbb{E}[g] = \|\Delta_j\|_\Phi^2 - \|\Delta_k\|_\Phi^2 \geq G$ by part (a). By Assumption~\ref{ass:bounded} the quadratic terms are bounded by $C_1 Q^2 \mathrm{diam}_\mathrm{F}(\Theta)^2$ a.s., and the cross term is conditionally $\sigma_m$-sub-Gaussian scaled by at most $C_1 Q\,\mathrm{diam}_\mathrm{F}(\Theta)$; hence $\|g - \mathbb{E}g\|_{\psi_1} \leq C_2 K$ with $K = Q^2\mathrm{diam}_\mathrm{F}(\Theta)^2 + \sigma_m Q\, \mathrm{diam}_\mathrm{F}(\Theta)$. Bernstein's inequality gives, for the empirical mean $\bar{g}$ over $n_m$ samples, $\mathbb{P}(\bar{g} \leq 0) \leq 2\exp[-c\,n_m \min(G^2/K^2, G/K)]$. A union bound over the at most $M(M-1)$ ordered pairs shows all empirical assignments are simultaneously correct with probability $\geq 1 - \delta$ provided $n_m \geq C(K/G)^2\log(M^2/\delta)$. \qed

\paragraph{Relation to supervised recovery.}

The preceding results distinguish three aspects of expert recovery: recovery accuracy under task supervision, recovery of the unordered vertex set from client uploads, and identification of the expert-to-task correspondence. We next relate the supervised upper bound of Theorem~\ref{thm:recovery} to the two identifiability lower bounds.

Theorem~\ref{thm:vertex} addresses recovery of the vertex \emph{set} from the upload pool alone. Under the $\gamma$-interior condition its lower bound is linear in $\gamma$ whenever $\gamma\leq\mathrm{sep}/(2\,\mathrm{diam}_{\mathrm F}(\Theta))$. Moreover, every $\gamma$-interior population has anchor impurity at least $(M-1)\gamma/M$. Thus the lower bound captures the same linear dependence on distance from the simplex boundary as the anchor term in Theorem~\ref{thm:recovery}; for families in which $\varepsilon=\Theta(\gamma)$, the two scale identically up to conditioning and simplex geometry. Theorem~\ref{thm:vertex} applies to pool-only recovery. FedSEE's supervised program uses labeled public loss to identify task directions, while its anchor condition controls the geometric projection of each task vertex onto the convex hull of the uploads (Remark~\ref{rem:anchor_two_reasons}).

Furthermore, a sharp distinction exists between recovering the vertex set and identifying its task correspondence. Boundary information can make the unordered vertex set recoverable, but it cannot by itself determine which vertex corresponds to which task. As anchor clients approach purity, the supervised recovery error vanishes, whereas the task-indexed label-blind floor of Theorem~\ref{thm:label} remains fixed. The task-labeled public loss supplies an identification margin: Proposition~\ref{prop:margin} shows that the population loss for task $m$ separates its correct vertex from the alternatives by at least $\tfrac14\sigma_{\min}(\Phi)\,\mathrm{sep}^2$. Thus task-indexed side information is necessary to establish the true vertex--task correspondence; in FedSEE, this information is supplied by the labeled public data.

\subsection{Mixture Estimation and Static Reconstruction}
\label{app:mixture_estimation}

Given a client upload $\hat{\theta}_i$ and the recovered experts, define its barycentric mixture estimate by
\begin{equation}
    \hat{\alpha}_i := \underset{a\in\Delta^{M-1}}{\arg\min}\; \left\| \hat{\theta}_i - \sum_{m=1}^{M} a_m\theta_m^{\mathrm{ext}} \right\|_{\Phi}^{2}. \label{eq:gen_bary}
\end{equation}

The following result quantifies the accuracy of this estimator in terms of the local upload error, expert-recovery error, feature conditioning, and simplex geometry. We first record a geometric inequality used in the analysis.

\begin{lemma}[Centered combinations]
\label{lem:centered}
For any $c \in \mathbb{R}^M$ with $\mathbf{1}^\top c = 0$, $\|\sum_m c_m \theta_m^*\|_\mathrm{F} \geq \|c\|_\infty h$.
\end{lemma}

\paragraph{Proof.} If $c=0$, the claim is immediate. Otherwise, let $m^* := \arg\max_m |c_m|$. Since $\mathbf{1}^\top c = 0$, $\sum_m c_m \theta_m^* = c_{m^*}(\theta_{m^*}^* - \sum_{k \neq m^*}\beta_k \theta_k^*)$ with $\beta_k := -c_k/c_{m^*}$, $\sum_k \beta_k = 1$, an affine combination of $\{\theta_k^* : k \neq m^*\}$. Thus
\[
\left\|\sum_m c_m\theta_m^*\right\|_\mathrm{F} = |c_{m^*}| \left\|\theta_{m^*}^*-\sum_{k\neq m^*}\beta_k\theta_k^*\right\|_\mathrm{F} \geq |c_{m^*}|\, \mathrm{dist}_\mathrm{F} \left( \theta_{m^*}^*, \mathrm{aff}\{\theta_k^*:k\neq m^*\} \right) \geq \|c\|_\infty h.
\]

\qed

\begin{proposition}[Barycentric mixture estimation]
\label{prop:mixture_est}
Under Assumptions~\ref{ass:linear} and~\ref{ass:bounded} with affinely independent experts ($h > 0$), let $\hat{\alpha}_i$ solve the barycentric projection~\eqnref{eq:gen_bary} using an upload with $\|\hat{\theta}_i - \theta_i^*\|_\mathrm{F} \leq \delta_{\mathrm{loc}}$ and recovered experts with $\max_m\|\theta_m^{\mathrm{ext}} - \theta_m^*\|_\mathrm{F} \leq \xi_{\max}$ (Proposition~\ref{prop:finitesample}). Then
\begin{equation}
    \|\hat{\alpha}_i - \alpha_i\|_1 \;\leq\; \frac{2M}{h}\Bigl(\kappa(\Phi)^{1/2}\, (\delta_{\mathrm{loc}} + \xi_{\max}) + \xi_{\max}\Bigr) \;=:\; \eta_C .
\end{equation}
In particular, $\eta_C\to0$ as $\delta_{\mathrm{loc}}\to0$ and $\xi_{\max}\to0$, and $\hat{\alpha}_i=\alpha_i$ when $\delta_{\mathrm{loc}}=\xi_{\max}=0$.
\end{proposition}

\paragraph{Proof.} Write $\hat{\theta}_m := \theta_m^{\mathrm{ext}}$, $E_m := \hat{\theta}_m - \theta_m^*$
with $\|E_m\|_\mathrm{F} \leq \xi_{\max}$, and $e_i := \hat{\theta}_i - \theta_i^*$ with $\|e_i\|_\mathrm{F} \leq \delta_{\mathrm{loc}}$. Let $a^* := \alpha_i$, $\hat{a} := \hat{\alpha}_i$, and $c := \hat{a} - a^*$, so that $\mathbf{1}^\top c = 0$ (both lie in $\Delta^{M-1}$).

\textbf{Step 1. Residual at the truth.} Using $\theta_i^* = \sum_m \alpha_{im} \theta_m^*$ (Theorem~\ref{thm:mixture}), the residual of the projection objective~\eqnref{eq:gen_bary} at $a^*$ is
\begin{equation}
    \rho^* := \sum_m a_m^* \hat{\theta}_m - \hat{\theta}_i = \sum_m \alpha_{im} E_m - e_i, \qquad \|\rho^*\|_\Phi \leq \sigma_{\max}(\Phi)^{1/2} (\xi_{\max} + \delta_{\mathrm{loc}}) =: r^* ,
\end{equation}
using $\sum_m \alpha_{im} = 1$ and $\|\cdot\|_\Phi \leq \sigma_{\max}(\Phi)^{1/2}\|\cdot\|_\mathrm{F}$.

\textbf{Step 2. Near-optimal residual.} By optimality of $\hat{a}$ and feasibility of $a^*$ for the convex program~\eqnref{eq:gen_bary}, $\|\sum_m \hat{a}_m \hat{\theta}_m - \hat{\theta}_i\|_\Phi \leq \|\rho^*\|_\Phi \leq r^*$. Since $\sum_m \hat{a}_m \hat{\theta}_m - \hat{\theta}_i = \sum_m c_m \hat{\theta}_m + \rho^*$, the triangle inequality gives
\begin{equation}
    \Bigl\|\sum_m c_m \hat{\theta}_m\Bigr\|_\Phi \leq 2 r^* , \qquad\text{hence}\qquad \Bigl\|\sum_m c_m \hat{\theta}_m\Bigr\|_\mathrm{F} \leq \frac{2 r^*}{\sigma_{\min}(\Phi)^{1/2}} = 2\kappa(\Phi)^{1/2}(\xi_{\max} + \delta_{\mathrm{loc}}).
\end{equation}

\textbf{Step 3. Lower bound via the simplex height.} Since $\mathbf{1}^\top c = 0$, Lemma~\ref{lem:centered} gives $\|\sum_m c_m \theta_m^*\|_\mathrm{F} \geq \|c\|_\infty\, h$. Using $\|\sum_m c_m E_m\|_\mathrm{F} \leq \|c\|_1 \xi_{\max} \leq 2\xi_{\max}$ (as $\|c\|_1 \leq 2$ on the simplex),
\begin{equation}
    \Bigl\|\sum_m c_m \hat{\theta}_m\Bigr\|_\mathrm{F} \geq \Bigl\|\sum_m c_m \theta_m^*\Bigr\|_\mathrm{F} - \Bigl\|\sum_m c_m E_m\Bigr\|_\mathrm{F} \geq \|c\|_\infty\, h - 2\xi_{\max} .
\end{equation}

\textbf{Step 4. Combine.} Chaining Steps 2 and 3, $\|c\|_\infty\, h \leq 2\kappa(\Phi)^{1/2} (\xi_{\max}+\delta_{\mathrm{loc}}) + 2\xi_{\max}$, and using $\|c\|_1 \leq M\|c\|_\infty$,
\begin{equation}
    \|\hat{\alpha}_i - \alpha_i\|_1 = \|c\|_1 \leq \frac{2M}{h}\Bigl(\kappa(\Phi)^{1/2} (\delta_{\mathrm{loc}}+\xi_{\max}) + \xi_{\max}\Bigr) = \eta_C.
\end{equation}
\qed
 
The dependence on $h$ shows that greater affine separation among the task experts improves the conditioning of barycentric mixture estimation. The same geometric quantity also appears in the pool-only vertex-recovery lower bound, although it plays a different role in that result. We next propagate the mixture-estimation and expert-recovery errors to the reconstruction of the client's optimal static adapter.

\begin{corollary}[Static barycentric reconstruction]
\label{cor:personalization}
Under Assumptions~\ref{ass:linear} and~\ref{ass:bounded}, with the barycentric mixture estimate $\hat{\alpha}_i$ satisfying $\|\hat{\alpha}_i-\alpha_i\|_1\leq\eta_C$ (Proposition~\ref{prop:mixture_est}), the reconstructed static adapter satisfies
\begin{equation}
    \mathcal{L}_i\!\left(\sum_m \hat{\alpha}_{im} \theta_m^{\mathrm{ext}}\right) - \mathcal{L}_i(\theta_i^*) \;\leq\; \sigma_{\max}(\Phi)\Delta_i^2,
\end{equation}

where
\[
\Delta_i := \eta_C\,\mathrm{diam}_\mathrm{F}(\Theta)+\xi_{\max}.
\]
Hence $\Delta_i\to0$ yields recovery of the optimal static adapter $\theta_i^*$.
\end{corollary}

\paragraph{Proof.}
Define the reconstructed static adapter as
\begin{equation}
    \theta_i^{\mathrm{rec}} := \sum_{m=1}^M \hat{\alpha}_{im}\theta_m^{\mathrm{ext}},
\end{equation}
and let
\begin{equation}
    D_\Theta := \mathrm{diam}_{\mathrm F}(\Theta), \qquad \xi_{\max} := \max_{m\in[M]} \|\theta_m^{\mathrm{ext}}-\theta_m^*\|_{\mathrm F}.
\end{equation}
By Proposition~\ref{prop:mixture_est},
\begin{equation}
    \|\hat{\alpha}_i-\alpha_i\|_1 \leq \eta_C.
\end{equation}

Under Assumption~\ref{ass:linear}, the quadratic form in~\eqnref{eq:loss_quadratic} and the barycentric identity $\theta_i^*=\sum_m\alpha_{im}\theta_m^*$ imply
\begin{equation}
    \mathcal{L}_i(\theta)-\mathcal{L}_i(\theta_i^*) = \|\theta-\theta_i^*\|_\Phi^2
\end{equation}
for every $\theta$. Therefore,
\begin{equation}
    \mathcal{L}_i(\theta_i^{\mathrm{rec}}) -\mathcal{L}_i(\theta_i^*) = \|\theta_i^{\mathrm{rec}}-\theta_i^*\|_\Phi^2 \leq \sigma_{\max}(\Phi) \|\theta_i^{\mathrm{rec}}-\theta_i^*\|_{\mathrm F}^2. \label{eq:static_rec_risk}
\end{equation}

It remains to bound the parameter-space reconstruction error. Using $\theta_i^*=\sum_m\alpha_{im}\theta_m^*$ and adding and subtracting $\sum_m\hat{\alpha}_{im}\theta_m^*$ gives
\begin{align}
    \theta_i^{\mathrm{rec}}-\theta_i^* &= \sum_{m=1}^M (\hat{\alpha}_{im}-\alpha_{im})\theta_m^* + \sum_{m=1}^M \hat{\alpha}_{im} (\theta_m^{\mathrm{ext}}-\theta_m^*). \label{eq:static_rec_decomp}
\end{align}
For the first term, since both $\hat{\alpha}_i$ and $\alpha_i$ belong to $\Delta^{M-1}$,
\[
\sum_{m=1}^M (\hat{\alpha}_{im}-\alpha_{im})=0.
\]
Fix any $m_0\in[M]$. Then
\begin{align}
    \left\| \sum_{m=1}^M (\hat{\alpha}_{im}-\alpha_{im})\theta_m^* \right\|_{\mathrm F} &= \left\| \sum_{m=1}^M (\hat{\alpha}_{im}-\alpha_{im}) (\theta_m^*-\theta_{m_0}^*) \right\|_{\mathrm F} \\
    &\leq \sum_{m=1}^M |\hat{\alpha}_{im}-\alpha_{im}|\, \|\theta_m^*-\theta_{m_0}^*\|_{\mathrm F} \\
    &\leq D_\Theta \|\hat{\alpha}_i-\alpha_i\|_1 \\
    &\leq \eta_C D_\Theta. \label{eq:static_rec_mixture}
\end{align}
For the second term, because $\hat{\alpha}_i\in\Delta^{M-1}$,
\begin{align}
    \left\| \sum_{m=1}^M \hat{\alpha}_{im} (\theta_m^{\mathrm{ext}}-\theta_m^*) \right\|_{\mathrm F} &\leq \sum_{m=1}^M \hat{\alpha}_{im} \|\theta_m^{\mathrm{ext}}-\theta_m^*\|_{\mathrm F} \\
    &\leq \xi_{\max}. \label{eq:static_rec_expert}
\end{align}
Combining~\eqnref{eq:static_rec_decomp}--\eqnref{eq:static_rec_expert} yields
\begin{equation}
    \|\theta_i^{\mathrm{rec}}-\theta_i^*\|_{\mathrm F} \leq \eta_C D_\Theta+\xi_{\max}.
\end{equation}
Define
\begin{equation}
    \Delta_i := \eta_C D_\Theta+\xi_{\max}. \label{eq:static_rec_delta}
\end{equation}
Substituting this bound into~\eqnref{eq:static_rec_risk} gives
\begin{equation}
    \mathcal{L}_i\!\left( \sum_{m=1}^M \hat{\alpha}_{im}\theta_m^{\mathrm{ext}} \right) - \mathcal{L}_i(\theta_i^*) \leq \sigma_{\max}(\Phi)\Delta_i^2.
\end{equation}

Hence, whenever $\eta_C\to0$ and $\xi_{\max}\to0$, we have $\Delta_i\to0$, and the reconstructed static adapter converges to the optimal static adapter in both parameter error and population excess risk. \qed

\subsection{Routing and Task-Aligned Specialization}
\label{app:routing}
We first formalize task-aligned specialization and then characterize the routing conditions required to approach it.

\begin{definition}[Task-aligned specialization]
\label{def:spec}
Fix a mixture of $M$ task distributions $\{\mathcal{P}_m\}_{m=1}^M$ whose true experts $\{\theta_m^*\}_{m=1}^M$ form a non-degenerate simplex ($h > 0$). A recovered system---a candidate expert set $\mathcal{V} = \{V_1, \ldots, V_M\}$ together with per-client routers $\{p_i(\cdot)\}_{i=1}^N$, produced by \emph{any} procedure---is \emph{$(\eta,\beta)$-specialized} if there exists a permutation $\pi \in S_M$ such that:
\begin{enumerate}

    \item[\textup{(i)}] \emph{Expert recovery (parameter space):} $\displaystyle \max_{m \in [M]} \|V_{\pi(m)}-\theta_m^*\|_\mathrm{F} \leq \eta$.
    
    \item[\textup{(ii)}] \emph{Routing concentration (function space):} for every client $i$ and every task $m$ in its support, $\displaystyle \mathbb{E}_{x \sim \mathcal{P}_m}\!\bigl[p_{i,\pi(m)}(x)\bigr] \geq 1 - \beta$.
\end{enumerate}
\end{definition}

The permutation $\pi$ accounts for the standard identifiability up to global relabeling that arises in mixture models and separable non-negative matrix factorization \citep{nmf}. Accordingly, Definition~\ref{def:spec} is invariant to a global permutation of the recovered experts. FedSEE additionally seeks task-indexed alignment, corresponding to $\pi=\mathrm{id}$. The task-identity result shows that label-free observations alone cannot determine this correspondence, whereas task-labeled recovery resolves it under the stated recovery conditions.

Condition~(ii) is defined at the input level rather than at the client level. Thus, a mixed client may route inputs from different tasks to different experts rather than being assigned as a whole to a single component. In particular, Condition~(ii) implies
\[
\|\bar p_i-\Pi_\pi\alpha_i\|_1 \leq 2\beta, \qquad \bar p_i := \mathbb{E}_{x\sim\mathcal D_i}[p_i(x)].
\]

\begin{remark}[Ideal specialization]
\label{rem:ideal_spec}
The task-indexed ideal corresponds to $(\eta,\beta)=(0,0)$ with $\pi=\mathrm{id}$. Under Definition~\ref{def:spec}, each task-$m$ input is then routed to $\theta_m^*$ and the predictor attains the oracle risk
\[
\Lambda_i := \sum_{m=1}^M \alpha_{im}\sigma_m^2 l.
\]
By the mixing-gap characterization established above, the best static adapter incurs an additional risk $G_i$, which is strictly positive for every genuinely mixed client and vanishes for a pure client.
\end{remark}

The results below separate two requirements for approaching this ideal. Mean expert usage must be calibrated to the client's mixture, while the routing policy must remain sufficiently input dependent to specialize across tasks.

\paragraph{Proof of Theorem~\ref{thm:routing_gap}.}
We separate the argument into the static contribution determined by $\bar p_i$ and the additional contribution induced by input-dependent routing.

\paragraph{Step 1: Risk of the constant mean router.}
For any $q\in\Delta^{M-1}$, let
\[
\theta(q):=\sum_{m=1}^M q_m\theta_m^\star.
\]
Under Assumption~\ref{ass:linear}, the quadratic client risk satisfies the variance identity
\[
\mathcal{L}_i(\theta)-\mathcal{L}_i(\theta_i^\star) = \|\theta-\theta_i^\star\|_\Phi^2,
\]
where
\[
\theta_i^\star = \sum_{m=1}^M\alpha_{im}\theta_m^\star
\]
by Theorem~\ref{thm:mixture}. Hence
\begin{equation}
\mathcal{L}_i(\theta(q))-\mathcal{L}_i(\theta_i^\star) = \left\| \sum_{m=1}^M(q_m-\alpha_{im})\theta_m^\star \right\|_\Phi^2. \label{eq:constant_router_exact}
\end{equation}

Set $c:=q-\alpha_i$. Since both $q$ and $\alpha_i$ lie in the simplex,
\[
\mathbf 1^\top c=0.
\]
By norm equivalence and Lemma~\ref{lem:centered},
\[
\left\| \sum_m c_m\theta_m^\star \right\|_\Phi^2 \ge \sigma_{\min}(\Phi) \left\| \sum_m c_m\theta_m^\star \right\|_F^2 \ge \sigma_{\min}(\Phi)h^2\|c\|_\infty^2.
\]
Therefore
\begin{equation}
\mathcal{L}_i(\theta(q))-\mathcal{L}_i(\theta_i^\star) \ge \sigma_{\min}(\Phi)h^2 \|q-\alpha_i\|_\infty^2. \label{eq:constant_router_lower}
\end{equation}

For the upper bound, let 
\[
s:=\frac12\|c\|_1.
\]
If $s=0$ the claim is immediate. Otherwise define 
\[
a_m:=\frac{(c_m)_+}{s}, \qquad b_m:=\frac{(-c_m)_+}{s}.
\]
Because $\mathbf 1^\top c=0$, both $a$ and $b$ belong to $\Delta^{M-1}$, and 
\[
\sum_m c_m\theta_m^\star = s \left( \sum_m a_m\theta_m^\star - \sum_m b_m\theta_m^\star \right).
\]
Both terms in parentheses lie in $\operatorname{conv}\{\theta_m^\star\}_{m=1}^M$. The diameter of this convex hull is $D_\Theta$, so
\[
\left\| \sum_m c_m\theta_m^\star \right\|_F \le sD_\Theta = \frac{D_\Theta}{2}\|c\|_1.
\]
Consequently,
\begin{equation}
\mathcal{L}_i(\theta(q))-\mathcal{L}_i(\theta_i^\star) \le \frac{\sigma_{\max}(\Phi)D_\Theta^2}{4} \|q-\alpha_i\|_1^2. \label{eq:constant_router_upper}
\end{equation}

\paragraph{Step 2: Reduction of an arbitrary router to its mean.}
Now consider an arbitrary routing policy $p_i(\cdot)$ and set
\[
q:=\bar p_i.
\]
For brevity write
\[
\theta_p(x):=\theta_{p_i}(x), \qquad \bar\theta_p:=\theta(\bar p_i).
\]

Let $T\in[M]$ denote the latent task of a sample from client $i$. Conditioned on $T=m$, Assumption~\ref{ass:linear} gives
\[
y=f_0(x)+\theta_m^\star\phi(x)+\zeta_m, \qquad \mathbb E[\zeta_m\mid x]=0.
\]
Since $p_i(x)$ is a function of the input and does not depend on $\zeta_m$, the noise cross-term vanishes conditionally on $x$. Therefore
\begin{align}
\mathcal{R}_i-\mathcal{L}_i(\bar\theta_p) = \mathbb E\Big[ & \|(\theta_T^\star-\theta_p(x))\phi(x)\|_2^2 \nonumber\\
&- \|(\theta_T^\star-\bar\theta_p)\phi(x)\|_2^2 \Big]. \label{eq:routing_reduce_exact}
\end{align}

For every $x$, both $\theta_p(x)$ and $\bar\theta_p$ lie in the convex hull of the experts. Hence
\[
\|\theta_T^\star-\theta_p(x)\|_F\le D_\Theta, \qquad \|\theta_T^\star-\bar\theta_p\|_F\le D_\Theta.
\]
Using $\|\phi(x)\|_2\le Q$, both corresponding signal residuals have Euclidean norm at most $QD_\Theta$. Thus
\begin{align}
\bigl|\mathcal{R}_i-\mathcal{L}_i(\bar\theta_p)\bigr| &\le 2QD_\Theta\, \mathbb E \left[ \|(\theta_p(x)-\bar\theta_p)\phi(x)\|_2 \right] \nonumber\\
&\le 2Q^2D_\Theta\, \mathbb E \left[ \|\theta_p(x)-\bar\theta_p\|_F \right]. \label{eq:routing_variation_param}
\end{align}

It remains to relate parameter variation to routing variation. For any two probability vectors $a,b\in\Delta^{M-1}$, the same positive--negative decomposition used above gives
\begin{equation}
\left\| \sum_m(a_m-b_m)\theta_m^\star \right\|_F \le \frac{D_\Theta}{2}\|a-b\|_1. \label{eq:simplex_lipschitz}
\end{equation}
Applying~\eqnref{eq:simplex_lipschitz} pointwise with $a=p_i(x)$ and $b=\bar p_i$ yields
\[
\|\theta_p(x)-\bar\theta_p\|_F \le \frac{D_\Theta}{2} \|p_i(x)-\bar p_i\|_1.
\]
Substituting into~\eqnref{eq:routing_variation_param},
\begin{equation}
\bigl|\mathcal{R}_i-\mathcal{L}_i(\theta(\bar p_i))\bigr| \le Q^2D_\Theta^2 V_i(p_i). \label{eq:routing_variation_bound}
\end{equation}

\paragraph{Step 3: Combine the two components.}
Using
\[
\mathcal{R}_i-\mathcal{L}_i(\theta_i^\star) = \bigl[ \mathcal{L}_i(\theta(\bar p_i)) -\mathcal{L}_i(\theta_i^\star) \bigr] + \bigl[ \mathcal{R}_i-\mathcal{L}_i(\theta(\bar p_i)) \bigr],
\]
the lower bound follows by combining \eqnref{eq:constant_router_lower} with the negative side of \eqnref{eq:routing_variation_bound}, and the upper bound follows by combining \eqnref{eq:constant_router_upper} with the positive side of \eqnref{eq:routing_variation_bound}. \qed

The preceding result characterizes the roles of mean calibration and routing variation relative to the optimal static adapter. We next give a complementary guarantee showing when recovered experts and task-aligned routing approach the routed oracle.

\begin{corollary}[Oracle risk from expert recovery and routing]
\label{cor:oracle_risk}
Under Assumptions~\ref{ass:linear} and~\ref{ass:bounded}, suppose
\[
\max_m\|\theta_m^{\mathrm{ext}}-\theta_m^*\|_\mathrm{F} \leq \xi_{\max},
\]
and, for every task $m$ in client $i$'s support, the router satisfies
\[
\mathbb{E}_{x\sim\mathcal{P}_m}[p_{im}(x)]\geq1-\beta .
\]
Let $D_\Theta:=\mathrm{diam}_\mathrm{F}(\Theta)$. The population risk $\mathcal{R}_i$ of the routed predictor~\eqnref{eq:objective} using the recovered experts satisfies
\begin{equation}
    \mathcal{R}_i-\Lambda_i \;\leq\; Q^2\!\left[ \xi_{\max}^2+ \bigl(2\xi_{\max}D_\Theta+D_\Theta^2\bigr)\beta \right]. \label{eq:oracle_specialization}
\end{equation}
Consequently, $\xi_{\max}\to0$ and $\beta\to0$ imply $\mathcal{R}_i\to\Lambda_i$.
\end{corollary}

\paragraph{Proof.}

For an input from task $m$, define $r_m(x):=1-p_{im}(x)$. By Assumption~\ref{ass:linear} and $\mathbb{E}[\zeta_m\mid x]=0$, subtracting the oracle noise risk gives
\begin{equation}
\mathcal{R}_i-\Lambda_i = \sum_m\alpha_{im}\, \mathbb{E}_{x\sim\mathcal{P}_m} \left[ \left\| \left( \sum_k p_{ik}(x)\theta_k^{\mathrm{ext}}-\theta_m^* \right)\phi(x) \right\|_2^2 \right].
\end{equation}
Using $\|\phi(x)\|_2\leq Q$,
\begin{equation}
\mathcal{R}_i-\Lambda_i \leq Q^2\sum_m\alpha_{im}\, \mathbb{E}_{x\sim\mathcal{P}_m} \left\| \sum_k p_{ik}(x)\theta_k^{\mathrm{ext}}-\theta_m^* \right\|_\mathrm{F}^2 .
\end{equation}

For task $m$,
\begin{align}
\left\| \sum_k p_{ik}(x)\theta_k^{\mathrm{ext}}-\theta_m^* \right\|_\mathrm{F} &\leq \left\| \sum_k p_{ik}(x)(\theta_k^{\mathrm{ext}}-\theta_k^*) \right\|_\mathrm{F} + \left\| \sum_{k\neq m}p_{ik}(x)(\theta_k^*-\theta_m^*) \right\|_\mathrm{F} \\
&\leq \xi_{\max}+D_\Theta r_m(x).
\end{align}
Since $0\leq r_m(x)\leq1$, we have $r_m(x)^2\leq r_m(x)$, while the routing-concentration assumption gives $\mathbb{E}_{\mathcal{P}_m}[r_m(x)]\leq\beta$. Therefore
\begin{equation}
\mathbb{E}_{\mathcal{P}_m} \left[\bigl(\xi_{\max}+D_\Theta r_m(x)\bigr)^2\right] \leq \xi_{\max}^2+ \bigl(2\xi_{\max}D_\Theta+D_\Theta^2\bigr)\beta .
\end{equation}
Averaging over $m$ with weights $\alpha_{im}$ and using $\sum_m\alpha_{im}=1$ yields~\eqnref{eq:oracle_specialization}. \qed

We next examine the consequence of imposing a fixed mean-usage target. Proposition~\ref{prop:balanced_specialization} in the main text shows that uniform mean usage conflicts with task-aligned specialization whenever the client mixture is non-uniform.

\paragraph{Proof of Proposition~\ref{prop:balanced_specialization}.}

Let
\[
q_{im} := \mathbb E_{x\sim\mathcal P_m}[p_i(x)] \in\Delta^{M-1}.
\]
If client $i$ satisfies Definition~\ref{def:spec}(ii) with permutation $\pi$, then
\[
q_{im,\pi(m)}\ge 1-\beta.
\]
Moreover,
\[
\bar p_i = \sum_m\alpha_{im}q_{im}, \qquad \Pi_\pi\alpha_i = \sum_m\alpha_{im}e_{\pi(m)}.
\]
Hence
\begin{align}
\|\bar p_i-\Pi_\pi\alpha_i\|_1 &\le \sum_m\alpha_{im} \|q_{im}-e_{\pi(m)}\|_1 \\
&= 2\sum_m\alpha_{im} \bigl(1-q_{im,\pi(m)}\bigr) \le 2\beta.
\end{align}
If additionally $\|\bar p_i-u\|_1\le\delta_{\rm bal}$, then the reverse triangle inequality gives
\[
2\beta \ge \|\bar p_i-\Pi_\pi\alpha_i\|_1 \ge \|\Pi_\pi\alpha_i-u\|_1 - \|\bar p_i-u\|_1.
\]
Since $u$ is permutation invariant,
\[
\|\Pi_\pi\alpha_i-u\|_1 = \|\alpha_i-u\|_1.
\]
Therefore
\[
\beta \ge \frac12 \bigl( \|\alpha_i-u\|_1-\delta_{\rm bal} \bigr)_+.
\]
For exact uniform mean usage, $\delta_{\rm bal}=0$, and hence
\[
\beta\ge\frac12\|\alpha_i-u\|_1.
\]
Thus $\beta>0$ for every non-uniform client. \qed

\begin{corollary}[Cost of anchoring to a fixed target]
\label{cor:anchor_cost}
Let $t\in\Delta^{M-1}$ and suppose $\|\bar p_i-t\|_\infty\le\chi$. Then
\begin{equation}
\mathcal{R}_i-\mathcal{L}_i(\theta_i^*) \ge \sigma_{\min}(\Phi)h^2 \bigl(\|\alpha_i-t\|_\infty-\chi\bigr)_+^2 -Q^2D_\Theta^2V_i(p_i). \label{eq:anchor_lb}
\end{equation}
\end{corollary}

\paragraph{Proof.} By the reverse triangle inequality,
\[
\|\bar p_i-\alpha_i\|_\infty \ge \bigl( \|\alpha_i-t\|_\infty -\|\bar p_i-t\|_\infty \bigr)_+ \ge \bigl(\|\alpha_i-t\|_\infty-\chi\bigr)_+.
\]
Substituting this inequality into the lower bound of Theorem~\ref{thm:routing_gap} yields~\eqnref{eq:anchor_lb}. \qed

\begin{corollary}[Personalization cost of balanced routing]
\label{cor:balanced_cost}
Taking $t=u=\tfrac1M\mathbf 1$, any policy satisfying $\|\bar p_i-u\|_\infty\le\chi$ obeys
\begin{equation}
\mathcal{R}_i-\mathcal{L}_i(\theta_i^*) \ge \sigma_{\min}(\Phi)h^2 \bigl(\|\alpha_i-u\|_\infty-\chi\bigr)_+^2 -Q^2D_\Theta^2V_i(p_i). \label{eq:balanced_lb}
\end{equation}
Hence the routed predictor remains separated from the optimal static adapter whenever the mean-mismatch term exceeds the routing-variation term.
\end{corollary}

\paragraph{Proof.}
Apply Corollary~\ref{cor:anchor_cost} with $t=u=\frac{1}{M}\mathbf 1$. \qed

These results show that a fixed uniform target can conflict with both task-aligned specialization and personalized performance. We next study a separate limitation concerning how the router representation itself can distinguish different client mixtures. We first record a standard Lipschitz property of the softmax map that will be used in the routing-sensitivity analysis.

\begin{lemma}[Softmax is $\tfrac{1}{2}$-Lipschitz in $\ell_2$]
\label{lem:softmax}
For all $a, b$, $\|\mathrm{softmax}(a) - \mathrm{softmax}(b)\|_2 \leq \tfrac{1}{2}\|a - b\|_2$.
\end{lemma}

\paragraph{Proof.} The Jacobian at logits $z$ with $p = \mathrm{softmax}(z)$ is
$J(z) = \mathrm{diag}(p) - pp^\top$. For a unit $u$, $u^\top J u = \mathrm{Var}_{e \sim p}(u_e) \leq \tfrac{1}{4}(\max_e u_e - \min_e u_e)^2 \leq \tfrac{1}{2}\|u\|_2^2$ by Popoviciu's inequality, so $\|J(z)\|_2 \leq \tfrac12$; integrate along the segment from $a$ to $b$. \qed

\paragraph{Sensitivity of routing statistics to task geometry.} To characterize when mean routing statistics can distinguish client
mixtures, consider the stylized router-input model
\begin{equation}
    x = \mu + \delta_m + \xi \qquad \text{under task } m,
\end{equation}
where $\mu$ is shared across tasks, $\delta_m$ is task specific, and $\xi$ is centered with a task-independent distribution. Let $\Pi_r$ project onto the top-$r$ principal subspace of the pooled second moment, and define
\[
\tau:=\max_{m\neq m'} \|\Pi_r(\delta_m-\delta_{m'})\|_2, \qquad \rho_\perp:=\max_{m\neq m'} \|(I-\Pi_r)(\delta_m-\delta_{m'})\|_2.
\]

\begin{proposition}[Routing-geometry sensitivity]
\label{prop:router}
Let $g(x) = W_r x + b_r$ be any linear router, $p(x) = \mathrm{softmax}(g(x))$, and $\bar{p}_m := \mathbb{E}[p(x) \mid \text{task } m]$. Then for all $m \neq m'$,
\begin{equation}
    \|\bar{p}_m - \bar{p}_{m'}\|_2 \leq \tfrac{1}{2}\bigl(\|W_r \Pi_r\|_2\,\tau + \|W_r(I - \Pi_r)\|_2\,\rho_\perp\bigr) =: D_{\mathrm{rt}}. \label{eq:router_bound}
\end{equation}
Moreover a client with mixture $\alpha$ has $\bar{p}(\alpha) = \sum_m \alpha_m \bar{p}_m$, and for any $\alpha, \alpha'$,
\begin{equation}
    \|\bar{p}(\alpha) - \bar{p}(\alpha')\|_2 \leq \|\alpha - \alpha'\|_1 \cdot D_{\mathrm{rt}}. \label{eq:mixture_bound}
\end{equation}
\end{proposition}

\paragraph{Proof.}
Since $\xi$ has the same distribution under every task, couple the task-$m$ and task-$m'$ inputs using the same realization of $\xi$:
\[
x_m=\mu+\delta_m+\xi, \qquad x_{m'}=\mu+\delta_{m'}+\xi.
\]
By Jensen's inequality and Lemma~\ref{lem:softmax}, (the bias $b_r$ cancels),
\[
\|\bar{p}_m - \bar{p}_{m'}\|_2 \leq \mathbb{E}_\xi \|\mathrm{softmax}(g(x_m)) - \mathrm{softmax}(g(x_{m'}))\|_2 \leq \tfrac12\|W_r(\delta_m - \delta_{m'})\|_2.
\]
\textbf{Subspace split.} Let $\Delta:=\delta_m-\delta_{m'}$. Decomposing $\Delta$ into its components inside and outside the principal subspace gives
\[
\|W_r\Delta\|_2 \leq \|W_r\Pi_r\|_2\|\Pi_r\Delta\|_2 + \|W_r(I-\Pi_r)\|_2\|(I-\Pi_r)\Delta\|_2 \leq \|W_r\Pi_r\|_2\tau + \|W_r(I-\Pi_r)\|_2\rho_\perp,
\]
proving~\eqnref{eq:router_bound}. For the client-level statement, linearity of expectation gives $\bar p(\alpha)=\sum_m\alpha_m\bar p_m$. Since $\sum_m(\alpha_m-\alpha_m')=0$,

\[
\|\bar{p}(\alpha) - \bar{p}(\alpha')\|_2 = \|\sum_m(\alpha_m - \alpha'_m)(\bar{p}_m - \bar{p}_1)\|_2 \leq \|\alpha - \alpha'\|_1 D_{\mathrm{rt}},
\]
which proves the result. \qed

The second inequality is a forward sensitivity bound. It shows that a change in the client mixture of size $\|\alpha-\alpha'\|_1$ changes the mean routing statistic by at most $D_{\rm rt}\|\alpha-\alpha'\|_1$. Consequently, when $D_{\rm rt}$ is small, substantially different mixtures may induce similar mean-routing statistics. This motivates obtaining mixture information from supervised expert recovery rather than inferring it solely from router statistics.

\paragraph{Mixture-aware calibration in the small-logit regime.}
We now relate the mean router response to the mixture-aware calibration target used by FedSEE. Fix a client $i$ and a router layer; let the router input have client mean $\bar{h}_i$, logits $z(h)=W_rh+b_r$, $p(h)=\mathrm{softmax}(z(h))$, $\bar{p}_i:=\mathbb{E}[p(h)]$, and $u:=\tfrac1M\mathbf{1}$. With $\Pi_M:=I_M-\tfrac1M\mathbf{1}\mathbf{1}^\top$, the router operates in the \emph{$\nu$-small-logit regime} if $\|\Pi_M z(h)\|_2\leq\nu$ a.s.

\begin{corollary}[Routing variation in the small-logit regime]
\label{cor:variation_small_logit}
If $p(h)=\mathrm{softmax}(z(h))$ satisfies $\|\Pi_M z(h)\|_2\le\nu$ almost surely, then
\[
V_i(p) := \mathbb E\|p(h)-\bar p_i\|_1 \le \frac{\sqrt M}{2}\,\nu.
\]
Consequently, Theorem~\ref{thm:routing_gap} has routing-variation slack at most
\[
\frac{\sqrt M}{2}\, Q^2D_\Theta^2\nu
\]
in this regime.
\end{corollary}

\paragraph{Proof.}
Softmax is invariant to adding the same constant to all logits, so
\[
p(h) = \mathrm{softmax}(\Pi_M z(h)).
\]
Since
\[
u = \mathrm{softmax}(0),
\]
Lemma~\ref{lem:softmax} gives, pointwise,
\[
\|p(h)-u\|_2 \le \frac12\|\Pi_M z(h)\|_2 \le \frac{\nu}{2}.
\]
Because $\bar p_i=\mathbb E[p(h)]$, the Euclidean variance identity yields
\[
\mathbb E\|p(h)-\bar p_i\|_2^2 = \mathbb E\|p(h)-u\|_2^2 - \|\bar p_i-u\|_2^2 \le \frac{\nu^2}{4}.
\]
Therefore, by Jensen's inequality and $\|v\|_1\le\sqrt M\|v\|_2$,
\[
V_i(p) = \mathbb E\|p(h)-\bar p_i\|_1 \le \sqrt M\, \mathbb E\|p(h)-\bar p_i\|_2 \le \frac{\sqrt M}{2}\nu.
\]
Substituting this bound into Theorem~\ref{thm:routing_gap} yields the stated routing-variation term. \qed

\begin{lemma}[Second-order softmax expansion]
\label{lem:expansion}
For all $z$, $\|\mathrm{softmax}(z)-u-\tfrac1M\Pi_Mz\|_2 \leq\tfrac34\|\Pi_Mz\|_2^2$. Hence in the $\nu$-small-logit regime,
\begin{equation}
    \bar{p}_i = u + \tfrac1M\Pi_M(W_r\bar{h}_i + b_r) + r_i, \qquad \|r_i\|_2 \leq \tfrac34\nu^2. \label{eq:meanrouting}
\end{equation}
\end{lemma}

\paragraph{Proof.} Let $\sigma(z) := \mathrm{softmax}(z)$, $J(z) = \mathrm{diag}(\sigma)
- \sigma\sigma^\top$. At the origin $\sigma(0) = u$, $J(0) = \tfrac1M\Pi_M$. Using $\|J(z) - J(z')\|_2 \leq 3\|p - p'\|_2 \leq \tfrac32\|z - z'\|_2$ (Lemma~\ref{lem:softmax}), $J$ is $\tfrac32$-Lipschitz. Applying Taylor's theorem to $\tilde z:=\Pi_Mz$ and using $\sigma(z)=\sigma(\tilde z)$ gives $\|\sigma(z)-u-J(0)\tilde z\|_2 \leq\tfrac34\|\tilde z\|_2^2$. Taking expectations over $h$ and using $\mathbb{E}[z(h)]=W_r\bar h_i+b_r$ gives~\eqnref{eq:meanrouting}. \qed

\begin{proposition}[KL-anchor calibration]
\label{prop:calibration}
In the $\nu$-small-logit regime: \textup{(a)} if $\mathrm{KL}(\tilde{c}_i \| \bar{p}_i) \leq \rho$ then
\[
\|\Pi_M(W_r\bar{h}_i+b_r)-M(\tilde{c}_i-u)\|_2 \le M(\sqrt{2\rho}+\tfrac34\nu^2);
\]
\textup{(b)} any router satisfying
\[
\Pi_M(W_r\bar{h}_i+b_r)=M(\tilde{c}_i-u)
\]
has
\[
\|\bar{p}_i-\tilde{c}_i\|_2\le\tfrac34\nu^2.
\]
\end{proposition}

\paragraph{Proof.}
For part~\textup{(a)}, Pinsker's inequality gives
\[
\|\tilde c_i-\bar p_i\|_2 \leq \|\tilde c_i-\bar p_i\|_1 \leq \sqrt{2\rho}.
\]
Rearranging~\eqnref{eq:meanrouting},
\[
\left\| \frac{1}{M}\Pi_M(W_r\bar h_i+b_r) - (\tilde c_i-u) \right\|_2 = \|\bar p_i-\tilde c_i-r_i\|_2.
\]
Using $\|r_i\|_2\leq\tfrac34\nu^2$ therefore yields
\[
\left\| \frac{1}{M}\Pi_M(W_r\bar h_i+b_r) - (\tilde c_i-u) \right\|_2 \leq \sqrt{2\rho}+\frac34\nu^2.
\]
Multiplying by $M$ proves part~\textup{(a)}.

For part~\textup{(b)}, substituting
\[
\Pi_M(W_r\bar h_i+b_r)=M(\tilde c_i-u)
\]
into~\eqnref{eq:meanrouting} gives
\[
\bar p_i-\tilde c_i=r_i.
\]
Hence
\[
\|\bar p_i-\tilde c_i\|_2 \leq \frac34\nu^2.
\]
\qed

\begin{remark}[Uniform versus mixture-aware calibration]
\label{rem:duality}
Under the simplified load-balancing objective $\|\bar p_i-u\|_2^2$ of \citet{wang2026myth}, the local expansion~\eqnref{eq:meanrouting} encourages the router's mean response toward the uniform target $u$. In contrast, FedSEE's KL term calibrates the same first moment toward the task-indexed target $\tilde c_i$. The uniform-usage result above shows that calibration toward $u$ is incompatible with exact specialization for a non-uniform client, whereas the result below quantifies the mean-routing error obtained from calibration toward $\tilde c_i$.
\end{remark}

\begin{corollary}[KL control of mean-routing mismatch]
\label{cor:kl_justify}
Let $\hat{\alpha}_i$ satisfy $\|\hat{\alpha}_i-\alpha_i\|_1\leq\eta_C$ and define $\delta_i^{\mathrm{proxy}} :=\|\tilde c_i-\hat{\alpha}_i\|_1$. If $\mathrm{KL}(\tilde c_i\|\bar p_i)\leq\rho$, then
\begin{equation}
\|\bar p_i-\alpha_i\|_1 \le \sqrt{2\rho}+\eta_C+\delta_i^{\mathrm{proxy}}. \label{eq:proxy_routing_bound}
\end{equation}
Under the assumptions of Theorem~\ref{thm:routing_gap} and when routing over the true experts, this further implies
\begin{equation}
\mathcal{R}_i-\mathcal{L}_i(\theta_i^*) \le \frac{\sigma_{\max}(\Phi)D_\Theta^2}{4} \left( \sqrt{2\rho}+\eta_C+\delta_i^{\mathrm{proxy}} \right)^2 + Q^2D_\Theta^2V_i(p_i). \label{eq:kl_ub}
\end{equation}
\end{corollary}

\paragraph{Proof.}
Pinsker's inequality gives
\[
\|\tilde c_i-\bar p_i\|_1 \le \sqrt{2\,\mathrm{KL}(\tilde c_i\|\bar p_i)} \le \sqrt{2\rho}.
\]
By the definition of $\delta_i^{\mathrm{proxy}}$ and Proposition~\ref{prop:mixture_est},
\[
\|\tilde c_i-\alpha_i\|_1 \le \|\tilde c_i-\hat\alpha_i\|_1 + \|\hat\alpha_i-\alpha_i\|_1 \le \delta_i^{\mathrm{proxy}}+\eta_C.
\]
Hence
\[
\|\bar p_i-\alpha_i\|_1 \le \|\bar p_i-\tilde c_i\|_1 + \|\tilde c_i-\alpha_i\|_1 \le \sqrt{2\rho} +\eta_C +\delta_i^{\mathrm{proxy}}.
\]
Substituting this bound into the upper inequality of Theorem~\ref{thm:routing_gap} gives~\eqnref{eq:kl_ub}. \qed

The bound separates three sources of mean-routing error: the KL calibration error $\rho$, the certified barycentric mixture-estimation error $\eta_C$, and the proxy error $\delta_i^{\mathrm{proxy}}$. The first two vanish under the corresponding optimization and recovery limits, while the third quantifies the additional approximation introduced by using the contribution-based proxy $\tilde c_i$ in place of the certified barycentric estimate.

\paragraph{Scope of the routing guarantee.}
Theorem~\ref{thm:routing_gap} separates two effects in routed risk. The mean-mismatch term quantifies calibration relative to the client mixture, while $V_i(p_i)$ controls how far input-dependent routing can move the risk from the corresponding static adapter. The latter term has no prescribed sign in the exact decomposition. Input-dependent routing may therefore improve upon the static floor and is precisely the mechanism needed to close the mixing gap.

FedSEE's KL anchor controls the mean-calibration component through Corollary~\ref{cor:kl_justify}. Reaching the routed oracle additionally requires accurate expert recovery and task-aligned input-level routing. Corollary~\ref{cor:oracle_risk} shows that as the routing concentration error $\beta$ and expert-recovery error $\xi_{\max}$ vanish, the routed risk approaches the oracle risk. Thus the KL anchor calibrates mean usage, while the task loss remains responsible for input-level specialization.

\paragraph{Shared-router aggregation.}
Condition~(ii) of Definition~\ref{def:spec} is task-conditioned rather than client-specific. Consequently, a single router can satisfy the specialization condition for all clients simultaneously. In particular, suppose $p^\star$ satisfies
\[
\mathbb{E}_{x\sim\mathcal{P}_m} [p^\star_m(x)] \geq 1-\beta
\]
for every task $m$. Then, for client $i$,
\[
\bar p_i = \sum_m \alpha_{im} \mathbb{E}_{x\sim\mathcal{P}_m}[p^\star(x)],
\]
and therefore
\[
\|\bar p_i-\alpha_i\|_1 \leq 2\beta.
\]
Thus, in the ideal task-aligned regime of Remark~\ref{rem:ideal_spec}, separate client-specific routers are not required to satisfy the routing-concentration condition.

A shared router can nevertheless be difficult to calibrate across clients when task information is weak in its input representation. Proposition~\ref{prop:router} bounds the separation between task-conditioned routing statistics, so substantially different client mixtures may induce similar mean expert usage. In this regime, Theorem~\ref{thm:routing_gap} identifies the resulting mean-calibration mismatch, while its variation term $V_i(p_i)$ captures the additional effect of input-dependent routing. The theorem therefore does not imply that router aggregation necessarily incurs a positive excess-risk penalty.

\subsection{Extensions Beyond the Exact Squared-Loss Setting}
\label{app:general_loss}
The preceding results exploit the exact quadratic geometry induced by the squared-loss model, while the experiments in Section~\ref{sec:experiments} optimize softmax cross-entropy. We therefore extend the analysis to general losses. The client risk remains an exact mixture of task risks, while the client optimum becomes a perturbed barycenter controlled by the curvature-dispersion parameter $\varkappa$. We then propagate this perturbation to expert recovery, specialize the analysis to softmax cross-entropy over a frozen representation, and discuss the corresponding implications for mixture estimation, routing, and identifiability.

\subsubsection{Setup and Assumptions}
\label{app:gl_setup}

We retain the model class defined in~\eqnref{eq:model_class}, $f_\theta(x) = f_0(x) + \theta\,\phi(x)$, where $(f_0,\phi)$ is frozen and $\theta \in \mathbb{R}^{l \times r}$ remains the sole trainable parameter block. We replace the squared error with a general loss function $\ell(f_\theta(x), y)$. Let $\mathcal{L}_m(\theta) := \mathbb{E}_{\mathcal{P}_m}[\ell(f_\theta(x),y)]$ denote the population risk for task $m$, and let $\Theta := \{\theta_m^*\}_{m=1}^M$ represent the set of true experts for $M \geq 2$. Since $\mathcal{D}_m^s$ contains i.i.d.\ samples from $\mathcal{P}_m$ (Assumption~\ref{ass:public}), $\widehat{\mathcal{L}}_m^s$ is the empirical counterpart of $\mathcal{L}_m$.

For analytical convenience, we vectorize the parameters: let $\vartheta := \mathrm{vec}(\theta) \in \mathbb{R}^{lr}$, such that $\nabla^2 \mathcal{L}_m(\theta) \in \mathbb{R}^{lr \times lr}$ and $\|\theta\|_\mathrm{F} = \|\vartheta\|_2$. We write $\mathcal{L}_m(\vartheta)$ and $\mathcal{L}_m(\theta)$ interchangeably. All Hessians in this section are defined with respect to this vectorization, and all matrix norms denote the spectral norm. When the loss is invariant along a subspace of the parameter block, as occurs for softmax cross-entropy under common shifts of all logits, the strong-convexity statements below are understood on the corresponding invariant complement $\mathcal{V}$. Appendix~\ref{app:gl_ce} makes this restriction explicit for the cross-entropy objective.
\begin{assumption}[Regular per-task risk]
\label{ass:regular_loss}
Let $\bar{\mathcal{C}} \subseteq \mathbb{R}^{l \times r}$ be a closed and convex set with interior $\mathcal{C} := \operatorname{int}\bar{\mathcal{C}}$. Assume (i) for every task $m$ and every client $i$, the unique minimizers $\theta_m^* := \arg\min_{\theta \in \bar{\mathcal{C}}} \mathcal{L}_m(\theta)$ and $\theta_i^* := \arg\min_{\theta \in \bar{\mathcal{C}}} \mathcal{L}_i(\theta)$ exist and lie strictly within $\mathcal{C}$, ensuring that $\nabla\mathcal{L}_m(\theta_m^*) = 0$ and $\nabla\mathcal{L}_i(\theta_i^*) = 0$; and (ii) on $\mathcal{C}$, each $\mathcal{L}_m$ is twice continuously differentiable and satisfies, for all $\theta, \theta' \in \mathcal{C}$,
\begin{equation}
    \mu I \;\preceq\; \nabla^2 \mathcal{L}_m(\theta) \;\preceq\; L I, \qquad \bigl\|\nabla^2 \mathcal{L}_m(\theta) - \nabla^2 \mathcal{L}_m(\theta')\bigr\| \;\leq\; L_H\,\|\theta - \theta'\|_\mathrm{F}, \label{eq:gl_regular}
\end{equation}
with $0 < \mu \leq L < \infty$ and $L_H \geq 0$.
\end{assumption}

\begin{definition}[Curvature dispersion]
\label{def:curvature_dispersion}
The curvature dispersion of the task family and its relative counterpart are defined as:
\begin{equation}
    \Delta_H \;:=\; \max_{m \neq m'}\;\sup_{\theta \in \mathcal{C}}\; \bigl\|\nabla^2\mathcal{L}_m(\theta) - \nabla^2\mathcal{L}_{m'}(\theta)\bigr\|, \qquad \varkappa \;:=\; \frac{\Delta_H + \tfrac{1}{2}L_H\,\mathrm{diam}_\mathrm{F}(\Theta)}{\mu}. \label{eq:gl_kappa}
\end{equation}
\end{definition}

Two observations contextualize the curvature dispersion $\varkappa$. First, $\Delta_H$ evaluates task divergence at a shared parameter; it characterizes the structural mismatch of the task landscapes rather than the spatial distance between their optima. Second, under the squared loss $\ell(\theta;x,y) = \|y - f_\theta(x)\|_2^2$, the Hessian evaluates to $\nabla^2\mathcal{L}_m(\theta) = 2\,\Phi_m \otimes I_l$ globally. Consequently, if tasks share an input covariance ($\Phi_m = \Phi$), this yields $\Delta_H = 0$ and $L_H = 0$, resulting in $\varkappa = 0$. Even under a heterogeneous squared loss ($\Phi_m \neq \Phi$), we have $\Delta_H = 2\max_{m \neq m'}\|\Phi_m - \Phi_{m'}\|_2$, $L_H = 0$, and $\mu = 2\min_m \sigma_{\min}(\Phi_m)$, such that $\varkappa = \rho_\Phi$. Thus, $\varkappa$ exactly reproduces the covariance-dispersion bias $\rho_\Phi$ from Remark~\ref{rem:heterogeneous} in Appendix~\ref{app:geometry}, directly confirming the internal consistency of the two analyses.

\subsubsection{General-Loss Geometry and Expert Recovery}
\label{app:gl_risk}
\label{app:gl_bary}
\label{app:gl_recovery}

\begin{proposition}[Mixture risk decomposition]
\label{prop:mixture_risk}
For any measurable loss function $\ell$ and any $\theta$ for which the expectations exist,
\begin{equation}
    \mathcal{L}_i(\theta) \;=\; \sum_{m=1}^{M} \alpha_{im}\,\mathcal{L}_m(\theta).
\end{equation}
\end{proposition}

\paragraph{Proof.}
This follows directly from~\eqnref{eq:mixture} and the linearity of expectation: the client distribution $\mathcal{D}_i$ is identically the mixture $\sum_m \alpha_{im}\mathcal{P}_m$, and the loss $\ell$ operates independently of the client index $i$. \qed

The risk decomposition therefore survives exactly beyond squared loss. What changes is the geometry of its minimizer. In general, $\theta_i^*$ need not coincide with the barycenter of the task optima, but the deviation can be controlled by the curvature dispersion introduced above.

\begin{theorem}[Perturbed barycenter under a general loss]
\label{thm:perturbed_bary}
Let Assumption~\ref{ass:regular_loss} hold. Define the exact barycenter $\theta_i^{\mathrm{bary}} := \sum_{m} \alpha_{im}\theta_m^*$ and the local radius $D_i := \max_{m}\|\theta_m^* - \theta_i^{\mathrm{bary}}\|_\mathrm{F} \leq \mathrm{diam}_\mathrm{F}(\Theta)$. The client population optimum $\theta_i^*$ satisfies:
\begin{equation}
    \bigl\|\theta_i^* - \theta_i^{\mathrm{bary}}\bigr\|_\mathrm{F} \;\leq\; \varkappa \, D_i \;\leq\; \varkappa\,\mathrm{diam}_\mathrm{F}(\Theta). \label{eq:gl_bary_bound}
\end{equation}
This bound is non-vacuous in the regime where $\varkappa < 1$. Consequently, $\mathrm{dist}_\mathrm{F}\bigl(\theta_i^*,\,\mathrm{conv}\{\theta_m^*\}\bigr) \leq \varkappa\,\mathrm{diam}_\mathrm{F}(\Theta)$, and when $\varkappa = 0$, the identity $\theta_i^* = \sum_m \alpha_{im}\theta_m^*$ of Theorem~\ref{thm:mixture} is recovered exactly.
\end{theorem}

\paragraph{Proof.}
By Proposition~\ref{prop:mixture_risk} and Assumption~\ref{ass:regular_loss}, the aggregate risk $\mathcal{L}_i = \sum_m \alpha_{im}\mathcal{L}_m$ is $\mu$-strongly convex on $\mathcal{C}$.

\emph{Step 1 (Integral representation).} Because interiority ensures $\nabla\mathcal{L}_m(\theta_m^*) = 0$, we apply the fundamental theorem of calculus to $\nabla\mathcal{L}_m$ along the segment $[\theta_m^*, \theta_i^*] \subset \mathcal{C}$. Utilizing the vectorized notation $\vartheta = \mathrm{vec}(\theta)$, we obtain:
\begin{equation}
    \nabla\mathcal{L}_m(\vartheta_i^*) \;=\; G_m\,(\vartheta_i^* - \vartheta_m^*), \qquad G_m \;:=\; \int_0^1 \nabla^2\mathcal{L}_m\bigl(\vartheta_m^* + t(\vartheta_i^* - \vartheta_m^*)\bigr)\,\mathrm{d}t . \label{eq:gl_Gm}
\end{equation}
Because each $G_m$ is an integrated average of Hessians evaluated within $\mathcal{C}$, it satisfies $\mu I \preceq G_m \preceq L I$. This representation remains exact.

\emph{Step 2 (First-order condition).} The optimality of $\theta_i^*$ for $\mathcal{L}_i$ dictates $\sum_m \alpha_{im}\nabla\mathcal{L}_m(\vartheta_i^*) = 0$. Substituting~\eqnref{eq:gl_Gm} yields $\sum_m \alpha_{im} G_m (\vartheta_i^* - \vartheta_m^*) = 0$. Defining $\bar{G} := \sum_m \alpha_{im}G_m \succeq \mu I$ and rearranging terms:
\begin{equation}
    \bar{G}\,\vartheta_i^* \;=\; \sum_m \alpha_{im} G_m \vartheta_m^* . \label{eq:gl_foc}
\end{equation}

\emph{Step 3 (Barycentric alignment).} Noting that $\sum_m \alpha_{im}(G_m - \bar{G}) = 0$, we subtract $\bar{G}\,\vartheta_i^{\mathrm{bary}}$ from both sides of~\eqnref{eq:gl_foc} to isolate the deviation:
\begin{equation}
    \bar{G}\bigl(\vartheta_i^* - \vartheta_i^{\mathrm{bary}}\bigr) \;=\; \sum_m \alpha_{im}\,(G_m - \bar{G})\,\vartheta_m^* \;=\; \sum_m \alpha_{im}\,(G_m - \bar{G})\,\bigl(\vartheta_m^* - \vartheta_i^{\mathrm{bary}}\bigr). \label{eq:gl_key}
\end{equation}

\emph{Step 4 (Bounding curvature variation).} Fix indices $m, k$. The integrands defining $G_m$ and $G_k$ evaluate to a parameter-space displacement of $(1-t)(\vartheta_m^* - \vartheta_k^*)$. Decoupling this into a task difference at a shared point and the integrated displacement yields:
\begin{equation}
    \|G_m - G_k\| \;\leq\; \Delta_H \;+\; L_H \int_0^1 (1-t)\,\|\vartheta_m^* - \vartheta_k^*\|_2\,\mathrm{d}t \;\leq\; \Delta_H + \tfrac{1}{2}L_H\,\mathrm{diam}_\mathrm{F}(\Theta).
\end{equation}
Because $\bar{G}$ is a convex combination of the matrices $G_k$, convexity of the spectral norm ensures this bound applies uniformly to $\|G_m - \bar{G}\|$.

\emph{Step 5 (Conclusion).} Taking the Euclidean norm $\|\cdot\|_2$ in~\eqnref{eq:gl_key} and utilizing $\|\bar{G}^{-1}\| \leq 1/\mu$, we find:
\begin{equation}
    \bigl\|\vartheta_i^* - \vartheta_i^{\mathrm{bary}}\bigr\|_2 \;\leq\; \frac{1}{\mu}\, \Bigl(\Delta_H + \tfrac{1}{2}L_H\,\mathrm{diam}_\mathrm{F}(\Theta)\Bigr) \sum_m \alpha_{im}\bigl\|\vartheta_m^* - \vartheta_i^{\mathrm{bary}}\bigr\|_2 \;\leq\; \varkappa\,D_i .
\end{equation}
Translating the Euclidean norm back to the equivalent Frobenius norm over the matrices satisfies~\eqnref{eq:gl_bary_bound}. \qed

\begin{remark}[Structural boundaries of exactness]
\label{rem:gl_lost}
\Eqnref{eq:gl_key} shows that exactness $\theta_i^* = \theta_i^{\mathrm{bary}}$ holds precisely when the directional condition $\sum_m \alpha_{im}\,(G_m - \bar{G})\,(\vartheta_m^* - \vartheta_i^{\mathrm{bary}}) = 0$ is met; for $M = 2$ this reduces to $(G_1 - G_2)(\vartheta_1^* - \vartheta_2^*) = 0$. This is a condition on the \emph{averaged} operators $G_m$ along a single direction, and is strictly weaker than $\Delta_H = 0$: non-quadratic families with uniformly separated Hessians can still be exactly barycentric for every $\alpha \in \Delta^{M-1}$. The bound~\eqnref{eq:gl_bary_bound} is therefore sufficient rather than necessary---it is the tightest control available from curvature magnitudes alone, and $\varkappa$ should be read as an upper bound on the deviation, not as a measure of it.
\end{remark}

\begin{remark}[Identifiability under general losses]
\label{rem:gl_identifiability}
The task-identity indeterminacy of Theorem~\ref{thm:label} follows from permutation symmetry and is therefore loss agnostic. In contrast, the vertex-set non-identifiability construction of Theorem~\ref{thm:vertex} uses the exact quadratic barycentric geometry. Under general losses, Theorem~\ref{thm:perturbed_bary} provides only an approximate simplex geometry controlled by $\varkappa$. Establishing a corresponding approximate vertex-set indistinguishability result would therefore require an additional perturbation argument and is not claimed here.
\end{remark}

The same perturbation also affects supervised expert extraction. For an $\varepsilon$-anchor client, the distance to its target task expert now has two sources: anchor impurity and curvature-induced deviation from the exact barycenter. This yields the following general-loss recovery guarantee.

\begin{theorem}[Expert recovery, general loss]
\label{thm:recovery_general}
Let Assumptions~\ref{ass:regular_loss} and~\ref{ass:anchor} hold, with exact client uploads $\{\theta_i^*\}$. The extraction program~\eqnref{eq:gen_extract} returns:
\begin{equation}
    \bigl\|\theta_m^{\mathrm{ext}} - \theta_m^*\bigr\|_\mathrm{F} \;\leq\; \sqrt{L/\mu}\;\cdot\;(\varepsilon + \varkappa)\;\cdot\; \mathrm{diam}_\mathrm{F}(\Theta).
\end{equation}
\end{theorem}

\paragraph{Proof.}
Let $i_m$ designate the exact anchor client for task $m$. By the triangle inequality and Theorem~\ref{thm:perturbed_bary},
\begin{align}
    \|\theta_{i_m}^* - \theta_m^*\|_\mathrm{F} &\;\leq\; \|\theta_{i_m}^{\mathrm{bary}} - \theta_m^*\|_\mathrm{F} + \varkappa\,\mathrm{diam}_\mathrm{F}(\Theta) \notag\\
    &\;\leq\; \Bigl\|\sum_{m' \neq m} \alpha_{i_m, m'}(\theta_{m'}^* - \theta_m^*)\Bigr\|_\mathrm{F} + \varkappa\,\mathrm{diam}_\mathrm{F}(\Theta) \;\leq\; (\varepsilon + \varkappa)\,\mathrm{diam}_\mathrm{F}(\Theta). \label{eq:gl_anchor}
\end{align}

The simplex vertex allocating all probability mass to $\theta_{i_m}^*$ is feasible for the program~\eqnref{eq:gen_extract}, ensuring $\mathcal{L}_m(\theta_m^{\mathrm{ext}}) \leq \mathcal{L}_m(\theta_{i_m}^*)$. Utilizing the $L$-smoothness of $\mathcal{L}_m$ at its minimizer $\theta_m^*$ and its $\mu$-strong convexity, we obtain:
\begin{equation}
    \tfrac{\mu}{2}\|\theta_m^{\mathrm{ext}} - \theta_m^*\|_\mathrm{F}^2 \;\leq\; \mathcal{L}_m(\theta_m^{\mathrm{ext}}) - \mathcal{L}_m(\theta_m^*) \;\leq\; \mathcal{L}_m(\theta_{i_m}^*) - \mathcal{L}_m(\theta_m^*) \;\leq\; \tfrac{L}{2}\|\theta_{i_m}^* - \theta_m^*\|_\mathrm{F}^2 .
\end{equation}
Substituting the anchor distance and taking the square root completes the proof. \qed

\begin{remark}[Other downstream implications]
\label{rem:gl_downstream}
Beyond expert recovery, the perturbation in Theorem~\ref{thm:perturbed_bary} also affects downstream quantities that rely on the exact barycentric geometry. For mixture estimation, if $\|\hat{\theta}_i-\theta_i^*\|_{\mathrm F}\leq\delta_{\mathrm{loc}}$, then
\[
\bigl\|\hat{\theta}_i-\theta_i^{\mathrm{bary}}\bigr\|_{\mathrm F} \leq \delta_{\mathrm{loc}} + \varkappa\,\mathrm{diam}_{\mathrm F}(\Theta),
\]
so the curvature-induced deviation enters as an additive upload error. The routing results of Appendix~\ref{app:routing} rely more directly on the quadratic risk geometry and therefore do not transfer verbatim. Extending them to general losses would require controlling the same curvature-induced deviation together with the strong-convexity and smoothness constants. When $\varkappa=0$, the exact barycentric geometry of the squared-loss analysis is recovered.
\end{remark}

The preceding results reduce the general-loss extension to controlling the curvature and conditioning quantities $\mu$, $L$, $\Delta_H$, and $L_H$. We next instantiate these quantities for softmax cross-entropy over the frozen-feature model; the composed-transformer approximation is treated separately in Appendix~\ref{app:fedsee_reduction}.

\subsubsection{Softmax Cross-Entropy over a Frozen Representation}
\label{app:gl_ce}
We specialize the general-loss analysis to softmax cross-entropy over the frozen representation in~\eqnref{eq:model_class}. Assume $l$ classes and let
\[
z_\theta(x)=f_0(x)+\theta\phi(x)\in\mathbb{R}^l, \qquad \ell(\theta;x,y) = -\log\bigl[\operatorname{softmax}(z_\theta(x))\bigr]_y .
\]
Writing $p_\theta(x)=\operatorname{softmax}(z_\theta(x))$, define the softmax covariance
\[
S(z):=\operatorname{diag}(p)-pp^\top .
\]
Because the softmax is invariant to adding the same scalar to every logit, we use the centered parameterization $\mathbf{1}^\top\theta=0$. The corresponding invariant complement is $\mathcal{V}=\mathbb{R}^r\otimes\mathbf{1}^\perp$, and all curvature bounds below are understood on this subspace.

\begin{proposition}[Cross-entropy curvature over a frozen representation]
\label{prop:ce_constants}
For every $\theta$, the Hessian evaluates to:
\begin{equation}
    \nabla^2 \mathcal{L}_m(\theta) \;=\; \mathcal{H}_m(\theta) \;:=\; \mathbb{E}_{\mathcal{P}_m}\Bigl[\,\phi(x)\phi(x)^\top \otimes S\bigl(z_\theta(x)\bigr)\Bigr]. \label{eq:ce_hessian}
\end{equation}
Suppose Assumption~\ref{ass:bounded} holds and each taskwise feature second moment
$\Phi_m:=\mathbb{E}_{\mathcal{P}_m}[\phi(x)\phi(x)^\top]$ is positive definite. Define
$\lambda_-:=\min_m\sigma_{\min}(\Phi_m)$ and
$\lambda_+:=\max_m\sigma_{\max}(\Phi_m)$.
Suppose further that $\mathcal{C} \subseteq \{\|\theta\|_{\mathrm{F}}\leq\theta_{\max}\}$ and let $B_0:=\sup_x\|f_0(x)\|_\infty$. Then
$\|z_\theta(x)\|_\infty\leq B_0+\theta_{\max}Q$, so every class probability satisfies
\[
\pi_{\min}\geq\frac{1}{l}\exp\bigl(-2(B_0+\theta_{\max}Q)\bigr).
\]
On the centered subspace $\mathcal{V}$,
\begin{equation}
l\,\pi_{\min}^2\lambda_- I
\;\preceq\;
\mathcal{H}_m(\theta)
\;\preceq\;
\tfrac{1}{2}\lambda_+ I,
\qquad
\frac{L}{\mu}
\leq
\frac{\lambda_+}{2l\pi_{\min}^2\lambda_-}.
\label{eq:ce_mu_L}
\end{equation}
Furthermore, the curvature dispersion $\Delta_H$ satisfies:
\begin{equation}
    \Delta_H \;=\; \max_{m \neq m'}\;\sup_{\theta \in \mathcal{C}}\; \bigl\|\mathcal{H}_m(\theta) - \mathcal{H}_{m'}(\theta)\bigr\| \;\leq\; Q^2 \cdot \max_{m \neq m'} \mathrm{TV}\bigl(\mathcal{P}_m^x, \mathcal{P}_{m'}^x\bigr), \label{eq:ce_delta}
\end{equation}
where $\mathcal{P}_m^x$ denotes the input marginal of task $m$. The Hessian-Lipschitz constant, measured in spectral norm against a Frobenius parameter displacement, satisfies $L_H \leq 2Q^3$.
\end{proposition}

\paragraph{Proof.}
\emph{Hessian.} Defining $z = f_0(x) + \theta\phi(x)$ and $\vartheta = \mathrm{vec}(\theta)$ yields $\partial z/\partial\vartheta = \phi(x)^\top \otimes I_l$. Because the Hessian of the softmax cross-entropy with respect to its logits is $S(z)$, the chain rule yields $\nabla^2_\vartheta \ell = (\phi\phi^\top) \otimes S(z)$ pointwise. Taking expectations recovers~\eqnref{eq:ce_hessian}.

\emph{Upper and Lower Bounds.} Because $S(z)\mathbf{1} = 0$, we define the orthogonal projector $P := I_l - \frac{1}{l}\mathbf{1}\mathbf{1}^\top$ onto $\mathbf{1}^\perp$. For any unit vector $v$, shift-invariance of the variance allows us to assume $v$ is centered ($v^\top \mathbf{1} = 0$). The geometric constraints $\max v \ge 0 \ge \min v$ and $(\max v)^2 + (\min v)^2 \le \|v\|_2^2 = 1$ guarantee $\max_c v_c - \min_c v_c \le \sqrt{2}$. Popoviciu's inequality therefore dictates $\mathrm{Var}_{c \sim p}(v_c) \le \frac{1}{4}(\max - \min)^2 \le \frac{1}{2}$, ensuring $S(z) \preceq \tfrac12 I_l$. Leveraging the logit bound $p_c \geq \pi_{\min}$, we simultaneously derive the lower bound: $\mathrm{Var}_{c \sim p}(v_c) = \tfrac12\sum_{c,d} p_c p_d (v_c-v_d)^2 \geq \tfrac{\pi_{\min}^2}{2} \sum_{c,d} (v_c-v_d)^2 = l\pi_{\min}^2 \|v\|_2^2$. This establishes $S(z) \succeq l\pi_{\min}^2 P$. Because $\phi(x)\phi(x)^\top \succeq 0$, tensoring preserves both orderings pointwise. Taking expectations gives
$\mathcal{H}_m(\theta)\succeq l\pi_{\min}^2(\Phi_m\otimes P)$ and
$\mathcal{H}_m(\theta)\preceq \tfrac12(\Phi_m\otimes I_l)$.
Restricting to $\mathcal{V}=\mathbb{R}^r\otimes\mathbf{1}^\perp$ and using the definitions of $\lambda_-$ and $\lambda_+$ yields~\eqnref{eq:ce_mu_L}.

\emph{Dispersion.} Evaluated at a common parameter $\theta$, the integrand $\Psi_\theta(x) := \phi(x)\phi(x)^\top \otimes S(z_\theta(x))$ constitutes a fixed function bounded by $\|\Psi_\theta(x)\| \leq \tfrac12 Q^2$. For bounded functions $\Psi$ and distributions $P_1, P_2$, the relation $\|\mathbb{E}_{P_1} \Psi - \mathbb{E}_{P_2} \Psi\| \leq 2\sup_x\|\Psi(x)\|\,\mathrm{TV}(P_1, P_2)$ under the convention $\mathrm{TV}(P_1, P_2) = \sup_A |P_1(A)-P_2(A)|$ guarantees~\eqnref{eq:ce_delta}.

\emph{Hessian-Lipschitz.} Using the Jacobian bound from the proof of Lemma~\ref{lem:expansion}, $\|S(z)-S(z')\|_2\leq\tfrac32\|z-z'\|_2$, and hence $\|\mathcal{H}_m(\theta)-\mathcal{H}_m(\theta')\|_2 \leq\tfrac32 Q^3\|\theta-\theta'\|_{\mathrm F} \leq2Q^3\|\theta-\theta'\|_{\mathrm F}$. \qed

\begin{remark}[Interpretation of the cross-entropy constants]
\label{rem:ce_reading}
The lower curvature bound deteriorates as the logit range increases, since $\mu \geq l\,\pi_{\min}^2\lambda_-$  and $\pi_{\min}$ decreases exponentially with $B_0+\theta_{\max}Q$. Consequently, the resulting upper bounds on the condition number $L/\mu$ and the perturbation parameter $\varkappa$ can become loose when the model is highly confident. The analysis is therefore most informative in a bounded-logit regime, enforced here through the parameter set $\mathcal{C}$ and the bound $\|\theta\|_{\mathrm F}\leq\theta_{\max}$. Under these conditions, softmax cross-entropy satisfies the local strong-convexity, smoothness, and Hessian-regularity requirements used by the preceding general-loss results.
\end{remark}

\subsection{Additional Model-Class Instantiations}
\label{app:instantiations}

The preceding analysis extends the framework along the loss dimension. A complementary question is how the shared-representation abstraction applies beyond the LoRA instantiation used by FedSEE. For the squared-loss results of Section~\ref{sec:theory}, the relevant quantities are determined jointly by the induced representation and the resulting expert geometry:
\begin{itemize}
    \item $\sigma_{\min}(\Phi)$ and $\sigma_{\max}(\Phi)$ characterize the conditioning of the shared feature geometry;
    \item $h$ and $\mathrm{diam}_\mathrm{F}(\Theta)$ characterize the geometry of the task-expert simplex and enter the recovery and routing guarantees;
    \item $Q$, the feature bound of Assumption~\ref{ass:bounded}, controls the finite-sample and routing-variation terms.
\end{itemize}
Any model whose induced $(f_0,\phi)$ representation and task experts satisfy the corresponding assumptions inherits the guarantees of Section~\ref{sec:theory} with the associated constants. Section~\ref{sec:FedSEE} develops the LoRA instantiation. We give two additional examples below.

\paragraph{Linear probing and last-layer personalization.}
Let $\phi$ be a frozen pre-trained encoder and let $\theta$ denote the trainable linear head. In this case, Assumption~\ref{ass:linear}(i) has the usual interpretation that each task is well specified within the shared representation. Assumption~\ref{ass:linear}(ii) additionally requires the tasks to induce a common feature second moment, which need not hold for a general pre-trained encoder because different task distributions may induce different feature covariances. When these covariances are only approximately shared, the covariance-mismatch analysis of Remark~\ref{rem:heterogeneous} in Appendix~\ref{app:geometry} provides the corresponding perturbed interpretation through $\rho_\Phi$.

\paragraph{Random features and fixed kernel representations.}
Let $\phi$ be a fixed finite-dimensional random-feature map, or a fixed feature representation associated with a kernel, and let $\theta$ denote the corresponding linear coefficient block. Bounded features directly satisfy the representation bound of Assumption~\ref{ass:bounded}, while feature non-degeneracy gives the corresponding covariance lower bound. The remaining conditions are not automatic. Task realizability requires the task predictors to lie in the span of the chosen feature map, and Assumption~\ref{ass:linear}(ii) requires the induced feature covariance to be shared across tasks. When the latter holds only approximately, Remark~\ref{rem:heterogeneous} again provides the appropriate covariance-perturbation interpretation.

These examples illustrate that the framework is not specific to LoRA, while its guarantees remain conditional on the induced representation and task geometry satisfying the stated assumptions. Directions for relaxing these assumptions are discussed in Section~\ref{sec:conclusion}.

\subsection{Deep-Model Extension for FedSEE}
\label{app:fedsee_reduction}
We next return to the deep FedSEE instantiation. The preceding analysis applies to the frozen-feature model in~\eqnref{eq:model_class}. At a single adapted projection, fixing $A^{\mathrm{global}}$ makes the routed contribution linear in the $B_m$ blocks, so the reduction is exact conditional on the incoming representation, as detailed in Appendix~\ref{app:fedsee_projection}. In the full transformer, however, an adapted projection changes the representations passed to subsequent layers, and query adaptation may also modify the attention weights. Proposition~\ref{prop:deep_reduction} controls the resulting deviation from a local first-order model, and Corollary~\ref{cor:deep_barycenter} translates this approximation into a bound on the departure from exact barycentric geometry.

\begin{proposition}[Deep-model perturbation]
\label{prop:deep_reduction}
Let $\vartheta$ collect the routed $B$ blocks across all adapted projections and let $F_{\vartheta}$ denote the resulting network. Fix a reference $\vartheta_0$ and suppose the parameter Jacobian is $H$-Lipschitz on a convex neighborhood $\|\vartheta-\vartheta_0\|_{\mathcal B}\le R_B$, where $\|\vartheta\|_{\mathcal B}^2$ is the sum of squared Frobenius norms of its constituent blocks. Then, for every $\Delta$ such that $\vartheta_0+\Delta$ remains in this neighborhood,
\begin{equation}
F_{\vartheta_0+\Delta}(x) = F_{\vartheta_0}(x)+J_0(x)\Delta+r_\Delta(x), \qquad \|r_\Delta(x)\|_2 \le \frac{H}{2}\|\Delta\|_{\mathcal B}^2, \label{eq:deep_linearization}
\end{equation}
where $J_0(x)$ is the Jacobian of $F_\vartheta(x)$ with respect to $\vartheta$ evaluated at $\vartheta_0$.
\end{proposition}

\paragraph{Proof.}

Fix an input $x$ and define the path
\[
g(t):=F_{\vartheta_0+t\Delta}(x), \qquad t\in[0,1].
\]
Because the line segment $\{\vartheta_0+t\Delta:t\in[0,1]\}$ lies in the assumed neighborhood, the fundamental theorem of calculus gives
\[
F_{\vartheta_0+\Delta}(x)-F_{\vartheta_0}(x) = \int_0^1 J_{\vartheta_0+t\Delta}(x)\Delta\,dt .
\]
Adding and subtracting $J_{\vartheta_0}(x)\Delta$ yields
\[
F_{\vartheta_0+\Delta}(x) = F_{\vartheta_0}(x) + J_{\vartheta_0}(x)\Delta + r_\Delta(x),
\]
where
\[
r_\Delta(x) := \int_0^1 \left( J_{\vartheta_0+t\Delta}(x) - J_{\vartheta_0}(x) \right)\Delta\,dt .
\]

Let $\|\cdot\|_{\mathrm{op}}$ denote the operator norm induced by $\|\cdot\|_{\mathcal B}$ on parameter perturbations and $\|\cdot\|_2$ on network outputs. By the $H$-Lipschitz assumption on the Jacobian,
\[
\left\| J_{\vartheta_0+t\Delta}(x) - J_{\vartheta_0}(x) \right\|_{\mathrm{op}} \le Ht\|\Delta\|_{\mathcal B}.
\]
Therefore,
\begin{align*}
\|r_\Delta(x)\|_2 &\le \int_0^1 \left\| J_{\vartheta_0+t\Delta}(x) - J_{\vartheta_0}(x) \right\|_{\mathrm{op}} \|\Delta\|_{\mathcal B}\,dt\\
&\le \int_0^1 Ht\|\Delta\|_{\mathcal B}^2\,dt\\
&= \frac{H}{2}\|\Delta\|_{\mathcal B}^2.
\end{align*}
Finally, for fixed $\vartheta_0$ and $x$, $J_{\vartheta_0}(x)\Delta$ is linear in the concatenated parameter perturbation $\Delta$. Thus the first-order network is a generalized affine predictor in $\Delta$ with input-dependent design operator $J_0(x)$. This form need not have the separable output-block structure $\theta\phi(x)$ of~\eqnref{eq:model_class}.\qed

\begin{assumption}[Local first-order geometry]
\label{ass:deep_linear}
After vectorizing $\Delta$, write the first-order predictor as
\[
F_{\mathrm{lin},v}(x) = F_{\vartheta_0}(x)+J_0(x)v .
\]
For each task $m$, there exists $v_m^*$ such that
\[
y = F_{\vartheta_0}(x)+J_0(x)v_m^*+\zeta_m, \qquad \mathbb E[\zeta_m\mid x]=0.
\]
On the analyzed parameter subspace, the taskwise Jacobian Gram matrices satisfy
\[
G_m := \mathbb E_{\mathcal P_m} [J_0(x)^\top J_0(x)] = G \succ 0
\]
for all $m$.
\end{assumption}

Under Assumption~\ref{ass:deep_linear}, the same quadratic argument as in Theorem~\ref{thm:mixture} gives $v_i^*=\sum_m\alpha_{im}v_m^*$ for the first-order client optimum. Since $\vartheta=\vartheta_0+v$ and $\sum_m\alpha_{im}=1$, the corresponding first-order optima also satisfy $\vartheta_{i,\mathrm{lin}}^* =\sum_m\alpha_{im}\vartheta_{m,\mathrm{lin}}^*$. The following corollary relates this geometry to the optima of the full deep model.

\begin{corollary}[Approximate barycentric geometry of the deep model]
\label{cor:deep_barycenter}
Suppose Assumption~\ref{ass:deep_linear} holds and that the task and client minimizers of both the deep and first-order risks lie in the parameter neighborhood of Proposition~\ref{prop:deep_reduction}. Suppose further that, on this neighborhood, each task risk and each client risk is uniformly within $\tau_{\mathrm{deep}}$ of its corresponding first-order risk, and that the first-order risks are $\mu$-strongly convex on the analyzed parameter subspace. Then
\[
\|\vartheta_m^*-\vartheta_{m,\mathrm{lin}}^*\|_{\mathcal B} \le 2\sqrt{\frac{\tau_{\mathrm{deep}}}{\mu}}
\]
for every task $m$, and likewise
\[
\|\vartheta_i^*-\vartheta_{i,\mathrm{lin}}^*\|_{\mathcal B} \le 2\sqrt{\frac{\tau_{\mathrm{deep}}}{\mu}}
\]
for every client $i$. Consequently,
\begin{equation}
\left\| \vartheta_i^* - \sum_{m=1}^M\alpha_{im}\vartheta_m^* \right\|_{\mathcal B} \le 4\sqrt{\frac{\tau_{\mathrm{deep}}}{\mu}}. \label{eq:deep_barycentric}
\end{equation}
\end{corollary}

\paragraph{Proof.}
Let $R$ denote one of the deep risks and $R_{\mathrm{lin}}$ its first-order counterpart, with
\[
\sup_{\vartheta}|R(\vartheta)-R_{\mathrm{lin}}(\vartheta)| \le\tau_{\mathrm{deep}}.
\]
Let $\vartheta^*$ and $\vartheta_{\mathrm{lin}}^*$ minimize $R$ and $R_{\mathrm{lin}}$, respectively. Then
\[
R_{\mathrm{lin}}(\vartheta^*) \le R(\vartheta^*)+\tau_{\mathrm{deep}} \le R(\vartheta_{\mathrm{lin}}^*)+\tau_{\mathrm{deep}} \le R_{\mathrm{lin}}(\vartheta_{\mathrm{lin}}^*) +2\tau_{\mathrm{deep}}.
\]
By $\mu$-strong convexity of $R_{\mathrm{lin}}$,
\[
R_{\mathrm{lin}}(\vartheta^*) - R_{\mathrm{lin}}(\vartheta_{\mathrm{lin}}^*) \ge \frac{\mu}{2} \|\vartheta^*-\vartheta_{\mathrm{lin}}^*\|_{\mathcal B}^2,
\]
and hence
\[
\|\vartheta^*-\vartheta_{\mathrm{lin}}^*\|_{\mathcal B} \le 2\sqrt{\frac{\tau_{\mathrm{deep}}}{\mu}}.
\]

Applying this once to client $i$ and once to every task, and using Assumption~\ref{ass:deep_linear} and the generalized affine barycentric identity established above,
\[
\vartheta_{i,\mathrm{lin}}^* = \sum_m\alpha_{im}\vartheta_{m,\mathrm{lin}}^*,
\]
gives
\begin{align*}
\left\| \vartheta_i^*-\sum_m\alpha_{im}\vartheta_m^* \right\|_{\mathcal B} &\le \|\vartheta_i^*-\vartheta_{i,\mathrm{lin}}^*\|_{\mathcal B}\\
&\quad+ \sum_m\alpha_{im} \|\vartheta_{m,\mathrm{lin}}^*-\vartheta_m^*\|_{\mathcal B}\\
&\le 4\sqrt{\frac{\tau_{\mathrm{deep}}}{\mu}},
\end{align*}
since $\sum_m\alpha_{im}=1$.\qed
\subsection{Recovery under Balanced Transport}
\label{app:balanced_recovery}

Theorem~\ref{thm:recovery} analyzes the supervised extraction problem in~\eqnref{eq:gen_extract} independently for each task. FedSEE instantiates this extractor through~\eqnref{eq:omega_opt} and optimizes the taskwise programs jointly under a balanced-transport constraint on the client contributions. This coupling reduces the joint feasible set relative to independent extraction and can therefore increase the taskwise extraction loss. The following proposition quantifies the resulting effect on expert recovery through the coupling excess $\delta_m^{\mathrm{BT}}$. The corresponding optimization procedure is described in Appendix~\ref{app:fedsee_sinkhorn}.

\begin{proposition}[Recovery under balanced coupling]
\label{prop:balanced_coupling}
Under Assumptions~\ref{ass:public}, \ref{ass:linear}, and~\ref{ass:anchor}, with exact client uploads and population public loss, let $\theta_m^{\mathrm{ind}}$ denote the independent extractor certified by Theorem~\ref{thm:recovery}, and let $\theta_m^{\mathrm{BT}}$ denote expert $m$ produced by the jointly balanced program. Define the coupling excess
\[
\delta_m^{\mathrm{BT}} := \mathcal L_m(\theta_m^{\mathrm{BT}}) - \mathcal L_m(\theta_m^{\mathrm{ind}}) \ge0.
\]
Then
\begin{equation}
\|\theta_m^{\mathrm{BT}}-\theta_m^*\|_{\mathrm F}^2 \le \kappa(\Phi)\, \varepsilon^2 \mathrm{diam}_{\mathrm F}(\Theta)^2 + \frac{\delta_m^{\mathrm{BT}}} {\sigma_{\min}(\Phi)}. \label{eq:balanced_coupling_bound}
\end{equation}
Hence the independent recovery guarantee is retained whenever the transport constraint is nonbinding ($\delta_m^{\mathrm{BT}}=0$). In particular, transport-compatible pure anchors give exact population recovery.
\end{proposition}

\paragraph{Proof.}
Because each row of the balanced solution remains a convex combination of the upload pool, $\theta_m^{\mathrm{BT}}$ is feasible for the corresponding independent extraction problem. Since $\theta_m^{\mathrm{ind}}$ minimizes the task loss over this feasible set,
\[
\mathcal L_m(\theta_m^{\mathrm{BT}}) \geq \mathcal L_m(\theta_m^{\mathrm{ind}}),
\]
and hence $\delta_m^{\mathrm{BT}}\geq0$.

Under Assumption~\ref{ass:linear}, the population task loss satisfies
\[
\mathcal L_m(\theta)-\mathcal L_m(\theta_m^*) = \|\theta-\theta_m^*\|_\Phi^2.
\]
By the definition of $\delta_m^{\mathrm{BT}}$,
\begin{align*}
    \|\theta_m^{\mathrm{BT}}-\theta_m^*\|_\Phi^2 &= \mathcal L_m(\theta_m^{\mathrm{BT}}) - \mathcal L_m(\theta_m^*) \\
    &= \mathcal L_m(\theta_m^{\mathrm{ind}}) - \mathcal L_m(\theta_m^*) + \delta_m^{\mathrm{BT}} \\
    &= \|\theta_m^{\mathrm{ind}}-\theta_m^*\|_\Phi^2 + \delta_m^{\mathrm{BT}}.
\end{align*}

The population recovery guarantee for the independent extractor gives
\[
\|\theta_m^{\mathrm{ind}}-\theta_m^*\|_\Phi \leq \sigma_{\max}(\Phi)^{1/2} \varepsilon\, \mathrm{diam}_{\mathrm F}(\Theta).
\]
Therefore,
\[
\|\theta_m^{\mathrm{BT}}-\theta_m^*\|_\Phi^2 \leq \sigma_{\max}(\Phi) \varepsilon^2 \mathrm{diam}_{\mathrm F}(\Theta)^2 + \delta_m^{\mathrm{BT}}.
\]
Finally, using
\[
\|A\|_{\mathrm F}^2 \leq \sigma_{\min}(\Phi)^{-1}\|A\|_\Phi^2
\]
gives
\[
\|\theta_m^{\mathrm{BT}}-\theta_m^*\|_{\mathrm F}^2 \leq \kappa(\Phi) \varepsilon^2 \mathrm{diam}_{\mathrm F}(\Theta)^2 + \frac{\delta_m^{\mathrm{BT}}}{\sigma_{\min}(\Phi)},
\]
which is~\eqnref{eq:balanced_coupling_bound}. \qed

\subsection{Stability of the Deployed Contribution Matrix}
\label{app:c_ema_stability}

FedSEE applies temporal smoothing to the balanced contribution matrices produced across communication rounds before using them for capacity assignment and router calibration. We next characterize the feasibility and tracking behavior of this smoothing step.

Define the balanced-transport polytope
\begin{equation}
    \mathcal C := \left\{ C\in\mathbb{R}_+^{M\times N}: C\mathbf{1}_N=\mathbf{1}_M,\; C^\top\mathbf{1}_M=\frac{M}{N}\mathbf{1}_N \right\}. \label{eq:transport_polytope}
\end{equation}
Every raw matrix $C^{*,(t)}$ produced by the balanced extraction lies in $\mathcal C$, and the initialization matrix $C^{(1)}$ is obtained from the same extraction program. Since $\mathcal C$ is convex, the EMA recursion in Proposition~\ref{prop:c_ema} preserves these constraints at every round. The proposition also quantifies how closely the smoothed matrix tracks a changing sequence of raw extraction matrices.

\begin{proposition}[Stability of the deployed contribution EMA]
\label{prop:c_ema}
Let $C^{*,(t)}\in\mathcal C$ denote the raw balanced contribution matrix returned by extraction at round $t$, let $C^{(1)}\in\mathcal C$, and let the deployed matrix follow
\[
C^{(t+1)} = (1-\lambda)C^{(t)} + \lambda C^{*,(t)}, \qquad \lambda\in(0,1).
\]
Then $C^{(t)}\in\mathcal C$ for every $t$.

Suppose additionally that the roundwise raw extraction matrices satisfy
\[
\|C^{*,(t+1)}-C^{*,(t)}\|_{\mathrm F} \leq \zeta_C^{(t)},
\]
and define
\[
\bar{\zeta}_C := \limsup_{t\to\infty}\zeta_C^{(t)}.
\]
Then
\begin{equation}
    \limsup_{t\to\infty} \|C^{(t)}-C^{*,(t)}\|_{\mathrm F} \leq \frac{\bar{\zeta}_C}{\lambda}. \label{eq:c_ema_tracking}
\end{equation}
In particular, if $\zeta_C^{(t)}\to0$, then
\[
\|C^{(t)}-C^{*,(t)}\|_{\mathrm F}\to0.
\]
If the raw extraction matrix is stationary, $C^{*,(t)}\equiv C^*$, then the stronger geometric bound holds:
\begin{equation}
    \|C^{(t)}-C^*\|_{\mathrm F} \leq (1-\lambda)^{t-1} \|C^{(1)}-C^*\|_{\mathrm F}. \label{eq:c_ema_stationary}
\end{equation}
\end{proposition}

\paragraph{Proof.}
Because $\mathcal C$ is convex and both $C^{(t)}$ and $C^{*,(t)}$ lie in $\mathcal C$,
\[
C^{(t+1)} = (1-\lambda)C^{(t)} + \lambda C^{*,(t)} \in\mathcal C.
\]
Since $C^{(1)}\in\mathcal C$, induction proves feasibility for every round.

For the tracking result, define
\[
E_C^{(t)} := C^{(t)}-C^{*,(t)}.
\]
Subtracting $C^{*,(t+1)}$ from the EMA recursion gives
\begin{align}
    E_C^{(t+1)} &= (1-\lambda)C^{(t)} + \lambda C^{*,(t)} - C^{*,(t+1)} \nonumber\\
    &= (1-\lambda)E_C^{(t)} - \bigl( C^{*,(t+1)}-C^{*,(t)} \bigr). \label{eq:c_ema_error}
\end{align}
Therefore,
\begin{equation}
    \|E_C^{(t+1)}\|_{\mathrm F} \leq (1-\lambda)\|E_C^{(t)}\|_{\mathrm F} + \zeta_C^{(t)}. \label{eq:c_ema_error_bound}
\end{equation}
Iterating this inequality yields
\begin{equation}
    \|E_C^{(t)}\|_{\mathrm F} \leq (1-\lambda)^{t-1} \|E_C^{(1)}\|_{\mathrm F} + \sum_{s=1}^{t-1} (1-\lambda)^{t-1-s} \zeta_C^{(s)}. \label{eq:c_ema_unroll}
\end{equation}

To control the final term, fix $\epsilon>0$. By the definition of $\bar{\zeta}_C$, there exists $S$ such that
\[
\zeta_C^{(s)} \leq \bar{\zeta}_C+\epsilon \qquad \text{for all }s\geq S.
\]
Split the convolution in~\eqnref{eq:c_ema_unroll} at $S$. The finite head satisfies
\[
\sum_{s=1}^{S-1} (1-\lambda)^{t-1-s} \zeta_C^{(s)} \longrightarrow 0
\]
as $t\to\infty$, while the tail is bounded by
\begin{align}
    \sum_{s=S}^{t-1} (1-\lambda)^{t-1-s} \zeta_C^{(s)} &\leq (\bar{\zeta}_C+\epsilon) \sum_{j=0}^{\infty}(1-\lambda)^j \nonumber\\
    &= \frac{\bar{\zeta}_C+\epsilon}{\lambda}.
\end{align}
The initialization term in~\eqnref{eq:c_ema_unroll} also vanishes geometrically. Taking $\limsup_{t\to\infty}$ and then letting $\epsilon\downarrow0$ proves~\eqnref{eq:c_ema_tracking}.

If $\zeta_C^{(t)}\to0$, then $\bar{\zeta}_C=0$, giving exact tracking. If $C^{*,(t)}\equiv C^*$, the drift term in \eqnref{eq:c_ema_error} vanishes identically, and direct iteration gives \eqnref{eq:c_ema_stationary}. \qed

\paragraph{Relation to recovery and mixture estimation.}
Proposition~\ref{prop:c_ema} concerns only the temporal behavior of the smoothed contribution matrix. It guarantees that the balanced-transport constraints are preserved and bounds the discrepancy between the smoothed matrix $C^{(t)}$ and the roundwise extraction matrix $C^{*,(t)}$. It does not imply that either matrix recovers the true client mixtures.

Mixture estimation is instead characterized by Proposition~\ref{prop:mixture_est}, which bounds the error of the barycentric estimate $\hat{\alpha}_i$. FedSEE uses the normalized contribution vector $\tilde c_i$ as a computational proxy for this estimate, with discrepancy $\delta_i^{\mathrm{proxy}} =\|\tilde c_i-\hat{\alpha}_i\|_1$.

Similarly, Proposition~\ref{prop:finitesample} applies to an upload pool satisfying its shared-representation, anchor, and estimation conditions. These are conditions on the upload pool rather than guarantees of the FedSEE training procedure. In particular, initialization trains $B_i$ jointly with client-specific $A_i$ before forming $A^{\mathrm{global}}$, while later uploads arise from joint routed training. The proposition applies whenever the resulting upload pool satisfies the stated conditions.


\end{document}